\documentclass{article}

\PassOptionsToPackage{numbers, compress}{natbib}  
\pdfoutput=1

\usepackage[final]{neurips_2022}

\usepackage[utf8]{inputenc} 
\usepackage[T1]{fontenc}    
\usepackage{hyperref}       
\usepackage{url}            
\usepackage{booktabs}       
\usepackage{amsfonts}       
\usepackage{nicefrac}       
\usepackage{microtype}      

\usepackage[dvipsnames]{xcolor}
\usepackage{sidecap} 
\usepackage{wrapfig}

\usepackage{setspace}

\usepackage{graphicx}
\usepackage{amsmath}
\usepackage{amssymb}
\usepackage{booktabs}
\usepackage{bm}
\usepackage{float}
\usepackage{multicol}
\usepackage{multirow}
\usepackage{caption}
\usepackage{graphicx}
\usepackage{adjustbox}

\usepackage{array}
\newcolumntype{P}[1]{>{\centering\arraybackslash}p{#1}}
\usepackage{dblfloatfix}
\usepackage{enumitem}
\usepackage{mathabx}

\usepackage{algpseudocode}
\usepackage{placeins}
\usepackage[linesnumbered,ruled,vlined]{algorithm2e}
\SetKwInput{KwInput}{Input}  
\SetKwInput{KwOutput}{Output}
\SetKwInput{KwRequire}{Require}
\SetKwInput{KwIP}{\em Importance Probing}
\SetKwInput{KwMA}{\em Main Adaptation}

\SetCommentSty{mycommfont}

\newcommand{\eg}{\textit{e.g.}}
\newcommand{\ie}{\textit{i.e.}}

\usepackage{amssymb}
\usepackage{pifont}
\newcommand{\cmark}{\ding{51}}%
\newcommand{\xmark}{\ding{55}}%

\title{Few-shot Image Generation via \\ Adaptation-Aware Kernel Modulation
}
\author{%
  Yunqing Zhao$^*$\\
  \texttt{yunqing\_zhao@mymail.sutd.edu.sg} \And 
  Keshigeyan Chandrasegaran$^*$\\
  \texttt{keshigeyan@sutd.edu.sg} \And
  Milad Abdollahzadeh$^*$\\
  \texttt{milad\_abdollahzadeh@sutd.edu.sg}\And
  Ngai-Man Cheung$^\dag$\\
  \texttt{ngaiman\_cheung@sutd.edu.sg}\and
  \\
  Singapore University of Technology and Design (SUTD)
}

\begin{document}

\maketitle
\def\thefootnote{*}
\footnotetext{Equal Contribution \hspace{3 mm} $^\dag$Corresponding Author}
\def\thefootnote{\arabic{footnote}}

\vspace{-0.3cm}
\begin{abstract}

Few-shot image generation (FSIG) aims to learn to generate new and diverse samples given an extremely limited number of samples from a domain, \eg, 10 training samples. 
Recent work has addressed the problem using transfer learning approach, leveraging a GAN pretrained on a large-scale source domain dataset and adapting that model to the target domain based on very limited target domain samples.
Central to recent FSIG methods are  {\em knowledge preserving criteria}, which aim to select a subset of 
source model's knowledge to be preserved into the adapted model.
However, a  {\em major limitation} of existing methods is that their
knowledge preserving
criteria consider {\em only source domain/source task}, and they   fail to consider {\em target domain/adaptation task} in  selecting source model's
knowledge, 
casting doubt on their suitability for setups of different {\em proximity} between source and target domain.
{\bf Our work} makes two contributions. As our first contribution, 
we revisit recent FSIG
works and their experiments.
Our important finding is that, under setups which
assumption of 
close proximity between 
source and target domains is relaxed,  
existing state-of-the-art 
(SOTA)
methods which 
consider only 
source domain
in knowledge preserving perform {\em no better} than a baseline fine-tuning method.
To address the limitation of existing methods, as our second contribution, 
we propose {\em Adaptation-Aware kernel
Modulation} (AdAM) to address general FSIG of different source-target domain proximity.
Extensive
experimental results show that the proposed method consistently achieves 
SOTA
performance across source/target domains of different proximity, including challenging setups when source and target domains are more apart. 
 Project Page:
  \href{https://yunqing-me.github.io/AdAM/}{{\color{RubineRed}{https://yunqing-me.github.io/AdAM/}}}
\end{abstract}

\section{Introduction}
\label{section1}

Generative Adversarial Networks (GANs) \cite{goodfellow2014GAN, brock2018bigGAN, karras2020styleganv2} have been applied to a range of important applications including image generation \cite{karras2018styleGANv1, karras2020styleganv2, choi2020starganv2}, image-to-image translation \cite{zhu2017cycleGAN, liu2019FUNit}, image editing \cite{lin2021anycostGANs, liu2021denet_imageediting}, 
anomaly detection \cite{lim2018doping}, and data augmentation \cite{tran2021data_aug_gan, chai2021ensembling}. 
However, a critical issue is that these GANs often require large-scale datasets and computationally expensive resources to achieve good performance.
For example,  StyleGAN \cite{karras2018styleGANv1} is trained on Flickr-Faces-HQ (FFHQ) \cite{karras2018styleGANv1} that contains 70,000 images.
However, in many practical applications only a few samples are available (\eg, 
photos of rare animal species / skin diseases).
Training a generative model is problematic in this low-data regime, where the generator often suffers from mode collapse or blurred generated images \cite{feng2021WhenGansReplicate, ojha2021fig_cdc, noguchi2019BSA}.
To address this, {\em few-shot image generation} (FSIG) studies the possibility of generating sufficiently diverse and high quality images, given very limited training data (\eg, 10 samples). FSIG also attracts an increasing interest for some downstream tasks, \eg, few-shot classification \cite{chai2021ensembling}. 

\begin{table*}[!t]
    \centering
    \caption{Transfer learning for few-shot image generation: Various criteria are proposed to {\em augment} baseline transfer learning to preserve subset of source model’s knowledge into the adapted model.
    }
    \begin{adjustbox}{width=\textwidth,center}
        \begin{tabular}{p{1.8cm}|P{9.6cm}|P{2.3cm}|P{3.0cm}}
        \toprule
        
        \textbf{Method} &\textbf{Knowledge preserving criteria}
        & Source domain/task \newline aware 
        & Target domain/adaptation \newline aware 
        \\ \hline
        
        TGAN \cite{wang2018transferringGAN} &
        Not available 
        & --
        & --
        \\ \hline
        FreezeD \cite{mo2020freezeD} & 
                Preservation of lower layers of the discriminator pre-trained on the {\em source} domain.
        & {\color{black}\cmark}
        & {\color{black}\xmark}
        \\ \hline
        EWC \cite{li2020fig_EWC} & 
                Preservation of weights important to the {\em source} generative model pre-trained on the {\em source} domain.
        & {\color{black}\cmark}
        & {\color{black}\xmark}
        \\ \hline
        CDC \cite{ojha2021fig_cdc} & 
                Preservation of pairwise distances of generated images by the {\em source} generative model pre-trained on the {\em source} domain.
        & {\color{black}\cmark}
        & {\color{black}\xmark}
        \\ \hline
        DCL \cite{zhao2022dcl} & 
                Preservation of multilevel semantic diversity of the generated images by the {\em source} generative model pre-trained on the {\em source} domain.
        & {\color{black}\cmark}
        & {\color{black}\xmark}
        \\ \hline
        {\bf AdAM  (Our work)} & 
        Preservation of kernels important in  {\em \textbf{adaptation}} of source model to {\em target}. 
        & {\color{black}\cmark}
        & {\color{black}\cmark}
         \\
        \bottomrule
        \end{tabular}
    \end{adjustbox}
    \label{table:criteria}
    \vspace{-6mm}
\end{table*}

{\bf FSIG with Transfer Learning.} Recent works in FSIG are based on transfer learning approach \cite{pan2009yang-qiang-transfer} \ie, leveraging the prior knowledge of a GAN pretrained on a large-scale, diverse source dataset (\eg, FFHQ \cite{karras2018styleGANv1} or ImageNet \cite{deng2009imagenet}) and adapting it to a target domain with very limited samples (\eg, face paintings \cite{yaniv2019faceofart}).
As only very limited samples are provided to define the underlying distribution, standard fine-tuning of a pre-trained GAN suffers from mode collapse: the adapted model can only generate samples closely resembling the given few shot target samples \cite{wang2018transferringGAN,ojha2021fig_cdc}.
Therefore, recent works \cite{li2020fig_EWC, ojha2021fig_cdc, zhao2022dcl}
have proposed to {\em augment} standard fine-tuning with different criteria to carefully preserve subset of source model's knowledge into the adapted model.
Various criteria has been proposed (Table~\ref{table:criteria}), and these 
{\em knowledge preserving criteria} have been central in recent FSIG research.
In general, these criteria aim to preserve subset of 
source model's knowledge which is deemed to be useful for target-domain sample generation, \eg, improving the diversity of target sample generation.

{\bf Research Gaps.} One major limitation of existing methods is that 
they consider {\em only} source domain in preserving subset of source model's knowledge into the adapted model.
In particular, these methods {\em fail to consider} target domain/adaptation task in selection of source model's knowledge (Table \ref{table:criteria}).
For example, EWC \cite{li2020fig_EWC} applies Fisher Information \cite{ly2017tutorial} to select important weights entirely based on the pretrained {\em source} model, and it aims to preserve these selected weights regardless of the target domain in adaptation.
Similar to EWC \cite{li2020fig_EWC}, CDC \cite{ojha2021fig_cdc} proposes an additional constraint to preserve pairwise distances of generated images by the {\em source} model, and there is no consideration of target domain/adaptation.
These {\em target/adaptation-agnostic} knowledge preserving criteria in recent works raise question regarding their suitability in different source/target domain setups.
It should be noted that  
existing FSIG works
(under very limited target samples) focus largely on setups where source and target domains are in {\em close proximity} (semantically) \eg, Human faces (FFHQ)$\rightarrow$Baby faces \cite{ojha2021fig_cdc,zhao2022dcl}, or Cars$\rightarrow$Abandoned Cars \cite{ojha2021fig_cdc,zhao2022dcl}.
It is unclear about their performance when source/target domains are more apart
(\eg, Human faces (FFHQ) $\rightarrow$ Animal faces \cite{choi2020starganv2}).

{\bf Contributions.} In this paper we take an important step to address these research gaps for FSIG. 
Specifically,
our work makes two contributions. {\bf As our first contribution}, we
revisit existing state-of-the-art (SOTA) algorithms and their experiments.
Importantly, we observe that when the close proximity assumption is relaxed in experiment setups and source/target domains are more apart,
existing 
SOTA
methods perform {\em no better} than a baseline fine-tuning method.
Our observation suggests that recent methods
considering only source domain/source task in knowledge preserving 
may not be suitable for {\em general} 
FSIG
when source and target domains are more apart.
To validate our claims,
we introduce additional experiments with 
different source/target domains,
analyze their proximity qualitatively and quantitatively, and examine existing methods under a unified framework.

Informed by our analysis, {\bf as our second contribution}, we propose an {\em adaptation-aware kernel modulation} approach 
to address general 
FSIG
of different source/target domain proximity.
In marked contrast to existing works which preserve
knowledge important to {\em source} task,
our method aims to preserve subset of source model's knowledge that are important to the {\em target} domain and the {\em adaptation} task.
More specifically,
we propose an {\em importance probing} algorithm to identify
kernels
which encode important knowledge for adaptation to the target domain. 
Then, we preserve the knowledge of these 
kernels
using a 
parameter-efficient {\em rank-constrained kernel modulation}.
We conduct 
extensive experiments to show that our proposed method consistently achieves 
SOTA
performance across source/target domains of different proximity, including challenging setups when source/target domains are more apart. 
Our main contributions are summarized as follows:
\begin{itemize}
    \item 
    {We revisit existing FSIG methods and experiment setups. Our study uncovers  issues with existing methods when applied to source/target domains of different proximity.}
    \item 
    {We propose Adaptation-Aware kernel Modulation (AdAM) for FSIG. Our method consistently achieves 
    SOTA
    performance both visually and quantitatively across source/target domains with different proximity. 
    }
\end{itemize}
\vspace{-2mm}

\begin{figure}[!t]
    \centering
    \includegraphics[width=\textwidth]{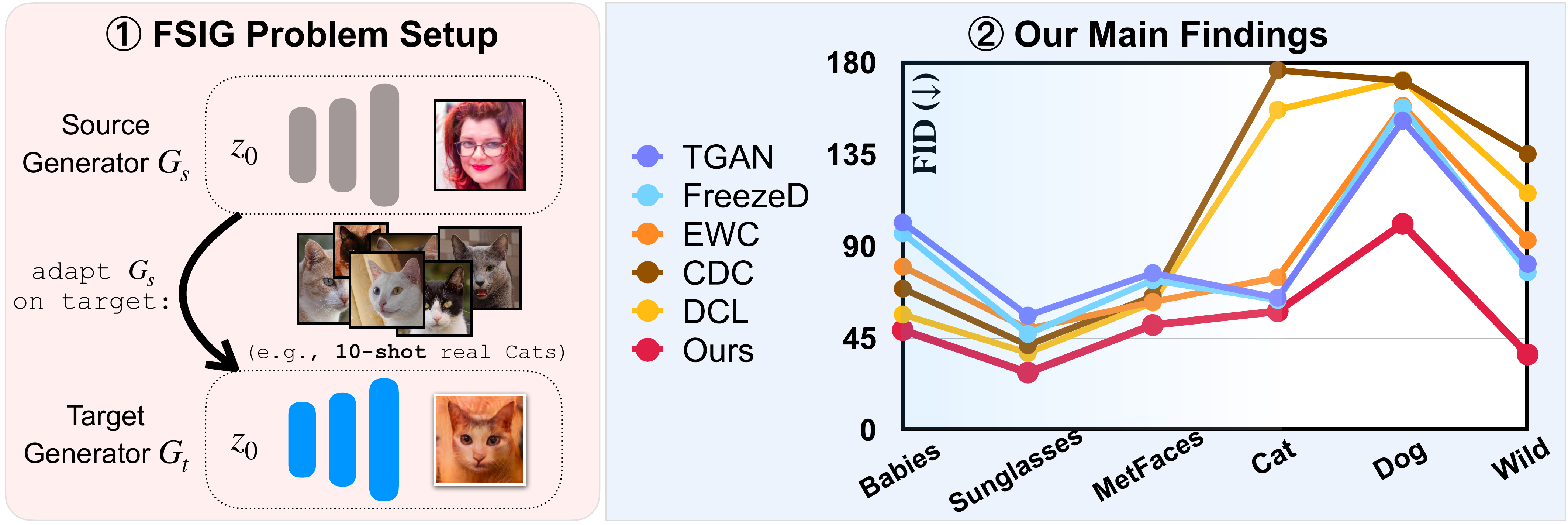}
    \includegraphics[width=\textwidth]{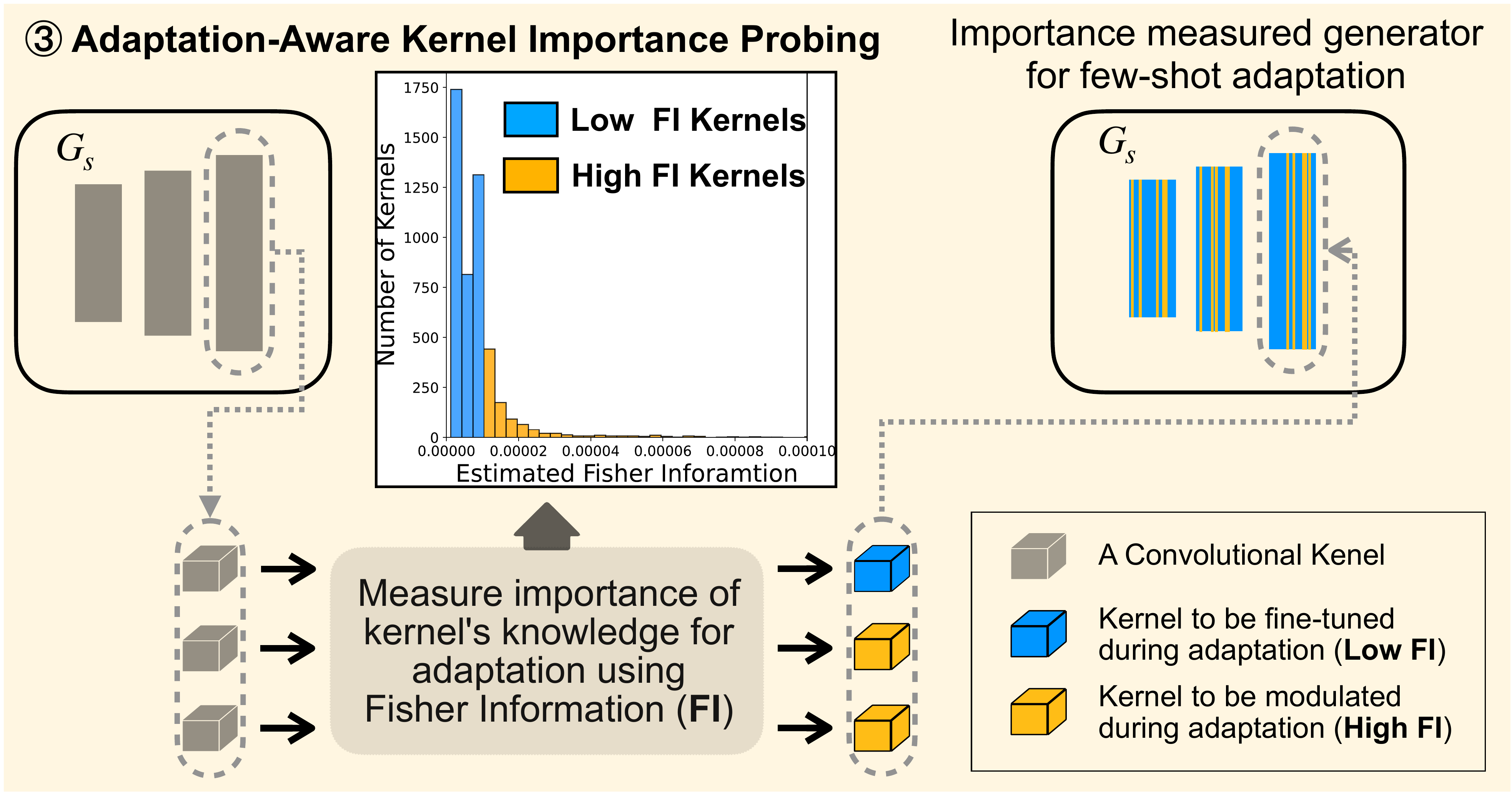}
    \vspace{-6mm}
    \caption{
    \textit{Overview and our contributions.}
    \textcircled{\raisebox{-0.9pt}{1}}: 
    We consider the problem of FSIG with Transfer Learning using very limited target samples (\ie 10-shot).
    \textcircled{\raisebox{-0.9pt}{2}}:
    Our work makes two contributions, $\bullet$ We discover that when the close proximity assumption between source-target domain is relaxed, SOTA FSIG methods (EWC \cite{li2020fig_EWC}, CDC \cite{ojha2021fig_cdc}, DCL \cite{zhao2022dcl})
    which consider only source domain/source task in knowledge preserving 
    perform no better than a baseline fine-tuning method (TGAN \cite{wang2018transferringGAN}) (Sec \ref{sec:proximity-transferability}).
    $\bullet$ We propose a novel adaptation-aware kernel modulation for FSIG that achieves SOTA performance across source / target domains with different proximity (Sec \ref{section4}). 
    \textcircled{\raisebox{-0.9pt}{3}} Schematic diagram of our proposed Importance Probing Mechanism: 
    We measure the importance of each kernel for the target domain after probing and 
    preserve source domain knowledge that is important for target domain adaptation (Sec \ref{section4}). The same operations are applied to discriminator.
    }
    \label{fig:overview}
    \vspace{-0.5cm}
\end{figure}

\section{Related Work}
\label{section2}

\textbf{Few-shot image generation.}
 Conventional few-shot learning \cite{finn2017maml, fei2006one, snell2017prototypical} aims at learning a discriminative classifier for classification \cite{guo2020awgim, sun2021explanation, milad2021revisit, zhao2022fs}, segmentation \cite{liu2020crnet_fs_seg, boudiaf2021few} or detection \cite{zhang2021pnpdet, fan2021generalized, gong2022meta} tasks.
 Differently, few-shot image generation (FSIG)
 \cite{ojha2021fig_cdc, li2020fig_EWC, zhao2022dcl}
 aims at learning a generator for new and diverse samples given extremely limited samples (\eg, 10 shots). Transfer learning has been applied to FSIG. 
 For example, Transferring GAN \cite{wang2018transferringGAN} (\textbf{TGAN}) applies simple GAN loss \cite{goodfellow2014GAN} to fine-tune all parameters of both the generator and the discriminator. \textbf{FreezeD} \cite{mo2020freezeD} fixes a few high-resolution discriminator layers during fine-tuning.
 To augment and improve simple fine-tuning, more recent 
 works have focused on preserving specific knowledge from the source models.  
 Elastic weight consolidation 
 (\textbf{EWC}) \cite{li2020fig_EWC} identifies important weights for the \textit{source} model and tries to preserve these weights. 
 Cross-domain Correspondence (\textbf{CDC}) \cite{ojha2021fig_cdc} preserves pair-wise distance of generated images from the source model to alleviate mode collapse.  
 Dual Contrastive Learning (\textbf{DCL}) \cite{zhao2022dcl} applies mutual information maximization to preserve multi-level diversity of the generated images by the source model.
 In this work, we observe that these 
 SOTA
 methods perform poorly when source and target domains are more apart. Therefore, their proposed source knowledge preservation criteria {\em may not} be generalizable. 
 Based on our analysis, we propose an adaptation-aware knowledge selection which is more {\em generalizable} for 
 source/target domains with different proximity.
 \vspace{-2mm}

\section{Revisiting FSIG through the Lens of Source--Target Domain Proximity}
\label{sec:proximity-transferability}

\begin{figure}
    
    \includegraphics[width=0.99\linewidth]{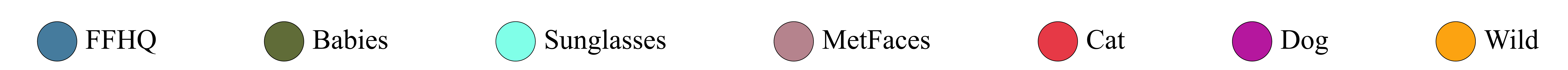} \\
    \begin{tabular}{ccc}
    \begin{minipage}{0.3\columnwidth}
    \hspace*{-5 mm}
     \includegraphics[width=1.0\linewidth]{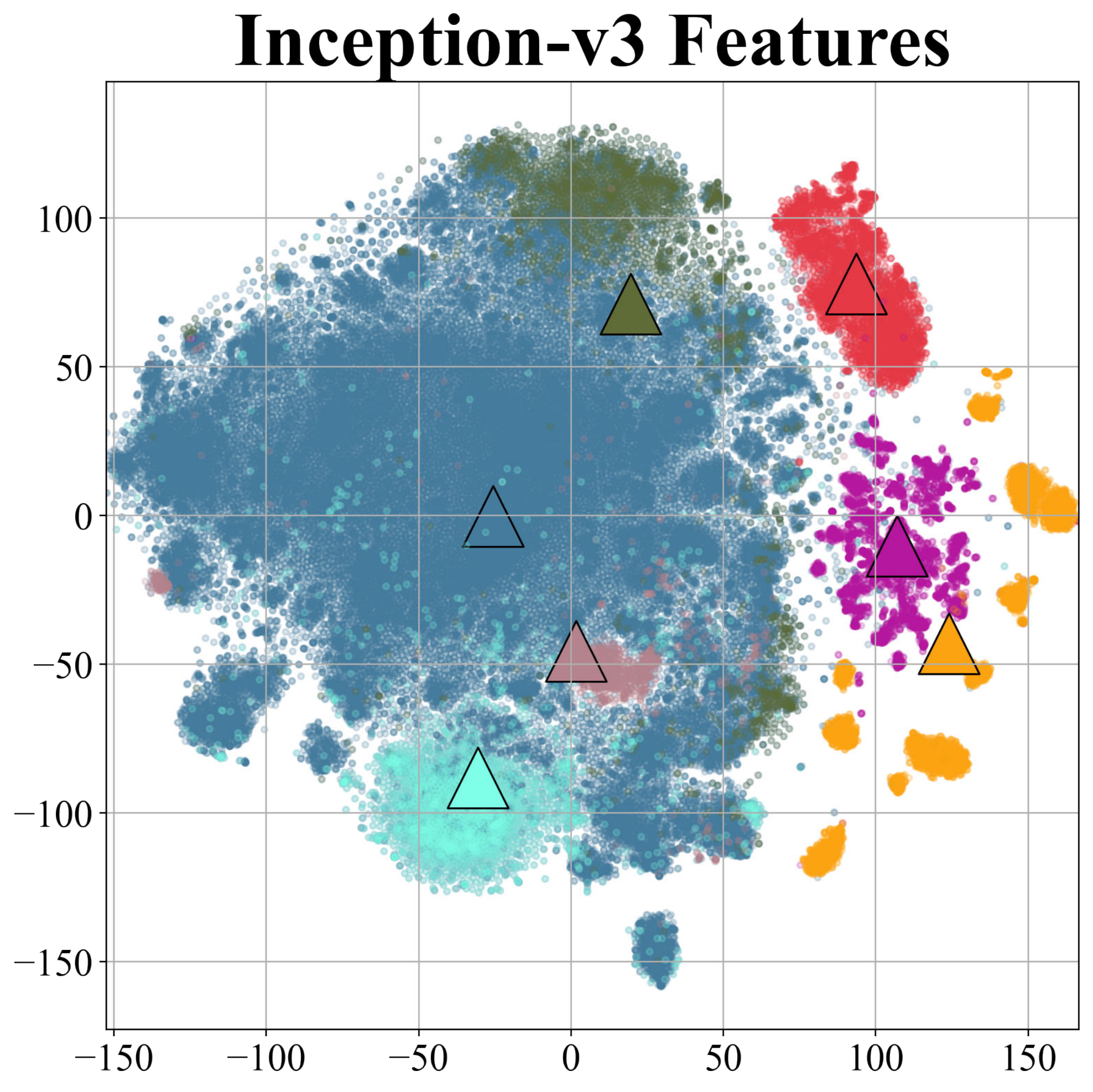}
     \end{minipage}
     
     & 
    
    \begin{minipage}{0.3\columnwidth}
    \hspace*{-5 mm}
     \includegraphics[width=1.0\linewidth]{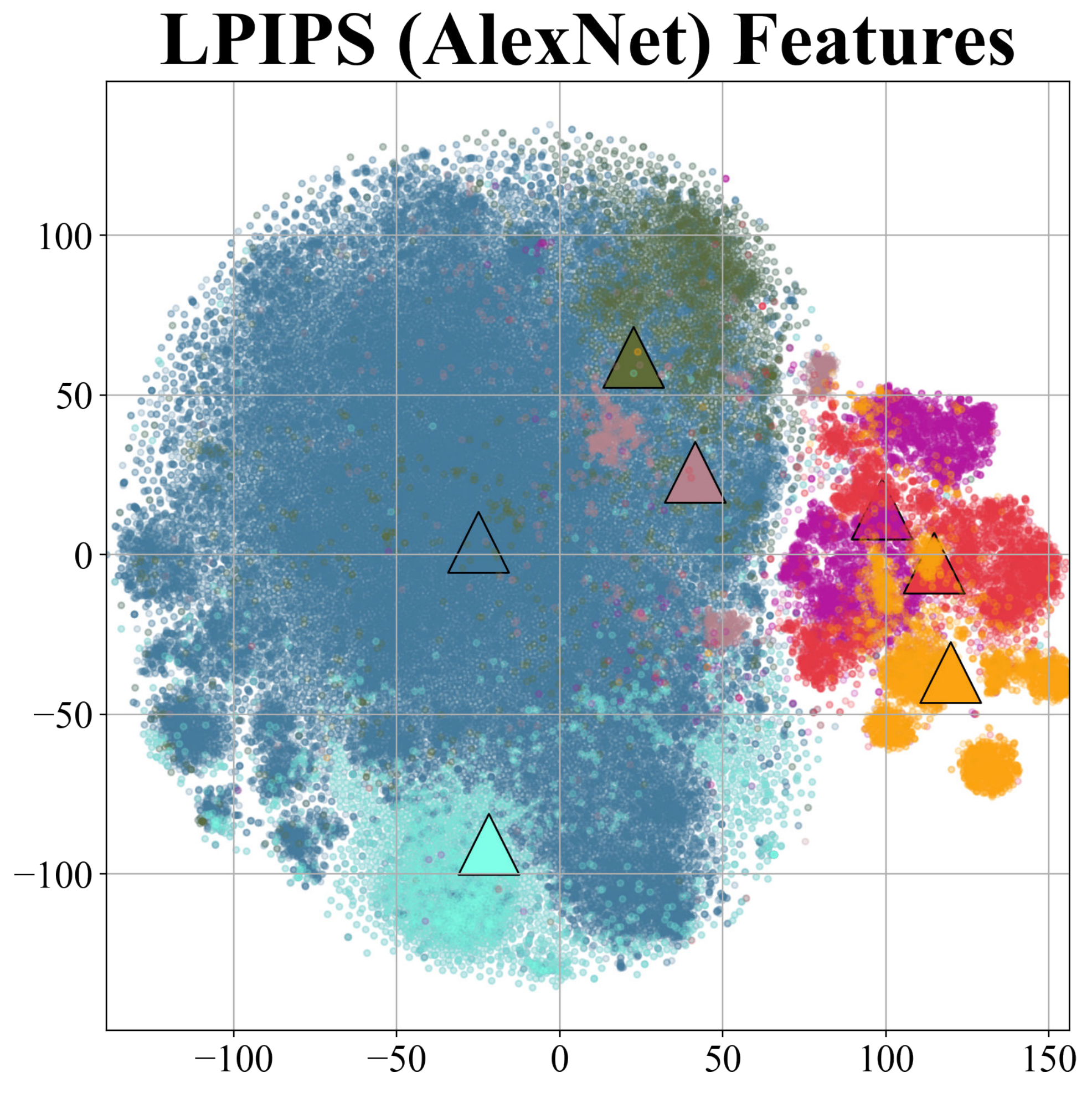}
     \end{minipage}
     
     & 
    
    \hspace*{-5 mm}
    \begin{minipage}{0.35\columnwidth}
    \begin{adjustbox}{width=1.0\columnwidth,center}
       \begin{tabular}{lccc}\toprule
        \textbf{Target Domain} &\textbf{Size} &\textbf{FID $\downarrow$} &\textbf{LPIPS $\downarrow$} \\ \toprule
        
        FFHQ \cite{karras2018styleGANv1} &70.0K &- &- \\ \midrule
        Babies \cite{ojha2021fig_cdc} &2.49K &147 &0.274 \\ \midrule
        Sunglasses \cite{ojha2021fig_cdc} &2.68K &108 &0.347 \\ \midrule
        MetFaces \cite{karras2020ADA} &1.33K &107 &0.358 \\ \midrule
        Cat \cite{choi2020starganv2} &5.15K &227 &0.479 \\ \midrule
        Dog \cite{choi2020starganv2} &4.74K &210 &0.442 \\ \midrule
        Wild \cite{choi2020starganv2} &4.74K &272 &0.484 \\ 

        \bottomrule
        \end{tabular}
    
    \end{adjustbox}
    \label{table:eta_set1}
    \end{minipage}
     \\
    \end{tabular}
  \vspace{-3mm}
  \caption{
\textit{Qualitative / Quantitative analysis of source-target domain proximity:}
We use FFHQ \cite{karras2020styleganv2} as the source domain.
We show 
source-target
domain proximity qualitatively by
visualizing Inception-v3 \textbf{(Left)} \cite{szegedy2016rethinking} and LPIPS \textbf{(Middle)} \cite{zhang2018lpips} -- using AlexNet \cite{krizhevsky2012alexnet} backbone -- features, 
and quantitatively using FID / LPIPS metrics \textbf{(Right)}. 
For feature visualization, we use t-SNE \cite{JMLR:v9:vandermaaten08a_tsne} and show centroids ($\bigtriangleup$) for all domains. 
FID / LPIPS is measured with respect to FFHQ. 
There are two important observations: 
\textcircled{\raisebox{-0.9pt}{1}} Common target domains used in existing FSIG works (Babies, Sunglasses, MetFaces) are notably proximal to the source domain (FFHQ). This can be observed from the feature visualization and verified by FID / LPIPS measurements.
\textcircled{\raisebox{-0.9pt}{2}} 
We clearly show using feature visualizations and FID / LPIPS measurements that additional setups -- Cat \cite{choi2020starganv2}, Dog \cite{choi2020starganv2} and Wild \cite{choi2020starganv2} -- represent target domains that are distant from the source domain (FFHQ).
We remark that large FID values in this analysis are reasonable due to the distance between the source (FFHQ) and 
different target domains 
as observed from centroid distance / feature variance.
The effect of limited sample size (target domains) for FID / LPIPS measurements are minimal and we include rich supportive studies in Supplementary.
Additional experiments and source/target setups in Supplementary to further support our analysis.
  }
\label{fig:proximity-visualization-measurements}
\vspace{-0.7cm}
\end{figure}

In this section, we revisit existing FSIG methods (10-shot) \cite{wang2018transferringGAN, mo2020freezeD, li2020fig_EWC, ojha2021fig_cdc, zhao2022dcl}
through the lens of source-target
domain proximity.
Specifically, we scrutinize the experimental setups of existing FSIG methods and 
observe 
that SOTA \cite{li2020fig_EWC, ojha2021fig_cdc, zhao2022dcl} largely focus on adapting to target domains that are (semantically) proximal to the source domain: 
Human Faces (FFHQ) $\rightarrow$ Baby Faces;
Human Faces (FFHQ) $\rightarrow$ Sunglasses;
Cars $\rightarrow$ Abandoned Cars; 
Church $\rightarrow$ Haunted Houses \cite{li2020fig_EWC, ojha2021fig_cdc, zhao2022dcl}.
This raises the question as to whether existing source-target domain setups sufficiently represent \textit{general} FSIG scenarios.
Particularly, real-world FSIG applications may not contain target domains that are always proximal to the source domain (\eg,: Human Faces (FFHQ) $\rightarrow$ Animal Faces).
Motivated by this, we conduct an in-depth qualitative and quantitative analysis on source-target
domain proximity where we introduce target domains that are distant from the source domain (Sec \ref{sub-sec:proximity-analysis}).
Our analysis uncovers an important finding:
\textbf{Under our additional setups where the assumption of close proximity between source and target domain is relaxed, existing SOTA FSIG methods \cite{li2020fig_EWC, ojha2021fig_cdc, zhao2022dcl}
which consider only source domain/source task in knowledge preserving
perform \textit{no better} than a baseline fine-tuning method.}
We show this is due to the strong focus of existing SOTA methods in preserving source domain knowledge, thereby not being able to adapt well to distant target domains (Sec \ref{sub-sec:proximity-sota-methods}) .

\subsection{Source--Target Domain Proximity Analysis}
\label{sub-sec:proximity-analysis}

\textbf{Introducing target domains with varying degrees of proximity to the source domain.}
In this section, we formally introduce
source-target
domain proximity with in-depth analysis to
scrutinize existing FSIG methods under different degrees of 
source-target
domain proximity.
Following prior FSIG works \cite{wang2018transferringGAN, mo2020freezeD, li2020fig_EWC, ojha2021fig_cdc, zhao2022dcl}, 
we use FFHQ \cite{karras2020styleganv2} as the source domain in this analysis. We remark that existing works largely consider different types of human faces as target domain (\ie: Babies \cite{ojha2021fig_cdc}, Sunglasses \cite{ojha2021fig_cdc}, MetFaces \cite{karras2020ADA}), 
To relax the close proximity assumption and study \textit{general} FSIG problems, we introduce more distant target domains namely Cat, Dog and Wild (from AFHQ \cite{choi2020starganv2}, consisting of 15,000 high-quality animal face images at 512 × 512 resolution) for our analysis.

\textbf{Characterizing source-target domain proximity.}
Given the wide success of deep neural network features in representing meaningful semantic concepts \cite{talebi2018learned, talebi2018nima, morozov2021on_ssl_gan_evaluation}, we visualize Inception-v3 \cite{szegedy2016rethinking} and LPIPS \cite{zhang2018lpips} features for source and target domains to qualitatively characterize domain proximity. 
Further, we use FID \cite{heusel2017FID} and LPIPS  distance to quantitatively characterize source-target domain proximity.
We remark that FID involves distribution estimation (first, second order moments) \cite{heusel2017FID} and LPIPS computes pairwise distances (learned embeddings) \cite{zhang2018lpips} between source / target domains.

\textbf{Analysis.}
Feature visualization and FID/ LPIPS measurement results are shown in Figure \ref{fig:proximity-visualization-measurements}.
Our results both qualitatively (columns 1, 2) and quantitatively (column 3) show that target domains used in existing works (Babies \cite{karras2020styleganv2}, Sunglasses \cite{karras2020styleganv2}, MetFaces \cite{karras2020ADA}) are notably proximal to the source domain (FFHQ), and our additionally introduced target domains (Dog, Cat and Wild \cite{choi2020starganv2}) are distant from the source domain thereby relaxing the close proximity assumption in existing FSIG works.

\subsection{FSIG methods under Relaxation of Close Domain Proximity Assumption}
\label{sub-sec:proximity-sota-methods}

Motivated by our analysis in Section \ref{sub-sec:proximity-analysis}, we investigate the performance of existing FSIG methods \cite{wang2018transferringGAN, mo2020freezeD, li2020fig_EWC, ojha2021fig_cdc, zhao2022dcl}
by relaxing the close proximity assumption between source and target domains. 
We investigate the performance of these FSIG methods across target domains of different proximity to the source domain,
which includes our additionally introduced target domains: Dog, Cat and Wild.
The FID results for FFHQ $\rightarrow$ Cat are: TGAN
(simple fine-tuning) 
\cite{wang2018transferringGAN}: 64.68, EWC \cite{li2020fig_EWC}: 74.61, CDC \cite{ojha2021fig_cdc}: 176.21, DCL \cite{zhao2022dcl}: 156.82. Full results can be found in Table \ref{table:fid_scores}.
\begin{wrapfigure}[25]{r}{0.43\textwidth}
\vspace{-0.205cm}
    \includegraphics[width=0.43\textwidth]{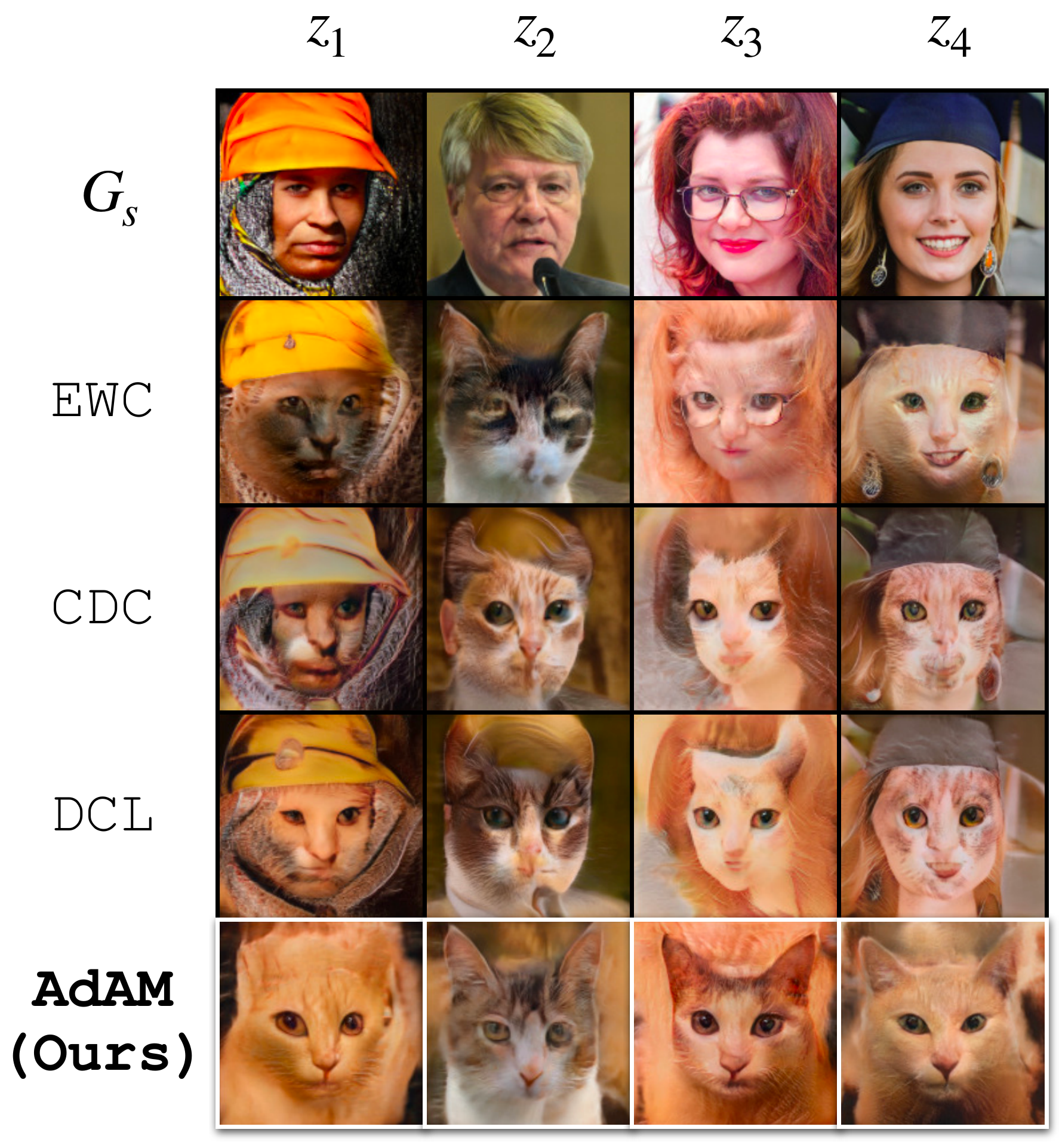}
\caption{
\small
$G_s$ is the source generator (FFHQ). Adapting from the source domain (FFHQ) to a distant target domain (Cat) using SOTA FSIG methods EWC \cite{li2020fig_EWC}, CDC \cite{ojha2021fig_cdc}, DCL \cite{zhao2022dcl} (rows 2, 3, 4) results in observable knowledge transfer that is not useful to the target domain. \ie: Source task knowledge such as \textit{Caps ($z_1, z_4$), Hair styles/color -- brown ($z_2$), red-hair ($z_3$), Eye glasses ($z_3$)} from FFHQ are transferred to Cats during adaptation which is not appropriate. 
Our method (last row) can alleviate these issues.
}
\label{fig:failure-cases-sota}
\end{wrapfigure}
We emphasize that our investigation uncovers an important finding: 
\textit{Under setups which the assumption of close proximity between source and target domain is relaxed (Dog, Cat, Wild), existing SOTA FSIG methods \cite{li2020fig_EWC, ojha2021fig_cdc, zhao2022dcl}}
perform \textit{no better} than a baseline method \cite{wang2018transferringGAN}. 
This can be consistently observed in Table \ref{table:fid_scores}.

This finding is critical as it exposes a serious drawback of SOTA FSIG methods \cite{li2020fig_EWC, ojha2021fig_cdc, zhao2022dcl} when close domain proximity (between source and target) assumption is relaxed.
We further analyse generated images from SOTA FSIG methods and observe that these methods are unable to adapt well to distant target domains due to \textit{only considering source domain / task in knowledge preservation.}
This can be clearly observed from Figure \ref{fig:failure-cases-sota}.
We remark that TGAN (simple baseline) \cite{wang2018transferringGAN} also suffers from severe mode collapse. 
Given that our investigation uncovers an important problem in SOTA FSIG methods, we tackle this problem in Sec \ref{section4}.
Figure \ref{fig:failure-cases-sota} (last row) shows a glimpse of our proposed method.

\section{Adaptation-Aware Kernel Modulation}
\label{section4}
We focus on this question: 
{\em ``Given a pretrained GAN on a source domain $\mathcal{D}_s$, and a few-samples from a target domain $\mathcal{D}_t$, which part of the source model's knowledge should be preserved, and which part should be updated, during the adaptation from $\mathcal{D}_s$ to $\mathcal{D}_t$?''}
In contrast to SOTA FSIG methods \cite{li2020fig_EWC, ojha2021fig_cdc, zhao2022dcl}, we propose an adaptation-aware FSIG  that also considers the target domain / adaptation task in deciding 
which part of the source model's knowledge to be preserved.
In a CNN, each {\em kernel} is responsible for a specific part of knowledge (\eg, pattern or texture). Similar behaviour is also observed for both generator \cite{bau2019gan} and discriminator \cite{bau2017network} in GANs.
Therefore, in this work, we make this knowledge preservation decision at the kernel level, \ie, \emph{\bfseries\boldmath  casting the knowledge preservation to a decision problem of whether a 
kernel is important when adapting from $\mathcal{D}_s$ to $\mathcal{D}_t$}.

Our FSIG algorithm has two
main steps: (i)  a lightweight {\em importance probing} step, and (ii)  {\em main  adaptation} step.
In the first step, \ie,  importance probing,  we adapt the model using a parameter-efficient design to the target domain for a limited number of iterations, 
and during this adaptation, we measure the importance of each individual kernel for the {\em target domain}.
The output of importance probing are decisions of importance / unimportance of individual kernels.
Then, in the second step, \ie, main adaptation, we preserve the knowledge of 
important kernels and update the knowledge of 
unimportant kernels.
The overview of the proposed system is shown in Figure~\ref{fig:overview}
and the pseudocode is shown in Algorithm \ref{alg:main}.

\begin{algorithm}[t]
	\DontPrintSemicolon
	\SetAlgoLined

\KwRequire{Pre-trained GAN: $G_s$ and $D_s$, $iter_{probe}$, $iter_{adapt}$, 
threshold quantile $t$,
learning rate $\alpha$
}

\KwIP{}

Freeze all kernels $\{ \mathbf{W}_i \}_{i=1}^{N}$ in pre-trained networks $G_s$, and $D_s$\\

Randomly initialize a modulation matrix $\mathbf{M}_i$ for each kernel $\mathbf{W}_i$ \\

\For{$k = 0$, $k{+}{+}$, while $k < iter_{probe}$} 
{
Perform kernel modulation for all kernels using Eqn.\ref{eq:kml} to obtain modulated weights $\hat{\mathbf{W}}$\\
Update $\mathbf{M} \leftarrow \mathbf{M} - \alpha \nabla_{\mathbf{M}} \mathcal{L}(G(z);\hat{\mathbf{W}})$
\tcc{\small lightweight, i.e., $iter_{probe} << iter_{adapt}$}

}

Measure importance of each kernel $\mathbf{W}_i$ by computing FI for the corresponding $\mathbf{M}_i$ using Eqn.\ref{eq:FI_vectors}\\

Compute the index set $\mathcal{A}$ of important kernels using  quantile $t$ of FI values as threshold\\ 

\KwMA{}

\eIf{$j \in \mathcal{A}$}
{Initialize the kernel by $\mathbf{W}_j$ and freeze the kernel, randomly initialize $\mathbf{M}_j$}
{Initialize the kernel by $\mathbf{W}_j$}

\For{$k = 0$, $k{+}{+}$, while $k < iter_{adapt}$}
{
\eIf{$j \in \mathcal{A}$}
{Modulate kernel using Eqn.\ref{eq:kml} to obtain modulated weights $\hat{\mathbf{W}_j}$\\
Update $\mathbf{M}_j \leftarrow \mathbf{M}_j - \alpha \nabla_{\mathbf{M}_j} \mathcal{L}(G(z);\hat{\mathbf{W}})$}
{Update $\mathbf{W}_j \leftarrow \mathbf{W}_j - \alpha \nabla_{\mathbf{W}_j} \mathcal{L}(G(z);\hat{\mathbf{W}})$}
}

\caption{Few-Shot Image Generation via Adaptation-Aware Kernel Modulation (AdAM)}
\label{alg:main}
\end{algorithm}

{\bf Proposed Importance Probing for FSIG.}
Our intuition for the 
proposed importance probing is: {\em ``The source GAN kernels have different levels of importance for each target domain.''}
For example, different subsets of kernels could be important when adapting a pretrained GAN on FFHQ to Babies~\cite{ojha2021fig_cdc} compared to adapting the same pretrained GAN to Cat~\cite{choi2020starganv2}. Therefore, we aim for a  
knowledge preservation criterion that is target domain/adaptation-aware (Table~\ref{table:criteria}).
In order to achieve adaptation-awareness, we propose a light-weight importance probing algorithm which considers adaptation from source to target domain.
There are two important design considerations: probing under (i) extremely limited number of target data and (ii) low computation overhead.

As discussed, in this  {\em importance probing} step, we adapt the source model to the target domain for a limited number of iterations and with a few available target samples.
During this short adaptation step, we measure the importance of kernel for the adaptation task.
To measure the importance, we use Fisher information (FI) which gives the {\em informative knowledge} of that kernel in handling adaptation task ~\cite{achille2019task2vec}.
Then, based on FI measurement, we classify kernels into
important / unimportant.  
These kernel-level importance decisions are then used in the next step, \ie, main adaptation.

In the main adaptation step, we propose to apply {\em kernel modulation} to achieve restrained update for the important kernels, and {\em simple fine-tuning} for the unimportant kernels. As will be discussed, the modulation is rank-constrained and has restricted degree-of-freedom;  therefore, it  is capable to preserve knowledge of the important kernels.
On the other hand, simple fine-tuning has large degree-of-freedom for updating knowledge of the unimportant kernels.
Furthermore, the rank-constrained kernel modulation is parameter-efficient.
Therefore, we also apply this rank-constrained kernel modulation {\em in the probing step} to determine the importance of kernels.

{\bf Kernel Modulation.}
The kernel modulation is used in the main adaptation step to preserve knowledge of important kernels into the adapted model. Furthermore, it is also used in the probing step as a parameter-efficient technique to determine importance of kernels.
Specifically, we apply Kernel ModuLation (KML) which is proposed very recently 
\cite{milad2021revisit}.
In \cite{milad2021revisit}, KML is proposed for multimodal few-shot {\em classification} (FSC).
In particular, in \cite{milad2021revisit},  KML has been found to be effective 
for 
knowledge transfer between different {\em classification} tasks of different modes under few-shot constraint.
Therefore, in our work, we apply KML for knowledge transfer between different {\em generation} tasks of different domains 
under limited target domain samples.

Specifically, in each convolutional layer of a CNN, the \emph{i\textsuperscript{th}} kernel of that layer $\mathbf{W}_i \in \mathbb{R}^{c_{in} \times k\times k}$ is convolved with the input feature $\mathbf{X} \in \mathbb{R}^{c_{in} \times h \times w}$ to the layer to produce the \emph{i\textsuperscript{th}} output channel (feature map) $\mathbf{Y}_i \in \mathbb{R}^{ h' \times w'}$, \ie, $\mathbf{Y}_i = \mathbf{W}_i*\mathbf{X} + b_i$, where $b_i \in \mathbb{R}$ denotes the bias term. 
Then, KML \cite{milad2021revisit} modulates $\mathbf{W}_i$ by multiplying it with the modulation matrix $\mathbf{M}_i \in \mathbb{R}^{c_{in} \times k\times k}$ plus an all-ones matrix $\mathbf{J} \in \mathbb{R}^{c_{in} \times k\times k}$:
\begin{equation}
    \label{eq:kml}
    \hat{\mathbf{W}}_i = \mathbf{W}_{i} \odot (\mathbf{J} + \mathbf{M}_{i})
\end{equation}
where $\odot$ denotes 
Hadamard multiplication. 
In Eqn.~\ref{eq:kml}, using $\mathbf{J}$ allows to learn the modulation matrix in a residual format.
Therefore, the modulation weights are learned as perturbations around the pretrained kernels which helps to preserve source knowledge. 
The exact pretrained kernel can also be transferred to the target model if it is optimal.
There are some important differences between discriminative version of KML in \cite{milad2021revisit} and our version, please see Supplementary for details.

This baseline 
KML learns an individual modulation parameter for each coefficient of the kernel. Therefore, it could suffer from {\em parameter explosion} when using in recent GAN architectures (\eg, more than 58M parameters in StyleGAN-V2 \cite{karras2020styleganv2}
\footnote{\href{https://github.com/rosinality/stylegan2-pytorch}{\textcolor{RubineRed}{https://github.com/rosinality/stylegan2-pytorch}}})
.
To address this issue, instead of learning the modulation matrix, we learn a {\em  low-rank} version of it
\cite{milad2021revisit,simon2020modulating}.
More specifically, for a Conv layer within CNN, with a total number of $d_{out}$ kernels to be modulated, instead of learning $\mathbf{M}=\{\mathbf{M}_i\}_{i=1}^{d_{out}}$, we learn two proxy vectors $\mathbf{m}_1 \in \mathbb{R}^{d_{out}}$, and $\mathbf{m}_2 \in \mathbb{R}^{(c_{in} \times k \times k)}$, and construct the modulation matrix using the outer product of these vectors, \ie, $\mathbf{M}=\mathbf{m}_1 \otimes \mathbf{m}_2 $. 
Furthermore, 
as we are using KML for adaptable knowledge preservation, we {\em freeze} the base kernel $\mathbf{W}_i$ during adaptation. Therefore, trainable parameters are $\mathbf{m}_1, \mathbf{m}_2$.
This reduces the number of trainable parameters significantly,
and has better performance on restraining the update of important kernels  (see Supplementary).
As it will be discussed later, the value of $d_{out}$ equals to the total number of kernels in a layer ($c_{out}$) during probing, and for main adaptation, it is determined by the output of our probing method ($d_{out} \leq c_{out}$).

{\bf Importance Measurement.}
Recall our FSIG has two main steps: (i)  importance probing step (Lines 1-8 in Algorithm \ref{alg:main}), and (ii) main
adaptation step (Lines 9-21 in Algorithm \ref{alg:main}).
In probing, we also apply KML as a parameter-efficient technique to determine importance of individual kernels.
In particular, for probing, 
we apply KML to all kernels (in both generator and discriminator) to identify which of the {\em modulated} kernels are important for the adaptation task. To measure the importance of the modulated kernels, we apply Fisher information  (FI) to the modulation parameters.
In our FSIG setup, for a modulated GAN with parameters $\Theta$, 
Fisher information $\mathcal{F}$
can be computed as:
\begin{equation}
    \label{eq:fisher_information}
    \mathcal{F}(\Theta) = \mathbb{E} \big[- \frac{\partial^2}{\partial\mathbf{\Theta}^2} \mathcal{L}(x|\Theta) \big]
\end{equation}
where $\mathcal{L}(x|\Theta)$ is the binary cross-entropy loss  computed using the output of the discriminator, and $x$ includes few-shot target samples, and fake samples generated by GAN.
Then, FI for a modulation matrix $\mathcal{F}(\mathbf{M}_i)$ can be computed by averaging over FI values of parameters within that matrix. As we are using the low-rank estimation to construct the modulation matrix, we can estimate $\mathcal{F}(\mathbf{M}_i)$ by FI values of the proxy vectors.
In particular, considering the outer product in low-rank approximation, we have $\mathbf{M}_i = reshape([\mathbf{m}_1^i\mathbf{m}_2^{1}, \dots, \mathbf{m}_1^i\mathbf{m}_2^{(c_{in}\times k \times k)}])$, where $|\mathbf{m}_2| = c_{in}\times k \times k$.
Then we use the unweighted average of FI for parameters of $\mathbf{m}_1$ and $\mathbf{m}_2$, proportional to their occurrence frequency in calculation of 
$\mathbf{M}_i$, as an estimate of $\mathcal{F}(\mathbf{M}_i)$ (details in Supplementary):
\begin{equation}
\label{eq:FI_vectors}
   \hat{\mathcal{F}}(\mathbf{M}_i) =
    \mathcal{F}(\mathbf{m}_1^i) + \frac{1}{|\mathbf{m}_2|} \sum_{j=1}^{|\mathbf{m}_2|}\mathcal{F}(\mathbf{m}_2^j)
\end{equation}
After calculating $\hat{\mathcal{F}}(\mathbf{M}_i)$ for all modulation matrices in both generator and discriminator, we use the  $t$\% quantile of these values as a threshold (separately for generator and discriminator) to decide
whether modulation of a kernel is important or unimportant for adaptation to the target domain.
If the modulation of a kernel is determined to  be important (during probing), the kernel is modulated using KML during main adaptation step; otherwise, the kernel is updated using simple fine-tuning during main adaptation.
In all setups, we perform probing for 500 iterations.
We remark that in probing only modulation parameters  $\mathbf{m}_1, \mathbf{m}_2$  are trainable, and FI is only computed on them, therefore the probing is a very lightweight step and can be performed with minimal overhead (details in Supplementary).
The output of  probing step are the decisions to apply kernel modulation or simple fine-tuning on individual kernels.
Then, based on these decisions, the main adaptation is performed.
The proposed FSIG scheme is summarized in Algorithm \ref{alg:main}.

\section{Empirical Studies}
\label{section5}

 \begin{figure*}[t]
     \centering
     \includegraphics[width=\textwidth]{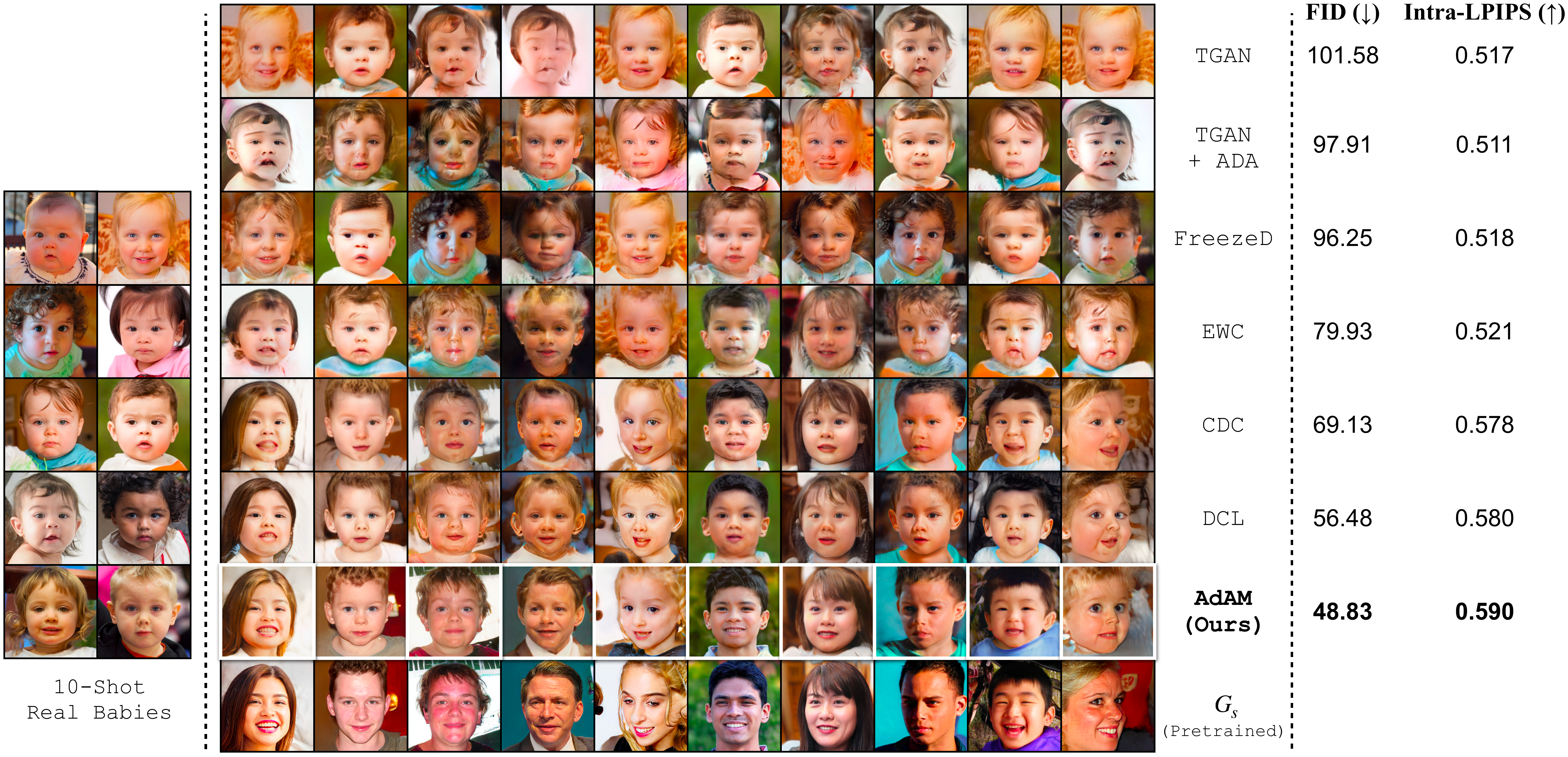}
     \includegraphics[width=\textwidth]{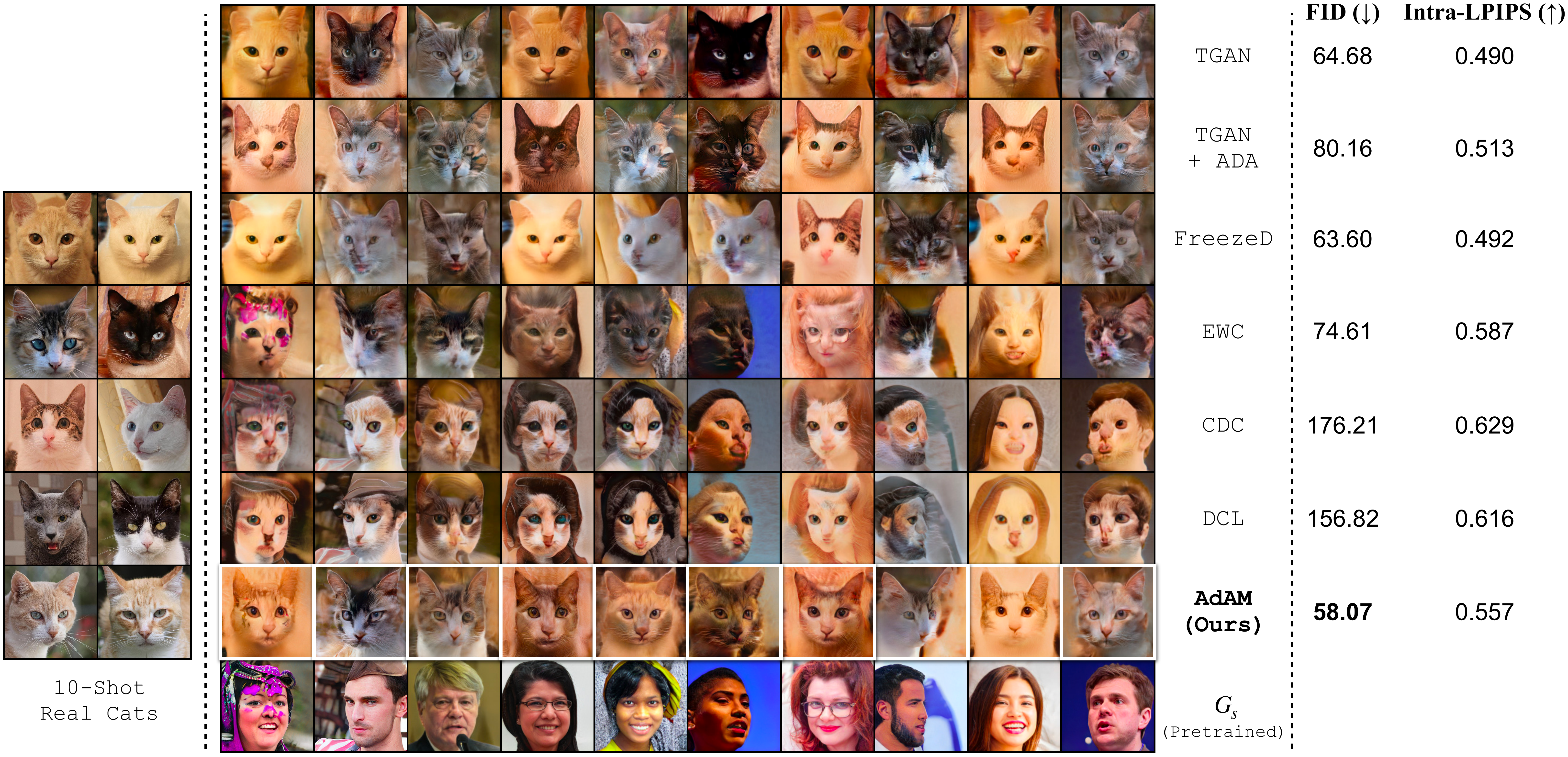}
     \caption{Qualitative and quantitative comparison of 10-shot image generation with different FSIG methods. Images of each column are from the same noise input. 
     \textbf{Left}: 10 real target images for few-shot adaptation.
     \textbf{Middle, Right}: For target domain with
     close proximity (\eg Babies, top), our method can generate high quality images with more refined details and diversity knowledge, achieving best FID and Intra-LPIPS socre.
     For target domain which is distant (\eg, Cat, bottom), TGAN/FreezeD overfit to the 10-shot samples and others fail. In contrast, our method preserves meaningful semantic features at different levels (\eg, posture and color) from source, achieving a good trade off between quality and diversity. In particular, our Intra-LPIPS approaches that of EWC, while our generated images have much better quality qualitatively and quantitatively.
     }
     \label{fig4}
     \vspace{-0.65cm}
 \end{figure*}

\subsection{Experiments / Results}
\begin{table}[t]  
    \centering
    \caption{
    {FSIG (10-shot) results: We report FID scores ($\downarrow$) of our proposed \textit{adaptation-aware} FSIG and compare with existing FSIG methods.
    We emphasize that Cat, Dog and Wild target domains are 
    additional experiments included in this work.
    (Sec \ref{sub-sec:proximity-analysis}).
    Our experiment results show two important findings: 
    \textbf{1)} Under setups which assumption of close proximity between source and target domains is relaxed (Cat, Dog, Wild), SOTA FSIG methods -- EWC, CDC, DCL --
    which consider only source domain in knowledge preserving
    perform \textit{no better} than a baseline fine-tuning method (TGAN).
    \textbf{2)} Our proposed adaptation-aware FSIG achieves SOTA performance in \textit{all} target domains due to 
    preserving source domain knowledge that is important for few-shot target domain adaptation.
    We generate 5,000 images using the adapted generator to evaluate FID on the whole target domain.
    We also report the corresponding KID, Intra-LPIPS and standard deviations in Supplementary.
    } 
    }
    \begin{adjustbox}{width=\columnwidth,center}
        \begin{tabular}{l| c c c c c c c c }
        \toprule
        \textbf{Target Domain}
         & \textbf{Babies} \cite{ojha2021fig_cdc}
         & \textbf{Sunglasses} \cite{ojha2021fig_cdc}
         & \textbf{MetFaces} \cite{karras2020ADA}
        & \textbf{AFHQ-Cat} \cite{choi2020starganv2}
        & \textbf{AFHQ-Dog} \cite{choi2020starganv2}
        & \textbf{AFHQ-Wild} \cite{choi2020starganv2}
         \\ 
        \hline
        TGAN \cite{wang2018transferringGAN} & 101.58 & 55.97 & 76.81 & 64.68 & 151.46 & 81.30    \\ \midrule
        TGAN+ADA \cite{karras2020ADA} & 97.91 & 53.64 & 75.82 & 80.16 & 162.63 & 81.55 \\ \midrule
        FreezeD \cite{mo2020freezeD} & 96.25 & 46.95 & 73.33 & 63.60 & 157.98 & 77.18 \\ \midrule
        EWC \cite{li2020fig_EWC} & 79.93 & 49.41 & 62.67 & 74.61 & 158.78 & 92.83   \\ \midrule
        CDC \cite{ojha2021fig_cdc} & 69.13 & 41.45 & 65.45 & 176.21 & 170.95 & 135.13  \\ \midrule
        DCL \cite{zhao2022dcl} & 56.48 & 37.66 & 62.35 & 156.82 & 171.42 & 115.93 \\ 
        \midrule
        \textbf{AdAM (Ours)} & \bm{$48.83$} & \bm{$28.03$} & \bm{$51.34$} & \bm{$58.07$} & \bm{$100.91$} & \bm{$36.87$}    \\
        \bottomrule
        \end{tabular}
        \vspace{-3 mm}
    \end{adjustbox}
\label{table:fid_scores}
\end{table}


\renewcommand{\arraystretch}{1}
\begin{figure}[h]
    \centering
    \includegraphics[width=\textwidth]{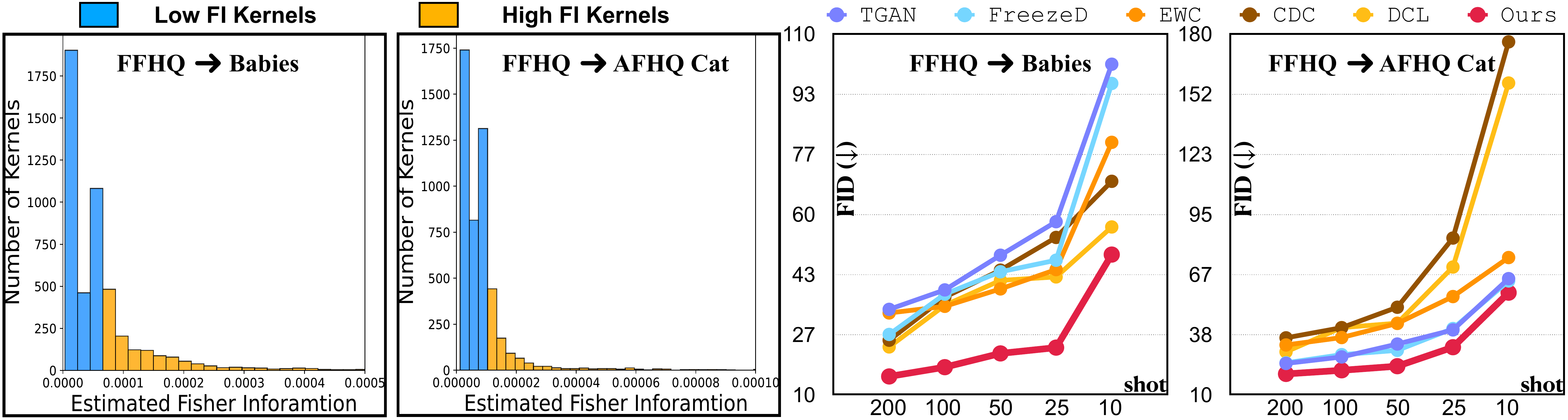}
    \begin{adjustbox}{width=0.85\textwidth}
        \begin{tabular}{l| c c c c c c c c }
        \toprule
        \textbf{Target Domain}
         & \textbf{Babies}
         & \textbf{Sunglasses}
         & \textbf{MetFaces}
        & \textbf{AFHQ-Cat}
        & \textbf{AFHQ-Dog}
        & \textbf{AFHQ-Wild}
         \\ 
        \hline
        \textbf{AdAM} (w/o probing) & 54.46 & 33.66  & 60.43 & 82.41 & 160.87 & 81.24    \\
        \midrule
        \textbf{AdAM} (Ours) & \bm{$48.83$} & \bm{$28.03$} & \bm{$51.34$} & \bm{$58.07$} & \bm{$100.91$} & \bm{$36.87$}    \\
        \bottomrule
        \end{tabular}
    \end{adjustbox}
    \caption{
    \textbf{(Top Left)} Our proposed IP identifies and preserves source kernels important (high FI) for target adaptation. \textbf{(Bottom)} FID score on different datasets. We validate the effectiveness of IP by modulating all kernels without IP. On the other hand, if we fine-tune all parameters without IP and modulation (TGAN), it suffers mode collapse (Table \ref{table:fid_scores} and Figure \ref{fig4}).     
    \textbf{(Top Right)} We evaluate the performance of different number of shots (10, 25, 50, 100, 200) on Babies and AFHQ-Cat. We show that our method consistently outperforms other FSIG methods in all setups. In Supplementary, we also show the generated images given different number of shots on more target domains. 
    }
    \label{table:ablation}
    \vspace{-0.65cm}
\end{figure}


\textbf{Experiment Details.} 
For fair comparison, we strictly follow prior works  \cite{wang2018transferringGAN, mo2020freezeD, li2020fig_EWC, ojha2021fig_cdc, zhao2022dcl} in the choice of GAN architecture, source-target adaptation setups and hyper-parameters. 
We use StyleGAN-V2 \cite{karras2020styleganv2} as the GAN architecture
and FFHQ as the source domain. 
Our experiments include setups with different source-target proximity: Babies/Sunglasses \cite{ojha2021fig_cdc}, MetFaces \cite{karras2020ADA} and Cat/Dog/Wild (AFHQ) \cite{choi2020starganv2} (See Sec. \ref{sec:proximity-transferability}).
Adaptation is performed with 256 x 256 resolution and batch size 4 on a Tesla V100 GPU. We apply importance probing and modulation on base kernels of both generator and discriminator. 
We focus on 10-shot target adaptation setup in the main paper.

\textbf{Qualitative Results.}
 We show generated images with our proposed AdAM along Baseline \cite{wang2018transferringGAN, mo2020freezeD} and SOTA FSIG methods \cite{li2020fig_EWC, ojha2021fig_cdc, zhao2022dcl} for two target domains, Babies and Cat with different degrees of proximity to FFHQ, before and after adaptation. 
 The results are shown in Figure \ref{fig4} top and bottom, respectively.
 By preserving source domain knowledge that is important for target domain, our proposed adaptation-aware FSIG method can generate substantially high quality images with high diversity for both Babies and Cat domains. We also include FID \cite{heusel2017FID} and Intra-LPIPS \cite{ojha2021fig_cdc} (for measuring diversity) to quantitatively show that our proposed method outperforms SOTA FSIG methods \cite{li2020fig_EWC, ojha2021fig_cdc, zhao2022dcl}.  
 We show more generated samples in Supplementary.

\textbf{Quantitative Results.}
We show complete FID ($\downarrow$) scores in Table \ref{table:fid_scores}.
Our proposed AdAM for FSIG achieves SOTA results across all target domains of varying proximity to the source (FFHQ).
We emphasize that it is achieved by preserving source domain knowledge that is important for target domain adaptation (Sec \ref{section4}). We also report Intra-LPIPS ($\uparrow$) as an indicator of diversity, as Figure \ref{fig4}.

\subsection{Analysis}

\textbf{Ablation study of Importance Probing.}
The goal of importance probing (denoted as ``IP'') is to identify kernels that are 
important for \textit{few-shot target adaptation} as shown in Figure \ref{table:ablation} (Top). 
To justify the effectiveness of our design choice, we perform an ablation study that discards the IP stage and regard all kernels as \textit{equally important} for target adaptation. 
Therefore, we simply modulate all kernels \textit{without any knowledge selection}. 
As one can observe from Figure \ref{table:ablation} (Bottom), knowledge selection plays a vital role in adaptation performance. Specifically, the significance of knowledge preservation is more evident when the target domains are distant from the source domain.

\textbf{Number of target samples (shots).}
The number of target domain training samples is an important factor that can impact the FSIG performance. In general, more target domain samples can allow better estimation of target distribution. 
We study the efficacy of our proposed method under different number of target domain samples.
The results are shown in Figure \ref{table:ablation}, and we show that our proposed adaptation-aware FSIG method consistently outperforms existing methods in all setups.

\section{Discussion}
\label{section6}

{\bf Conclusion.} Focusing on FSIG, we make two contributions.
First, we revisit current SOTA methods and their experiments. 
We discover that SOTA methods perform poorly in setups when source and target domains are more distant, as existing methods only consider source domain/task for knowledge preservation. Second, we propose a new FSIG method which is target/adaptation-aware (AdAM). Our proposed method outperforms previous work across all setups of different source-target domain proximity.
We include extended experiments and analysis in Supplementary.

{\bf Broader Impact.} Our work makes contribution to  generation of synthetic data in applications where sample collection is challenging, \eg, photos of rare animal species. This is an important contribution to many data-centric applications.
Furthermore, transfer learning of generative models using a few data sample enables data and computation-efficient model development. Our work has positive impact on environmental sustainability and reduction of greenhouse gas emission. 
While our work targets generative applications with limited-data, it parallely raises concerns regarding such methods being used for malicious purposes.
Given the recent success of forensic detectors \cite{Chandrasegaran_2022_ECCV,Wang_2020_CVPR,Chandrasegaran_2021_CVPR,frank2020leveraging}, we conduct a simple study using Color-Robust forensic detector proposed in \cite{Chandrasegaran_2022_ECCV} on our Babies and Cat datasets. 
We observe that the model achieves 99.8\% and 99.9\% average precision (AP) respectively showing that AdAM samples can be successfully detected.
We also remark that our work presents opportunities for improving knowledge transfer methods \cite{hinton_kd, chandrasegaran22_icml, heo2019comprehensive, evci2022head2toe} in a broader context.

{\bf Limitations.} While our experiments are extensive compared to previous works, in practical applications, there are many possible target domains which cannot be included in our experiments. However, as our method is target/adaptation aware, we believe our method can generalize better than existing SOTA which are target-agnostic.

\newpage
{
\small
\bibliographystyle{unsrt}
\bibliography{reference}
}

\newpage

\section*{Acknowledgment}
 {
 This project is partially supported by the grant RS-INSUR-00027.
 This work was also supported in part by the National Research Foundation, Singapore, under its AI Singapore Programmes (AISG) under Award AISG2-RP-2021-021 and Award AISG-100E2018-005; 
 and in part by the Singapore University of Technology and Design under Project PIE-SGP-AI-2018-01.
 We thank anonymous reviewers for their insightful comments.
 
 }

\section*{Supplementary Material}
 This Supplementary provides additional experiments, results, analysis and ablation studies to further support our contributions. The Supplementary materials are organized as follows:
  \begin{itemize}
    
     \item Section \textcolor{red}{\ref{sec_supp:ip}}: Proposed Importance Probing Algorithm : Details
     
     \begin{itemize}
         \item Section\textcolor{red}{\ref{sec_supp:ip_overhead}} Computational Overhead

         \item Section\textcolor{red}{\ref{sec_supp:kml_ops}} Illustration of Kernel Modulation Operations
         
         \item Section\textcolor{red}{\ref{sec_supp:ip_fim_approximation}} Fisher Information approximation using proxy vectors
     \end{itemize}

    \item Section \textcolor{red}{\ref{sec-supp:related_works}}: Discussion on Related Works
          
     \item Section \textcolor{red}{\ref{sec_supp:ablation}}: Ablation Studies and Additional Analysis on Importance Probing
     
     \item Section \textcolor{red}{\ref{sec-supp:extended_experiments}}: Extended Experiments and Results (and Visualizations)
     \begin{itemize}
        \item Section \textcolor{red}{\ref{subsec-supp:additional_source_target_domains}}: Additional Source / Target Domain Setups
        
        \item 
        {
        Section \textcolor{red}{\ref{subsec-supp:additional_gan_architectures}}: Additional GAN Architectures
        }
        
        \item 
        {
        Section \textcolor{red}{\ref{subsec-supp:class_saliency}}: Alternative characterization of importance measure
        }
        
        \item 
        {
        Section \textcolor{red}{\ref{subsec-supp:comparison_with_ada}}: Comparison with Adaptive Data Augmentation
        }
        
        \item 
        {
        Section \textcolor{red}{\ref{subsec-supp:ip_with_limited_samples}}: Importance probing with extremely limited number of samples
        }

     \end{itemize}

    \item 
    {
    Section \textcolor{red}{\ref{sec:rebuttal_visualizing_high_FI_kernels}}: Discussion: What form of visual information is encoded by high FI kernels?
    }
    
    \item Section \textcolor{red}{\ref{sec-supp:additional_main_paper_results}}:  Main Paper Experiments : Additional Results / Analysis
    
    \begin{itemize}
        \item Section \textcolor{red}{\ref{sec-supp:kid_intr_lpips}}: KID / Intra-LPIPS / Standard Deviation of Experiments
         
        \item Section  \textcolor{red}{\ref{sup-sec:extended_experiments_10_shot_results}}: 10-shot Adaptation Results
        
        \item Section \textcolor{red}{\ref{supp-sec:fid-analysis}}: FID measurements with limited target domain samples
        
    \end{itemize}

    \item 
    {
    Section \textcolor{red}{\ref{sec-supp:rebuttal_proximity_relaxation}}: Discussion: How much the proximity between the source and the target could be relaxed?
    }
    
    \item Section \textcolor{red}{\ref{sec-supp:checklist}}: Additional information for Checklist
    
    \begin{itemize}
        \item Section \textcolor{red}{\ref{sec-supp:societal_impact}}: Potential Societal Impact
        
        \item Section \textcolor{red}{\ref{sec-supp:amount_compute}}: Amount of Compute
        
    \end{itemize}
    
 \end{itemize}
 
 \textbf{Reproducibility.}
Project Page: \href{https://yunqing-me.github.io/AdAM/}{{\color{RubineRed}{https://yunqing-me.github.io/AdAM/}}}.
\newpage

\section{Proposed Importance Probing Algorithm: Details}
\label{sec_supp:ip}

\subsection{Computational Overhead}
\label{sec_supp:ip_overhead}

Our proposed Importance Probing (IP) algorithm to measure the importance of each individual kernel in the source GAN for the target-domain is lightweight. 
\textit{i.e.}: proposed importance probing only requires 8 minutes
compared to the adaptation step which requires $\approx$ 110 minutes
(Averaged over 3 runs for FFHQ $\rightarrow$ Cat adaptation experiment).
This is achieved using two design choices:
\begin{itemize}
    \item During IP, only modulation parameters are updated. Given that our modulation design is low-rank KML, the number of trainable parameters is significantly small compared to the actual source GAN. \textit{i.e.}: number of trainable parameters in our proposed IP is only 0.1M whereas the source GAN contains 30.0M trainable parameters.
    
    \item Our proposed IP is performed for limited number of iterations to measure the importance for the target domain. \textit{i.e.}: IP stage requires only 500 iterations to achieve a good performance for adaptation.
\end{itemize}

Complete details on number of trainable parameters and compute time for our proposed method and existing FSIG works are provided in Table \ref{tab:ip_computation_comparison}.
As one can observe, our proposed method (both IP and adaptation) is better than existing FSIG works in terms of trainable parameters and compute time.

\begin{table}
    \centering
    \caption{
    Comparison of training cost in terms of number of trainable parameters, training iterations and compute time for different FSIG methods. 
    FFHQ is the source domain and we show results for Babies (top) and Cat (bottom) target domains.
    One can clearly observe that our proposed IP is extremely lightweight and our KML based adaptation contains much less trainable parameters in the source GAN. 
    All results are measured in containerized environments using a single Tesla V100-PCIE (32 GB) GPU with batch size of 4. All reported results are averaged over 3 independent runs.
    }
        \begin{tabular}{c|cccc}
        \multicolumn{5}{c}{\textbf{FFHQ} $\rightarrow$ \textbf{Babies}}\\
         \toprule
         \textbf{Method} & \textbf{Stage} & \textbf{ \# trainable params (M)} &  \textbf{\# iteration} & \textbf{\# time} \\\hline
         TGAN \cite{wang2018transferringGAN} & Adaptation & 30.0 & 3000 & 110 mins \\ \midrule
         FreezeD \cite{mo2020freezeD} & Adaptation & 30.0 & 3000 & 110 mins \\ \midrule
         EWC \cite{li2020fig_EWC} & Adaptation & 30.0 & 3000 & 110 mins \\ \midrule
         CDC \cite{ojha2021fig_cdc} & Adaptation & 30.0 & 3000 & 120 mins\\ \midrule
         DCL \cite{zhao2022dcl} & Adaptation & 30.0 & 3000 & 120 mins \\
         \cline{1-5}
         \multirow{2}{*}{\textbf{AdAM (Ours)}} & IP & \textbf{0.105} & \textbf{500} & 8 mins\\ 
         \cline{2-5}
         & Adaptation & \textbf{18.9} & \textbf{1500}
         & 65min \\
         \bottomrule
        \end{tabular}
        
        \vspace{1em}
        \begin{tabular}{c|cccc}
        \multicolumn{5}{c}{\textbf{FFHQ} $\rightarrow$ \textbf{AFHQ-Cat}}\\
         \toprule
         \textbf{Method} & \textbf{Stage} & \textbf{ \# trainable params (M)} &  \textbf{\# iteration} & \textbf{\# time} \\\hline
         TGAN \cite{wang2018transferringGAN} & Adaptation & 30.0 & 6000 & 210 mins \\ \midrule
         FreezeD \cite{mo2020freezeD} & Adaptation & 30.0 & 6000 & 200 mins \\ \midrule
         EWC \cite{li2020fig_EWC} & Adaptation & 30.0 & 6000 & 220 mins \\ \midrule
         CDC \cite{ojha2021fig_cdc} & Adaptation & 30.0 & 6000 & 300 mins\\ \midrule
         DCL \cite{zhao2022dcl} & Adaptation & 30.0 & 6000 & 300 mins \\
         \cline{1-5}
         \multirow{2}{*}{\textbf{AdAM (Ours)}} & IP & \textbf{0.105} & \textbf{500} & 8 mins\\ 
         \cline{2-5}
         & Adaptation & \textbf{18.9} & \textbf{2500}
         & 110 mins \\
         \bottomrule
        \end{tabular}
    \label{tab:ip_computation_comparison}
\end{table}

\subsection{Kernel Modulation (KML) with rank-constrained operations}
\label{sec_supp:kml_ops}
\begin{figure}[t]
    \centering
        \includegraphics[width= \textwidth]{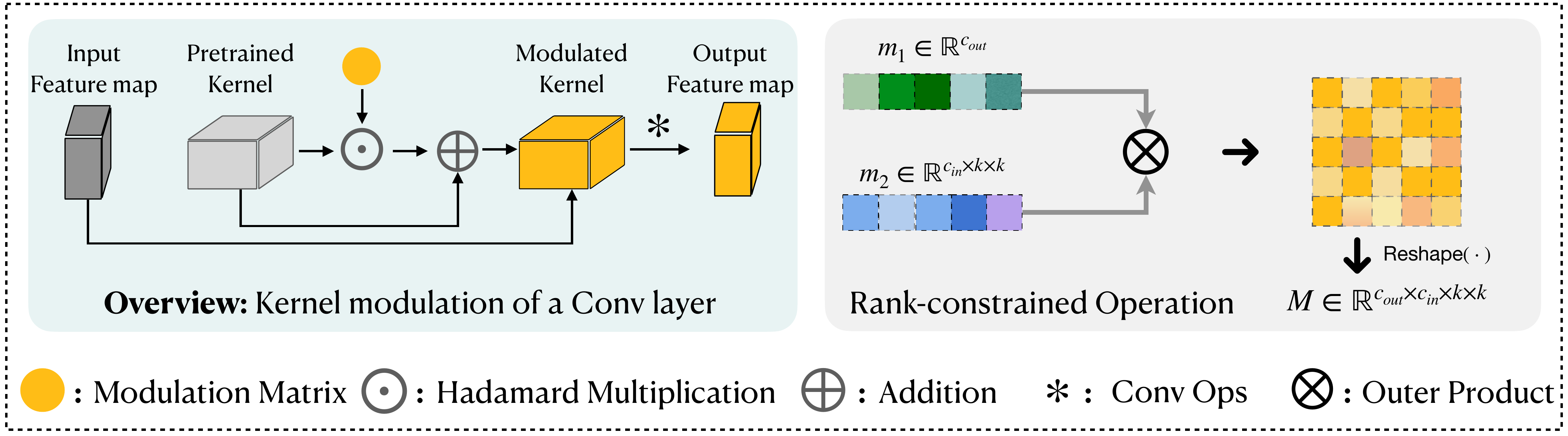}
    \caption{
    Illustration of Kernel Modulation operations. Here we use a convolutional kernel for instance. Similar operations are applied to the linear layer.
    }
    \label{fig:kml_ops}
\end{figure}

Here we show more details of KML, as supplement to the main paper, as Figure \ref{fig:kml_ops}.

\subsection{Fisher Information Approximation Using Proxy Vectors}
\label{sec_supp:ip_fim_approximation}

Recall in Sec.{\color{red} 4} of main paper, we consider low-rank approximation of modulation matrix using outer product of proxy vectors:  $\mathbf{M}_i = reshape([\mathbf{m}_1^{i}\mathbf{m}_2^1, \dots, \mathbf{m}_1^i\mathbf{m}_2^{c_{in} \times k \times k}])$, where $|\mathbf{m}_2|={(c_{in} \times k \times k)}$.
In order to calculate the FI of the modulation matrix, we start with the FI of each element in this matrix. Considering $m_{ij}=\mathbf{m}_1^i\mathbf{m}_2^j$, following equation can be derived by simple application of 
chain rule of 
differentiation:
\begin{equation}
    \frac{\partial\mathcal{L}}{\partial m_{ij}} = \frac{1}{2\mathbf{m}_2^j}\frac{\partial\mathcal{L}}{\partial \mathbf{m}_1^i} 
    + \frac{1}{2\mathbf{m}_1^i}\frac{\partial\mathcal{L}}{\partial \mathbf{m}_2^j}
\end{equation}
We use the square of the gradients to estimate the FI ~\cite{achille2019task2vec}. Therefore, the following equation can be obtained between the FI of these variables:

\begin{equation}
    \mathcal{F}(m_{ij}) = \frac{1}{{4\mathbf{m}_2^j}^2}\mathcal{F}(\mathbf{m}_1^i)
    + \frac{1}{{4\mathbf{m}_1^i}^2}\mathcal{F}(\mathbf{m}_2^j)
    + \frac{1}{2\mathbf{m}_1^i\mathbf{m}_2^j}\frac{\partial\mathcal{L}}{\partial \mathbf{m}_1^i}\frac{\partial\mathcal{L}}{\partial \mathbf{m}_2^j}
\end{equation}
Then, the FI of the modulation matrix $\mathbf{M}_i = [m_{i1}, m_{i2}, \dots] $, can be calculated as:
\begin{equation}
    \begin{split}
    \label{eqn:FI_full}
    \mathcal{F}(\mathbf{M}_i) & = \mathlarger{\mathlarger{\sum}}_{j=1}^{|\mathbf{m}_2|}\mathcal{F}(m_{ij}) \\ 
    & = \mathlarger{\mathlarger{\sum}}_{j=1}^{|\mathbf{m}_2|} (\frac{1}{{4\mathbf{m}_2^j}^2}\mathcal{F}(\mathbf{m}_1^i)
    + \frac{1}{{4\mathbf{m}_1^i}^2}\mathcal{F}(\mathbf{m}_2^j) + \frac{1}{2\mathbf{m}_1^i\mathbf{m}_2^j}\frac{\partial\mathcal{L}}{\partial \mathbf{m}_1^i}\frac{\partial\mathcal{L}}{\partial \mathbf{m}_2^j}) \\
    & = \mathcal{F}(\mathbf{m}_1^i) \mathlarger{\mathlarger{\sum}}_{j=1}^{|\mathbf{m}_2|}\frac{1}{{4\mathbf{m}_2^j}^2}
    + \frac{1}{{4\mathbf{m}_1^i}^2} \mathlarger{\mathlarger{\sum}}_{j=1}^{|\mathbf{m}_2|} \mathcal{F}(\mathbf{m}_2^j) \\ & + \frac{1}{2\mathbf{m}_1^i}\frac{\partial\mathcal{L}}{\partial \mathbf{m}_1^i} \mathlarger{\mathlarger{\sum}}_{j=1}^{|\mathbf{m}_2|} \frac{1}{\mathbf{m}_2^j} \frac{\partial\mathcal{L}}{\partial \mathbf{m}_2^j}
    \end{split}
\end{equation}
We empirically 
observed
that discarding (i) the cross-term (ii) the coefficients 
($\frac{1}{{4\mathbf{m}_2^j}^2}$,
$\frac{1}{{4\mathbf{m}_1^i}^2}$) 
in the importance of each kernel in Eqn.~\ref{eqn:FI_full}
results in a similar FID for the final adapted model. 
Therefore, 
the estimation can be simpler and more lightweight.
In particular, the following  (simpler) estimated version of $\mathcal{F}(\mathbf{M}_i)$ is used in our work:  
\begin{equation}
\label{eq:FI_vectors_supp}
   \hat{\mathcal{F}}(\mathbf{M}_i) =
    \mathcal{F}(\mathbf{m}_1^i) + \frac{1}{|\mathbf{m}_2|} \sum_{j=1}^{|\mathbf{m}_2|}\mathcal{F}(\mathbf{m}_2^j)
\end{equation}
Note that $\hat{\mathcal{F}}(\mathbf{M}_i)$ intuitively estimates the FI of the modulation matrix by a weighted average of its constructing parameters corresponding to their occurrence frequency
in calculation of 
$\mathbf{M}_i$.
We remark that in our implementation, for reporting all of the results in the main paper, and also the additional results in the supplementary, we have used this lightweight estimation Eqn.~\ref{eq:FI_vectors_supp} to calculate the importance of each kernel during importance probing.

\section{Discussion of Related Works}
\label{sec-supp:related_works}
In Sec.{\color{red}2} of the main paper, we discuss closely-related work of this paper that focuses on few-shot image generation (FSIG) under extremely limited data, i.e., 10 samples. Here, we review other related work.

\subsection{Image generation with less data}
Since the introduction of GANs \cite{goodfellow2014GAN}, there is a fair amount of work to focus on training of GANs with less data in recent literature, with efforts on introducing additional data augmentation methods \cite{karras2020ADA, tran2018distGAN}, regularization terms \cite{tseng2021regularizingGAN}, modifying GAN architectures \cite{zhao2020leveraging_icml_adafm}, and modification of filter kernels \cite{zhao2020leveraging_icml_adafm,cong2020gan_memory}.
Commonly, these works focus on setups with 
several thousands of images, i.e.: Flowers dataset \cite{nilsback2008oxford_flower} with 8,189 images in 
\cite{cong2020gan_memory},  10\% of ImageNet, or the entire AFHQ \cite{choi2020starganv2} dataset.
On the other hand, FSIG with extremely limited data (10 samples) poses unique challenges. In particular, as pointed out in \cite{ojha2021fig_cdc,zhao2022dcl}, severe mode collapse and loss in diversity are  critical challenges in FSIG that require special attention. 
We remark that in \cite{cong2020gan_memory}, a 
technique called AdaFM is introduced to update kernels. However, the underlying ideas and mechanism of AdaFM and our KML are quite different. 
AdaFM is inspired from style-transfer literature \cite{adain}, introduces independent scale and shift (scalar)  parameters to update individual channels of kernels to manipulate their styles. 
On the other hand, as discussed in the main paper, KML introduces a structural  
$\mathbf{J} + \mathbf{M}$, $\mathbf{M}=\mathbf{m}_1 \otimes \mathbf{m}_2 $, to update multiple kernels in a coordinated manner. 
In our experiment, we also test AdaFM in few-shot setups and compare its performance with KML.

\subsection{Discriminative kernel modulation}
As mentioned in the main paper, Kernel ModuLation (KML) is originally proposed in \cite{milad2021revisit} for adapting the model between different modes of few-shot classification (FSC) tasks.
However, due to some differences between the multimodal meta-learner in \cite{milad2021revisit}, and our transfer learning-based scheme, there are important differences in design choices when applying KML to our problem.
\textbf{\em First}, in contrast to FSC work \cite{milad2021revisit} which follows a {\em discriminative learning} setup, we aim to address a problem in a {\em generative learning} setup. 
\textbf{\em Second}, in FSC setup, the modulation parameters are generated during adaptation to target task with a pretrained modulation network trained on tens of thousands of few-shot tasks. 
So the modulation parameters are not directly learned for a target few-shot task.
In contrast, in our setup, the base kernel is frozen during the adaptation, and we directly learn the modulation parameters for a target domain/task using a very limited number of samples (e.g., 10-shot). 
\textbf{\em Finally}, in FSC, usually source and target tasks follow a same task distribution $p(\mathcal{T})$. In fact, in implementation, even though the classes are disjoint between source and target tasks, all of them are constructed using the data from the same domain (e.g., miniImageNet~\cite{vinyals2016matching}). However, in our setup, the source and target tasks/domains distributions could be very different (e.g., Human Faces (FFHQ) $\rightarrow$ Cats).

\section{Ablation Studies and Additional Analysis on Importance Probing}
\label{sec_supp:ablation}

In this section, we conduct extensive ablation studies to show the significance of our proposed method for FSIG. Similar to main paper analysis, we use FFHQ \cite{karras2020styleganv2} as the source domain, and use Babies and Cat \cite{choi2020starganv2} as target domains. 
The different approaches in the study are as follows:

\begin{itemize}
  \item TGAN \cite{wang2018transferringGAN}: The source GAN models pretrained on FFHQ are updated using {\em simple fine-tuning} with the 10 shot target samples.
  \item EWC \cite{li2020fig_EWC}: 
  Following \cite{li2020fig_EWC}, a L2 regularization is applied to all model weights to augment simple fine-tuning. The regularization is scaled by the importance of individual model weights as determined by the 
  FI of the model weights based on the {\em source} models.
  \item EWC + IP: We apply our probing idea on top of EWC. In the probing step, original EWC as discussed above is used but with a small number of iterations. At the end of probing, FI of the model weights based on the {\em updated} models is computed.
  Then, during main adaptation, this {\em target-aware} FI is used to scale the L2 regularization. In other words, EWC + IP is a target-aware version of EWC in \cite{li2020fig_EWC} using our probing idea.
  \item AdaFM \cite{cong2020gan_memory}: AdaFM modulation is applied to all kernels. 
  \item AdaFM + IP: We apply our probing idea on top of AdaFM. In the probing step, original AdaFM as discussed above is used but with a small number of iterations. At the end of probing, FI of AdaFM parameters is computed, and kernels are classified as important/unimportant using the same 75\% quantile threshold as in our work. 
  Then, during main adaptation, the important kernels are updated via AdaFM, and the unimportant kernels are updated via simple fine tuning. In other words, AdaFM + IP is a target-aware version of AdaFM using our probing idea.
  \item Ours w/o IP (\textit{i.e.} main adaptation only): KML modulation is applied to all kernels.
  \item Ours w/ Freeze: We apply our probing idea as discussed in the main paper, \textit{i.e.}, with KML applied to all kernels but adaptation with a small number of iterations.
  At the end of probing, FI of KML parameters is computed, and kernels are classified as important/unimportant using the same 75\% quantile threshold as in our work. 
  Then, during main adaptation, the important kernels are {\em frozen}, and the unimportant kernels are updated via simple fine tuning. In other words, this is similar to our proposed method except that kernel freezing is used in main adaptation instead of KML for important kernels.
  \item Ours w/ KML (\textit{i.e.} our main proposed method): 
  This is the method proposed in the main paper. 
  We apply our probing idea as discussed in the main paper, \textit{i.e.}, with KML applied to all kernels but adaptation with a small number of iterations.
  At the end of probing, FI of KML parameters is computed, and kernels are classified as important/unimportant using 75\% quantile threshold. 
  Then, during main adaptation, the important kernels are modulated using KML, and the unimportant kernels are updated via simple fine tuning. 
\end{itemize}

\begin{figure}[h]
    \centering
        \includegraphics[width= \textwidth]{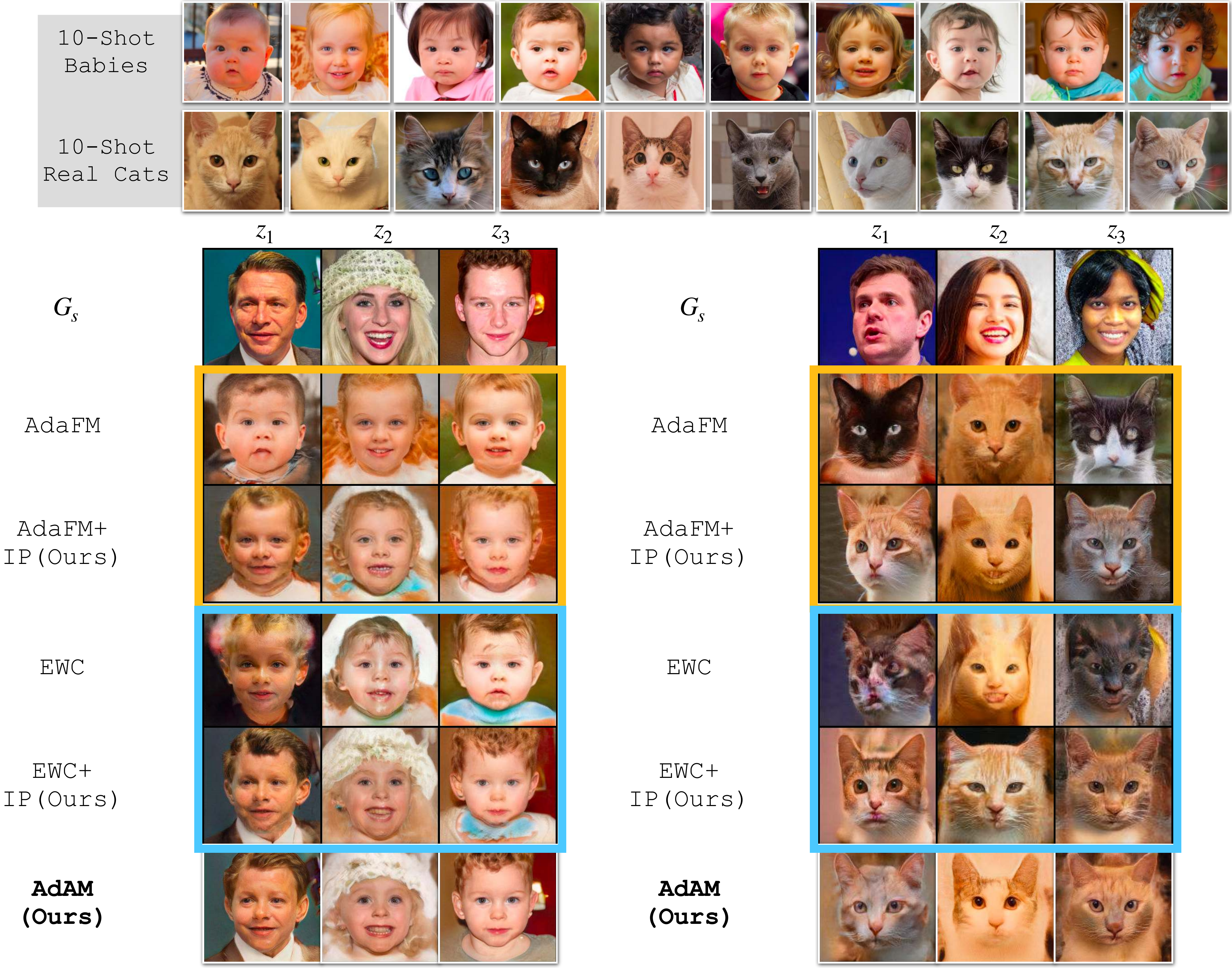}
    \caption{
    $G_s$ is the source generator (FFHQ).
    We show results for FFHQ $\rightarrow$ Babies (left) and FFHQ $\rightarrow$ Cat (right), similar to the main paper. 
    Applying our idea of importance probing to EWC \cite{li2020fig_EWC}, AdaFM \cite{cong2020gan_memory}, we observe better quality in FSIG. 
    This shows that our proposed idea on importance probing is principally a suitable approach to improve FSIG.
    One can also observe that images generated by our proposed method (with KML) has good quality compared to other methods. This is quantitatively confirmed in Table \ref{table-supp:ablation_results} .
    }
    \label{fig:ablation_ip}
\end{figure}

\textbf{Qualitative Results.}
We show generated images corresponding to all approaches discussed above in Figure \ref{fig:ablation_ip}. These results show that our proposed idea on importance probing is principally a suitable approach to improve FSIG by identifying kernels important for target domain adaptation. Figure
\ref{fig:ablation_ip} also shows that our proposed method can generate images with better quality.

\textbf{Quantitative results.}
We show FID / LPIPS results in Table 
\ref{table-supp:ablation_results}.
These results show that our proposed IP is principally a suitable approach for FSIG. 
This can be clearly observed when applying IP to EWC \cite{li2020fig_EWC} and AdaFM \cite{cong2020gan_memory}.
We remark that probing with KML (ours AdAM) is computationally much efficient compared to probing with EWC and AdaFM due to less number of trainable parameters.
Overall, we quantitatively show that our proposed method outperforms existing FSIG methods with IP, thereby generating images with a good balance between quality (FID $\downarrow$) and diversity (Intra-LPIPS $\uparrow$).
We also empirically observe that methods performing IP at kernel level (Ours w/ KML, AdaFM + IP) perform better than method performing IP at parameter level (EWC + IP).

\renewcommand{\arraystretch}{1}
\begin{table}[h]  
    \caption{
    Ablation studies for IP: 
    FFHQ \cite{karras2020styleganv2} is the source domain.
    We use Babies and Cats \cite{choi2020starganv2} as target domains. 
    We show FID (left) and Intra-LPIPS (right) results.
    For each method, best FID and LPIPS results are shown in \textbf{bold}.
    IP is performed for 500 iterations (where relevant).
    These results show that our proposed IP is principally a suitable approach for FSIG. This can be clearly observed when applying IP to EWC \cite{li2020fig_EWC} (EWC+IP) and AdaFM \cite{cong2020gan_memory} (AdaFM+IP).
    We also observe that methods performing IP at kernel level (Ours w/ KML, AdaFM + IP) perform better than method performing IP at parameter level (EWC + IP).
    Overall, we quantitatively show that our proposed method outperforms all existing FSIG methods with IP, thereby generating images with high quality (FID) and diversity (Intra-LPIPS).
    }
    \centering
   \begin{minipage}[t]{0.45\textwidth}
   \centering
        \begin{tabular}{l| c c }
        \toprule
        \textbf{Target Domain}
         & \textbf{Babies}
         & \textbf{Cat}
         \\ 
         & \multicolumn{2}{c}{FID ($\downarrow$)}  \\
                \hline
        TGAN \cite{wang2018transferringGAN} & 101.58 & 64.68 \\
        \hline
        EWC \cite{li2020fig_EWC}  & 79.93 & 74.61 \\
        EWC + [IP (Ours)] & \textbf{{70.80}} & \textbf{66.35} \\ \hline
        AdaFM \cite{cong2020gan_memory} & 62.90  & 64.44 \\
        AdaFM + [IP (Ours)] & \textbf{55.64} & \textbf{60.04}  \\ \hline
        Ours w/o IP & 54.46 & 82.41 \\
        Ours w/ Freeze [w/ IP] & 50.81 & 61.60 \\
        \textbf{AdAM} (w/ KML [w/ IP]) & \textbf{48.83} & \textbf{58.07} \\
        \bottomrule
        \end{tabular}
    \label{table:ablation-fid-supp}
    \end{minipage}
    \hspace{5 mm}
   \begin{minipage}[t]{0.45\textwidth}
   \centering
        \begin{tabular}{l| c c }
        \toprule
        \textbf{Target Domain}
         & \textbf{Babies}
         & \textbf{Cat}
         \\ 
         & \multicolumn{2}{c}{Intra-LPIPS ($\uparrow$)}   \\
        \hline
        TGAN \cite{wang2018transferringGAN} & 0.517 & 0.490 \\\hline
        EWC \cite{li2020fig_EWC} & 0.521 & \textbf{0.587} \\
        EWC + [IP (Ours)] & \textbf{0.625}  & 0.540 \\ \hline
        AdaFM \cite{cong2020gan_memory} & 0.568 & 0.525 \\
        AdaFM + [IP (Ours)] & \textbf{0.577} & \textbf{0.540}\\ \hline
        Ours w/o IP & \textbf{0.613} & 0.522 \\
        Ours w/ Freeze [w/ IP] & 0.581 & 0.559 \\
        \textbf{AdAM} (w/ KML [w/ IP]) & 0.590 & {0.557} \\
        \bottomrule
        \end{tabular}
    \label{table-supp:ablation_results}
    \end{minipage}
\end{table}


\section{Extended Experiments and Results}
\label{sec-supp:extended_experiments}

In this section, we conduct additional experiments to further support our findings and contributions. 

\subsection{Additional source / target domains}
\label{subsec-supp:additional_source_target_domains}
\begin{figure}

    \includegraphics[width=0.99\linewidth]{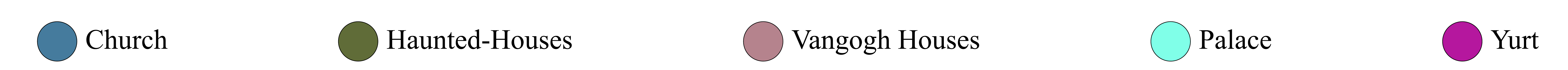} \\
    \vspace{-3 mm}
    \begin{tabular}{cc}
    
    \begin{minipage}{0.45\columnwidth}
     \includegraphics[width=1.0\linewidth]{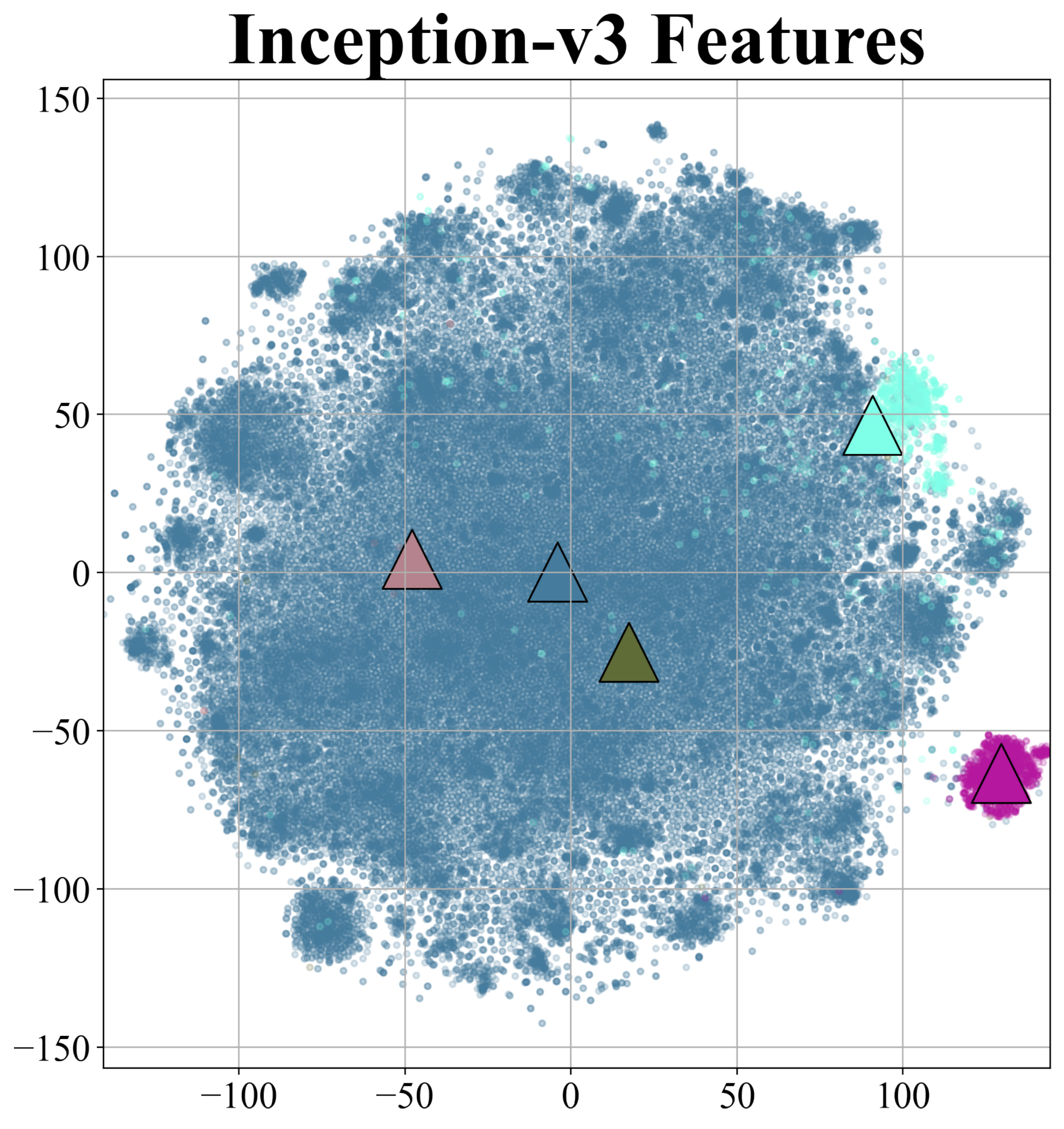}
     \end{minipage}
     
     & 

    \begin{minipage}{0.45\columnwidth}
     \includegraphics[width=1.0\linewidth]{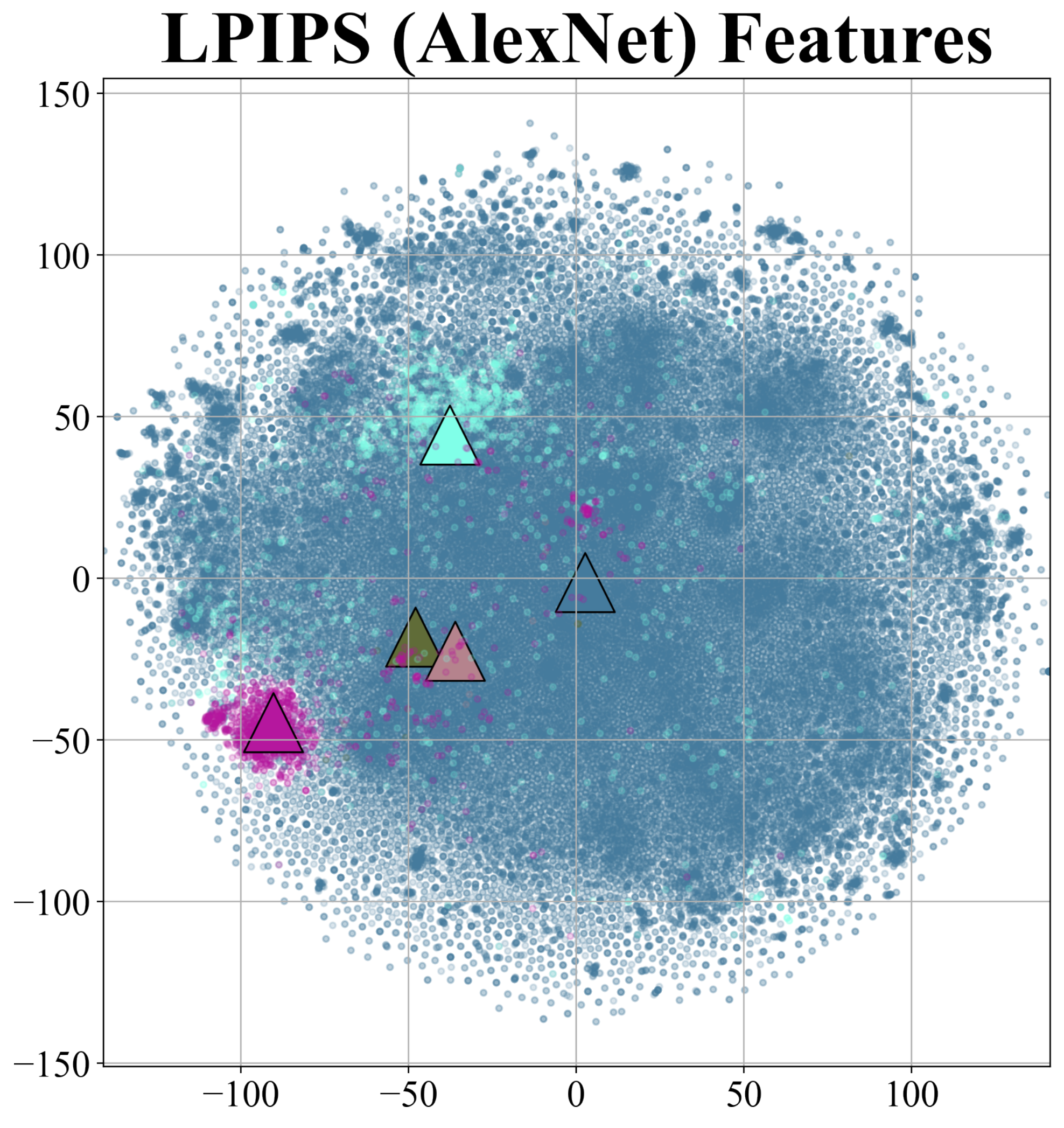}
     \end{minipage}
     
     \\
    \end{tabular}
  \caption{
\textit{Source-target domain proximity Visualization:}
We use Church as the source domain following \cite{ojha2021fig_cdc}.
We show 
source-target
domain proximity by
visualizing Inception-v3 (Left) \cite{szegedy2016rethinking} and LPIPS (Middle) \cite{zhang2018lpips} --using AlexNet \cite{krizhevsky2012alexnet} backbone-- features, 
and quantitatively using FID / LPIPS metrics (Right). 
For feature visualization, we use t-SNE \cite{JMLR:v9:vandermaaten08a_tsne} and show centroids ($\bigtriangleup$) for all domains. 
FID / LPIPS is measured with respect to FFHQ. 
There are 2 important observations: 
\textcircled{\raisebox{-0.9pt}{1}} Common target domains used in existing FSIG works (Haunted Houses, Van Gogh Houses) are notably proximal to the source domain (Church). This can be observed from the feature visualization and verified by FID / LPIPS measurements.
\textcircled{\raisebox{-0.9pt}{2}} 
We clearly show using feature visualizations and FID / LPIPS measurements that additional setups -- Palace \cite{deng2009imagenet} and Yurt \cite{deng2009imagenet} -- represent target domains that are distant from the source domain (Church).
We remark that due to availability of only 10-shot samples in the target domain, FID / LPIPS are not measured in these setups.
}
\label{fig-supp:proximity-visualization}
\end{figure}

Following \cite{ojha2021fig_cdc}, we conduct extended experiments using Church as the source domain. 
\cite{ojha2021fig_cdc} uses Haunted houses and Van Gogh Houses as target domains.
Similar to Sec.{\color{red}3} in the main paper, our analysis confirms that these target domains are closer to the source domain (Church). 
We additionally include palace and yurt as target domains to relax the close proximity assumption.
Proximity visualization is shown in Figure \ref{fig-supp:proximity-visualization}.

\textbf{Experiment Details.} 
For fair comparison, we strictly follow prior works  \cite{wang2018transferringGAN, mo2020freezeD, li2020fig_EWC, ojha2021fig_cdc, zhao2022dcl} in the choice of GAN architecture, source-target adaptation setups and hyper-parameters. 
We use StyleGAN-V2 \cite{karras2020styleganv2} as the GAN architecture
and FFHQ as the source domain. 
We use 256 x 256 resolution for adaptation. 
Adaptation is performed with batch size 4 on a single Tesla V100 GPU. We apply importance probing and modulation on base kernels of both generator and discriminator. 
We focus on 10-shot target adaptation setup.

\textbf{Results.}
Given that the target domain only contains 10 real images, following \cite{ojha2021fig_cdc}, we show the quality of FSIG for 10-shot adaption.
Qualitative analysis is shown in Figure \ref{fig:supp_failure_palace}. 
As one can observe, SOTA FSIG methods \cite{li2020fig_EWC, ojha2021fig_cdc, zhao2022dcl} are unable to adapt well to distant target domain (palace) due to \textit{due to only considering source domain / task in knowledge preservation.}
We remark that TGAN \cite{wang2018transferringGAN} suffers severe mode collapse.
We clearly show that our proposed adaptation-aware FSIG method outperforms existing FSIG works.

\begin{figure}[h]
    \centering
    \includegraphics[width=\textwidth]{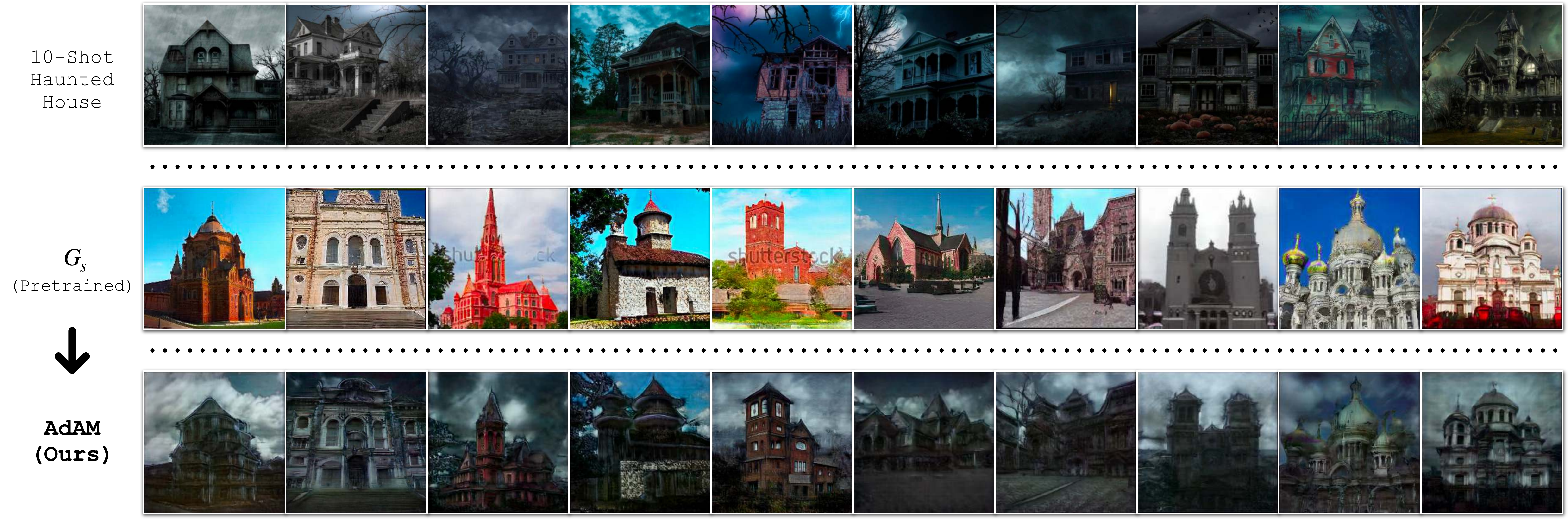}
    \caption{Church $\rightarrow$ Haunted House}
    \label{fig:supp_haunted}
\end{figure}

\begin{figure}[h]
    \centering
    \includegraphics[width=\textwidth]{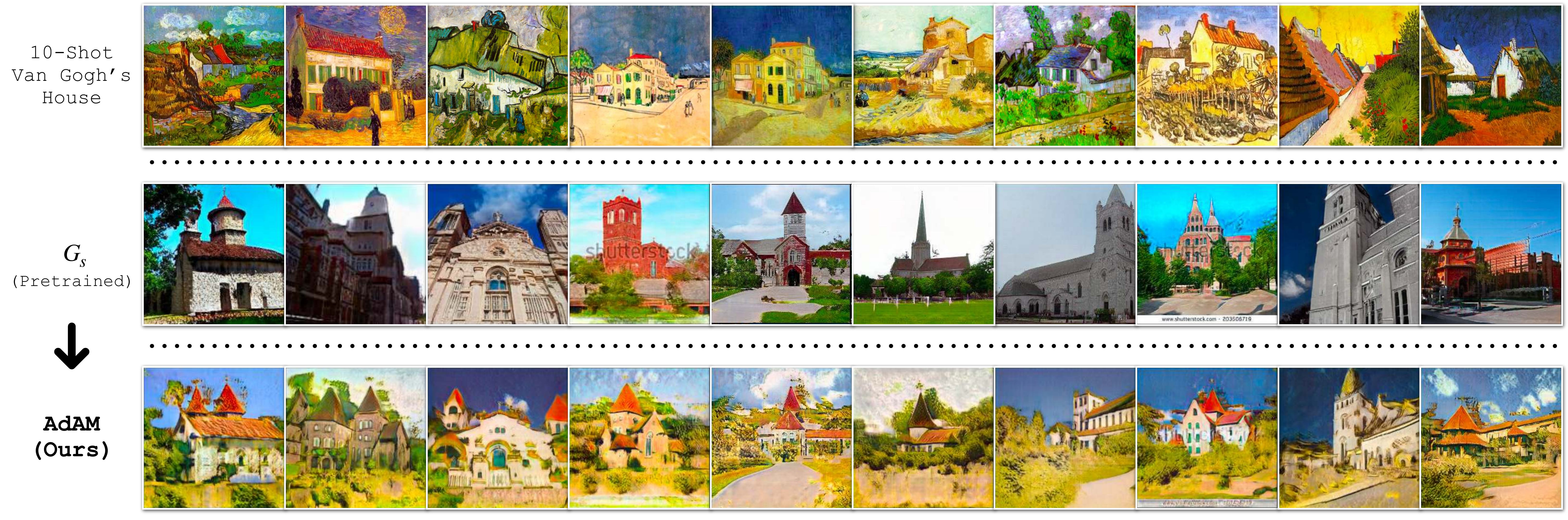}
    \caption{Church $\rightarrow$ Van Gogh's House}
    \label{fig:supp_van}
\end{figure}

\begin{figure}[h]
    \centering
    \includegraphics[width=\textwidth]{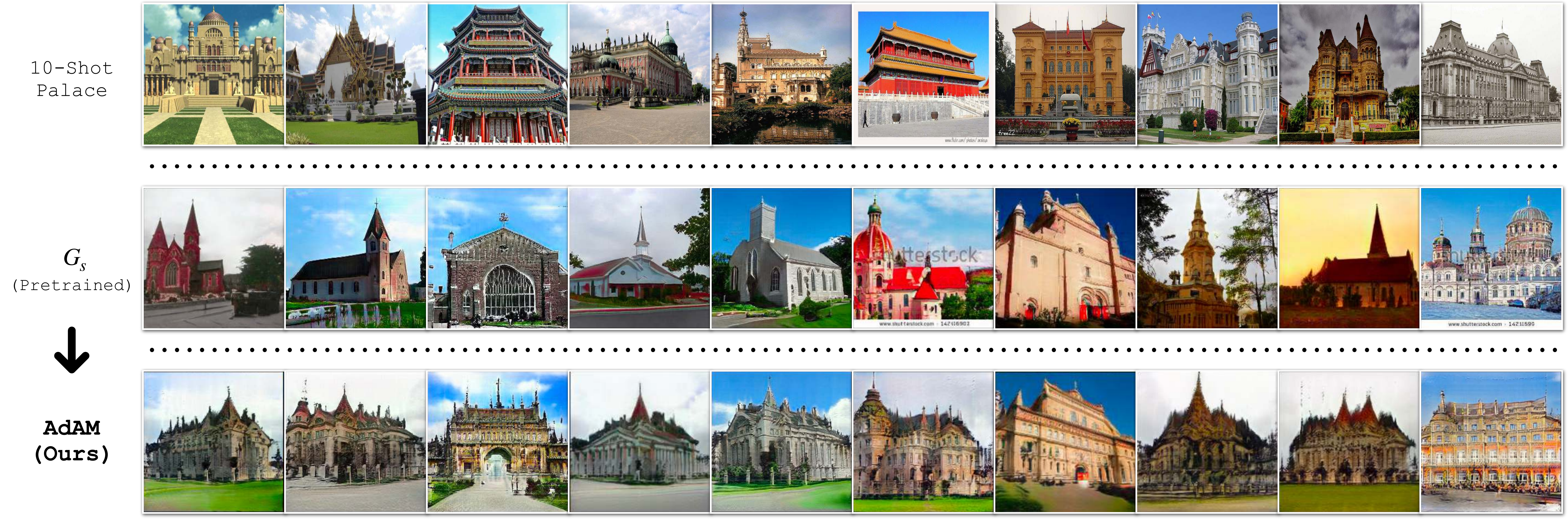}
    \caption{Church $\rightarrow$ Palace (distant domain)}
    \label{fig:supp_palace}
\end{figure}

\begin{figure}[h]
    \centering
    \includegraphics[width=\textwidth]{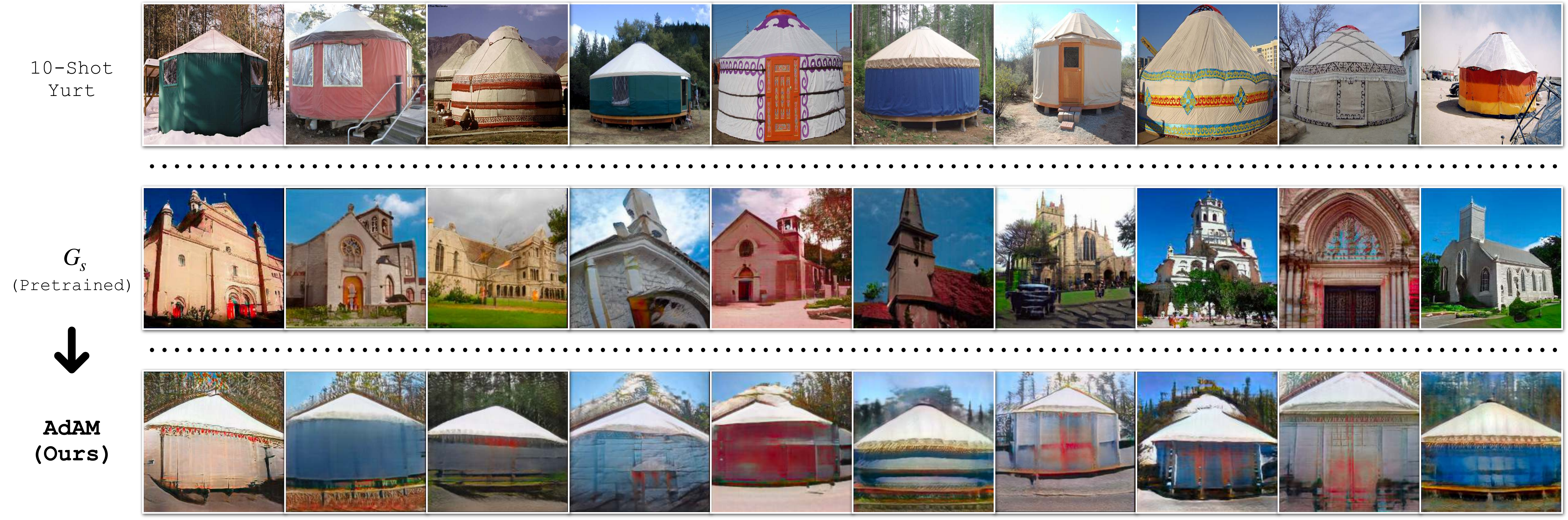}
    \caption{Church $\rightarrow$ Yurt (distant domain)}
    \label{fig:supp_yurt}
\end{figure}

Further, we show complete 10-shot adaptation results.
Results for Haunted Houses, Van Gogh Houses, Palace and Yurt are shown in Figures \ref{fig:supp_haunted}, \ref{fig:supp_van}, \ref{fig:supp_yurt}, \ref{fig:supp_palace} respectively. Other adaptation setups with FFHQ/Cars as source are shown in Figure \ref{fig:supp_sunglasses}, Figure \ref{fig:supp_metface}, Figure \ref{fig:supp_sketches}, Figure \ref{fig:supp_amedeo}, Figure \ref{fig:supp_otto}, Figure \ref{fig:supp_raphael} and Figure \ref{fig:supp_cars}.

\begin{figure}[h]
    \centering
    \includegraphics[width=\textwidth]{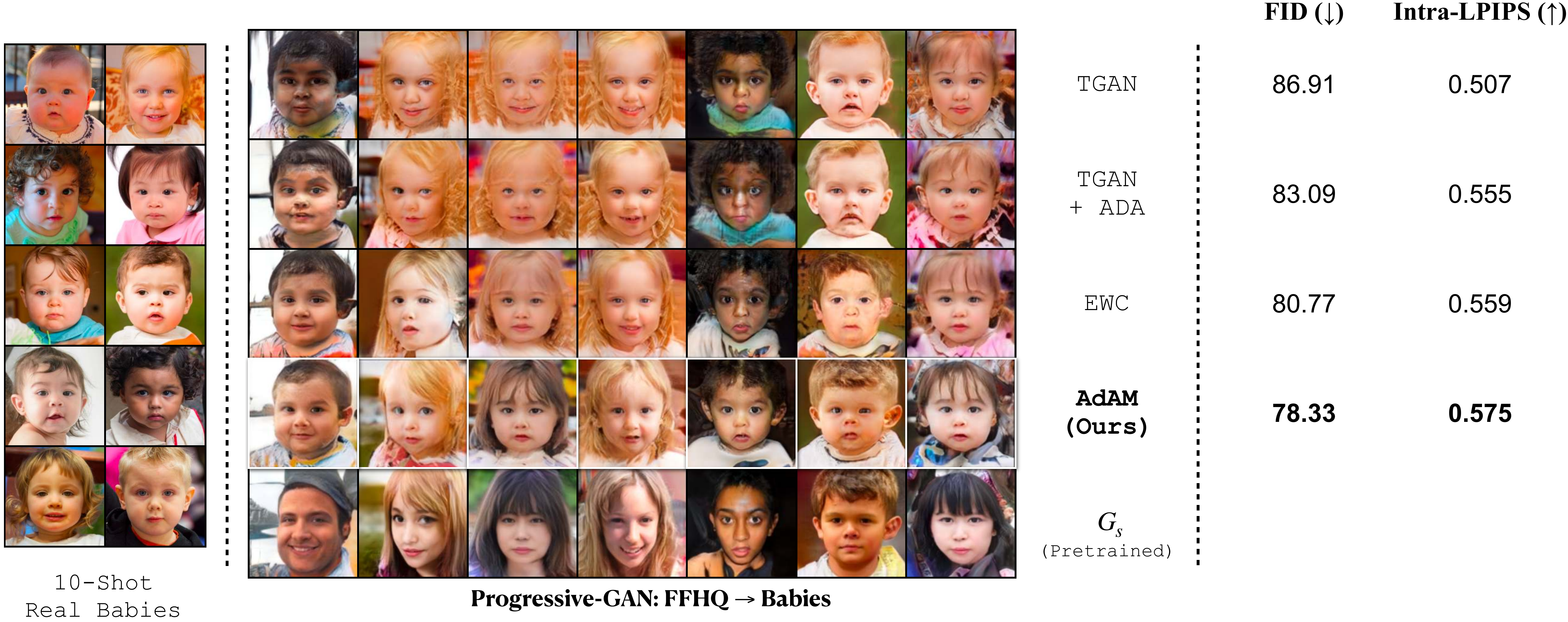}
    \caption{
    FFHQ $\rightarrow$ Babies 10-shot adaptation results using pre-trained ProGAN \cite{karras2017progan} generator. We include ADA results. As one can observe, our proposed method outperforms existing FSIG methods.}
    \label{fig:rebuttal_progan_babies}
\end{figure}

\begin{figure}[h]
    \centering
    \includegraphics[width=\textwidth]{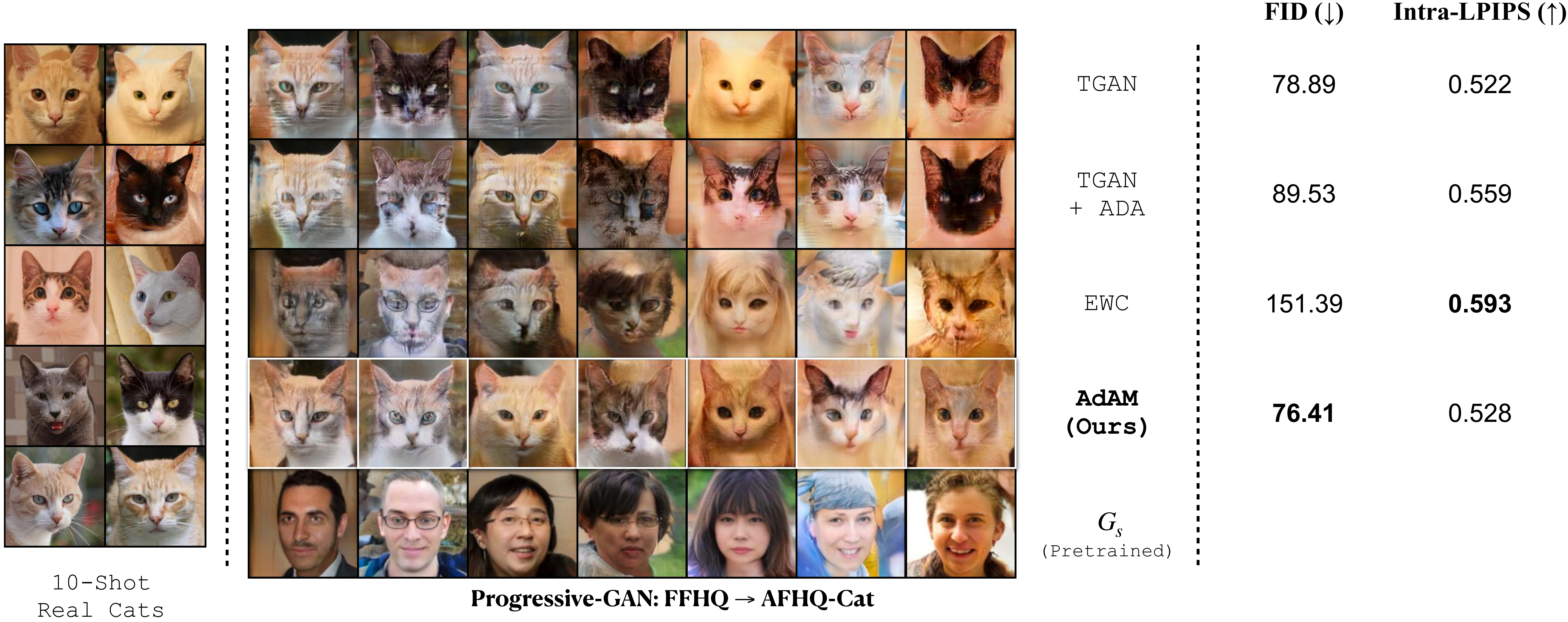}
    \caption{
    FFHQ $\rightarrow$ Cat 10-shot adaptation results using pre-trained ProGAN \cite{karras2017progan} generator. We include ADA results. As one can observe, our proposed method outperforms existing FSIG methods.}
    \label{fig:rebuttal_progan_cat}
\end{figure}

\begin{figure}[h]
    \centering
    \includegraphics[width=\textwidth]{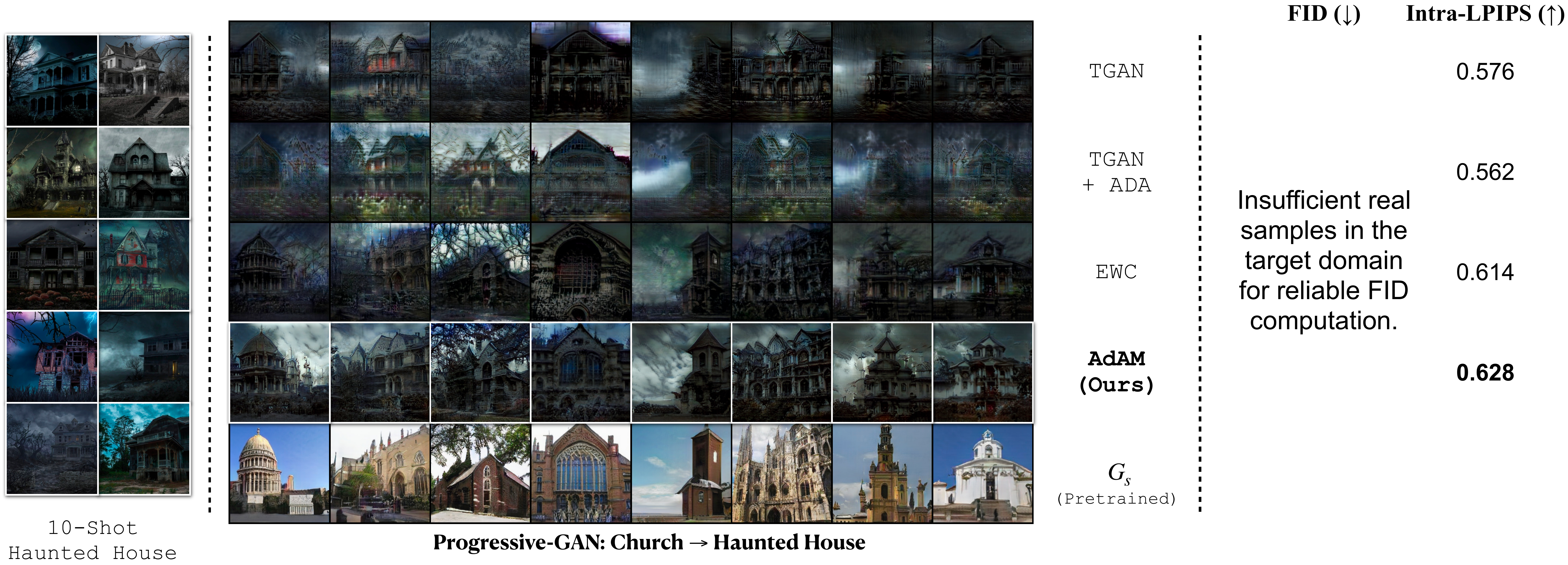}
    \caption{
    Church $\rightarrow$ Haunted Houses 10-shot adaptation results using pre-trained ProGAN \cite{karras2017progan} generator. We include ADA results.}
    \label{fig:rebuttal_progan_haunted_houses}
\end{figure}

\begin{figure}[h]
    \centering
    \includegraphics[width=\textwidth]{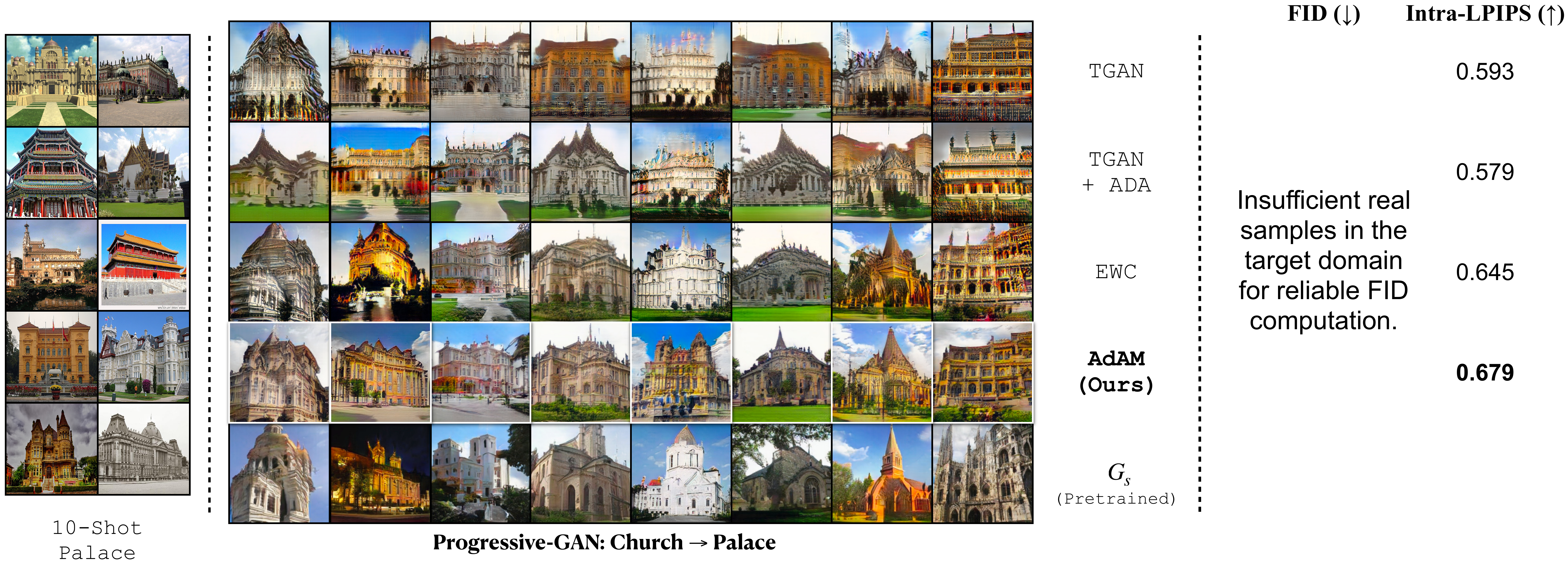}
    \caption{
    Church $\rightarrow$ Palace 10-shot adaptation results using pre-trained ProGAN \cite{karras2017progan} generator. We include ADA results.}
    \label{fig:rebuttal_progan_palace}
\end{figure}

{
\subsection{Additional GAN Architectures}
\label{subsec-supp:additional_gan_architectures}

We use an additional pre-trained GAN architecture, ProGAN \cite{karras2017progan}, to conduct FSIG experiments for FFHQ $\rightarrow$ Babies, FFHQ $\rightarrow$ Cat, Church $\rightarrow$ Haunted houses and Church $\rightarrow$ Palace setups. For fair comparison, we strictly follow the exact experiment setup discussed in Section \ref{subsec-supp:additional_source_target_domains}.

\textbf{Results.} We show complete qualitative and quantitative results for FFHQ $\rightarrow$ Babies, FFHQ $\rightarrow$ Cat adaptation in Figures \ref{fig:rebuttal_progan_babies} and \ref{fig:rebuttal_progan_cat} respectively. As one can observe, our proposed method consistently outperforms other baseline and SOTA FSIG methods with another pre-trained GAN model (ProGAN \cite{karras2017progan}), demonstrating the effectiveness and generalizability of our method. We also show qualitative results for Church $\rightarrow$ Haunted houses and Church $\rightarrow$ Palace adaptation in Figures \ref{fig:rebuttal_progan_haunted_houses} and \ref{fig:rebuttal_progan_palace} respectively.
}

{
\subsection{Alternative characterization of importance measure}
\label{subsec-supp:class_saliency}
In literature, Class Salience \cite{simonyan2013deep} (CS) is used as a property to explain which area/pixels of an input image stand out for a specific classification decision. Similar to the estimated Fisher Information (FI) used in our work, the complexity of CS is based on the first-order derivatives. Therefore, conceptually CS could have a connection with FI as they both use the knowledge encoded in the gradients.

We perform an experiment to replace FI with CS in importance probing and compare with our original approach. Note that, in \cite{simonyan2013deep}, CS is computed w.r.t. input image pixels. To make CS suitable for our problem, we modify it and compute CS w.r.t. modulation parameters. Similar to our approach in the main paper, we average the importance of all parameters within a kernel to calculate the importance of that kernel. Then we use these values during our importance probing to determine the important kernels for adapting from source to target domain (as Sec. 4 in our main paper). The results in Table \ref{table:cs-fi} are obtained with our proposed method using FI and CS during importance probing:
}
\begin{table}[h]  
    \caption{
    {
    In this experiment, we replace FI with CS in importance probing and compare with our original approach. We evaluate the performance under different source $\rightarrow$ target adaptation setups.}
    }
   \centering
        \begin{tabular}{l| c c | cc }
        \toprule
        \textbf{Domain}
         & \multicolumn{2}{c}{\textbf{FFHQ $\rightarrow$ Babies}}
         & \multicolumn{2}{c}{\textbf{FFHQ $\rightarrow$ Cat}}
         \\ 
         & {FID ($\downarrow$)} & {Intra-LPIPS ($\uparrow$)} & {FID ($\downarrow$)} & {Intra-LPIPS ($\uparrow$)}  \\
                \hline
        Class Salience \cite{simonyan2013deep}  & 52.46 & 0.582 & 61.68 & 0.556 \\
        Fisher Information (Ours) & \textbf{48.83} & \textbf{0.590} & \textbf{58.07} & \textbf{0.557} \\
        \bottomrule
        \end{tabular}
    \label{table:cs-fi}
\end{table}

{
Our results suggest that importance probing using FI (approximated by first-order derivatives) can perform better in selection of important kernels, leading to better performance (FID, intra-LPIPS) in the adapted models as shown in the Table \ref{table:cs-fi}.
}

\begin{figure}[h]
    \centering
    \includegraphics[width=\textwidth]{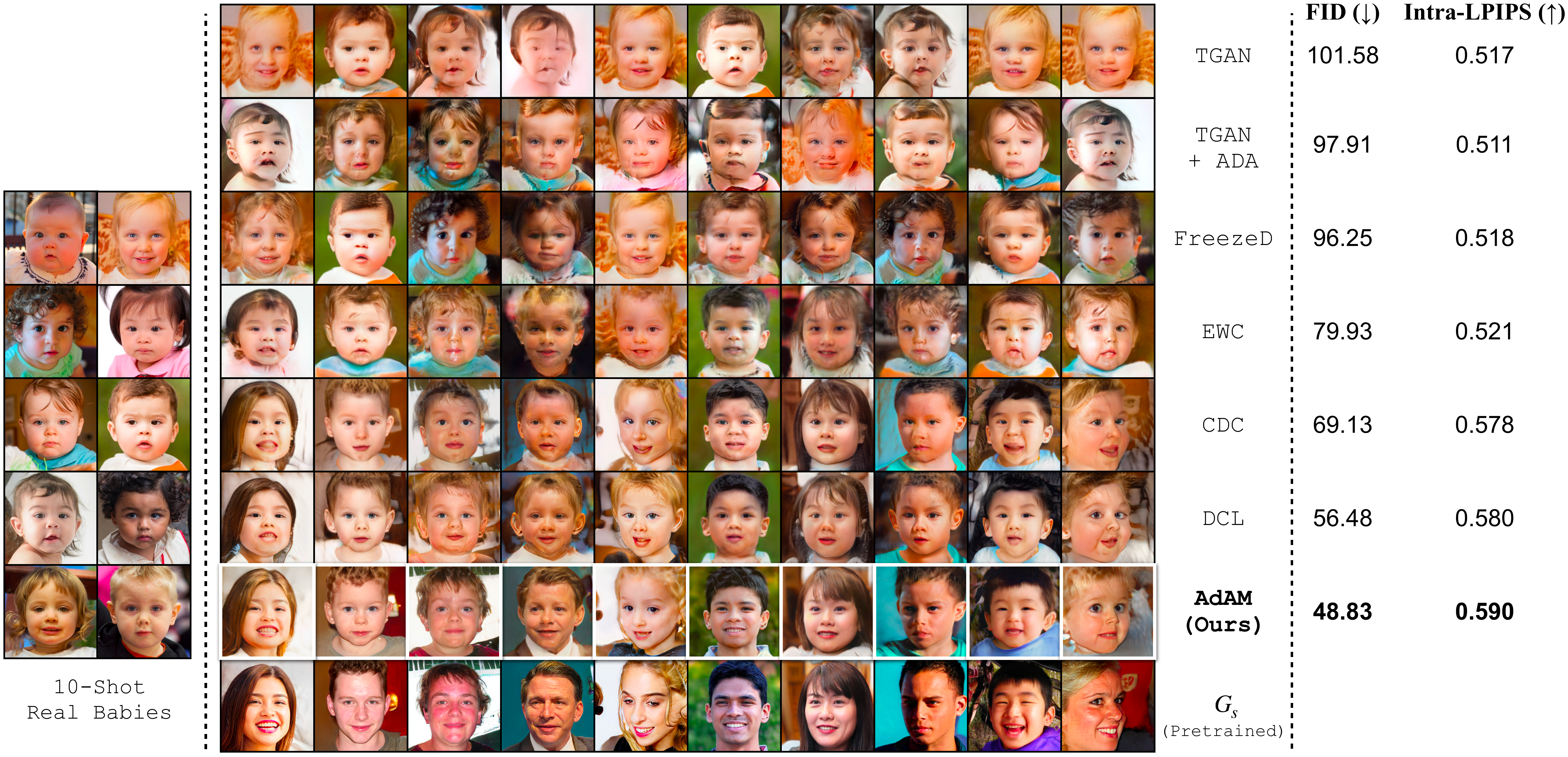}
    \caption{FFHQ $\rightarrow$ Babies results, including ADA \cite{karras2020ADA}.}
    \label{fig:supp_ada_babies}
\end{figure}

\begin{figure}[h]
    \centering
    \includegraphics[width=\textwidth]{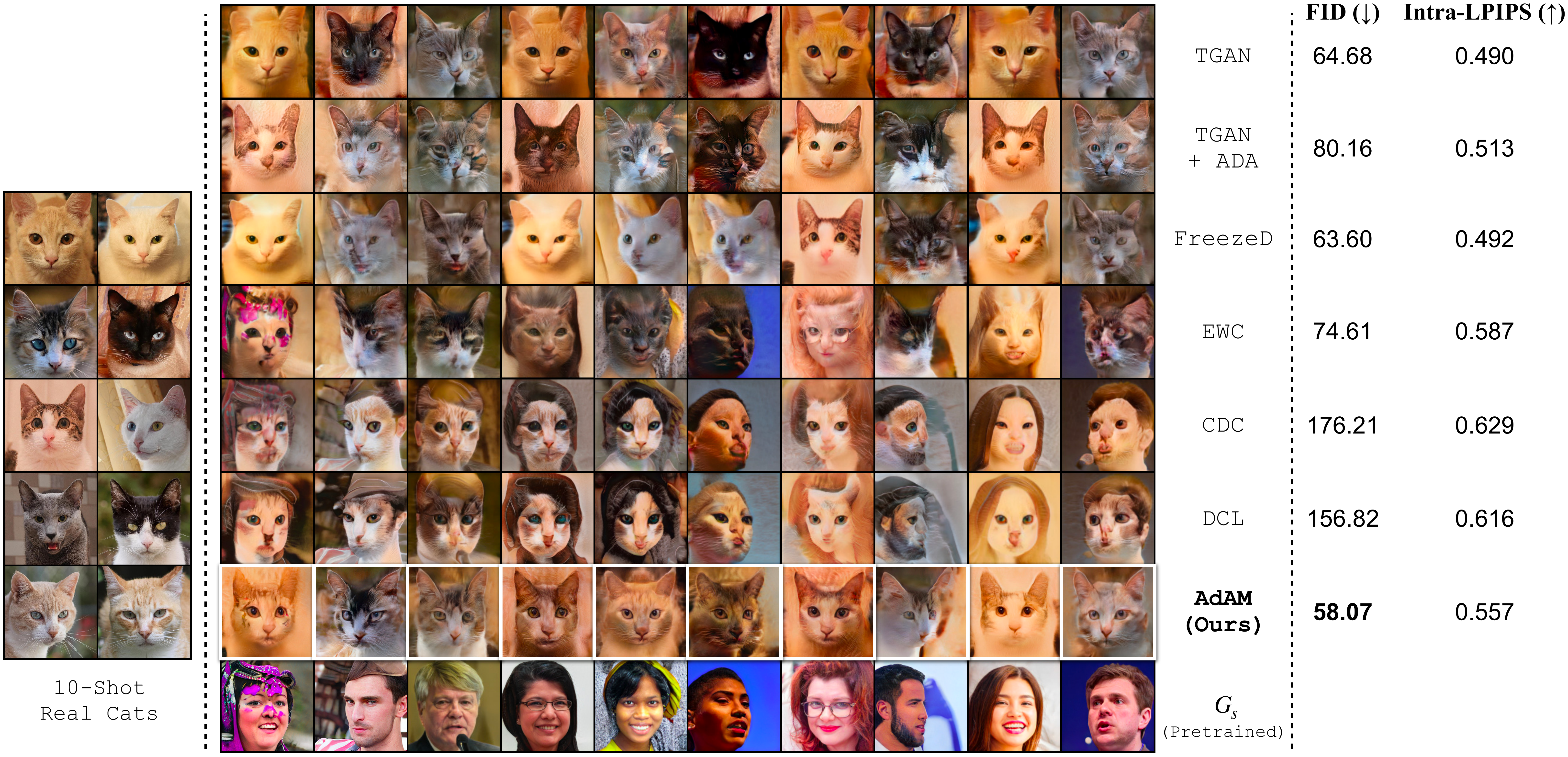}
    \caption{FFHQ $\rightarrow$ Babies results, including ADA \cite{karras2020ADA}.}
    \label{fig:supp_ada_cat}
\end{figure}

{
\subsection{Comparison with Adaptive Data Augmentation \cite{karras2020ADA}}
\label{subsec-supp:comparison_with_ada}

We additionally include the results of Adaptive Data Augmentation \cite{karras2020ADA} (ADA), as a supplement to Figure 4 in the main paper. We show that our proposed method consistently outperforms ADA in few-shot adaptation setups. The results are shown in Figures \ref{fig:supp_ada_babies} and \ref{fig:supp_ada_cat}.
}

\begin{figure}[h]
    \centering
    \includegraphics[width=\textwidth]{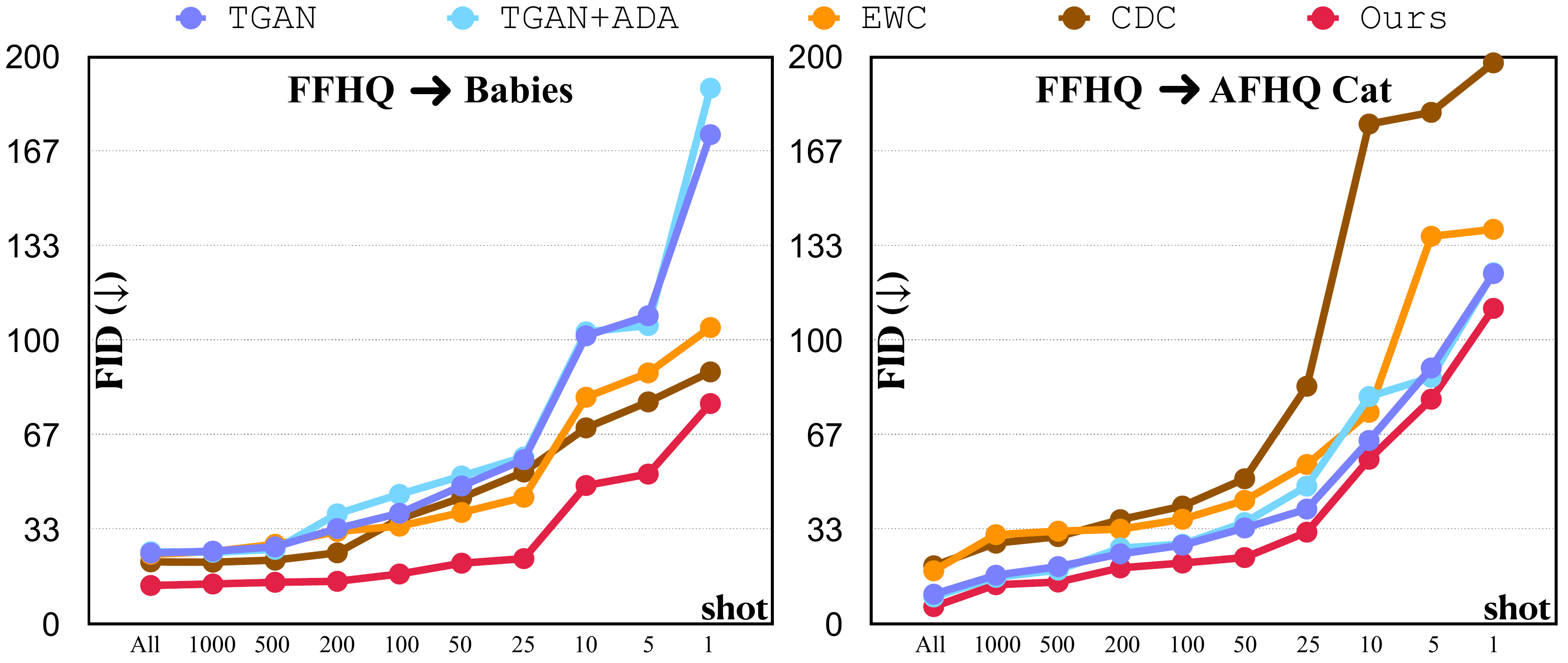}
    \caption{We add more data points based on Figure 6 in the main paper, and conduct experiments given extremely limited number of samples. We also include the entire dataset for adaptation.}
    \label{fig:fig6_extension}
\end{figure}

{
\subsection{Importance probing with extremely limited number of samples.}
\label{subsec-supp:ip_with_limited_samples}
In Figure 6 (main paper), we perform ablation studies to show that our method  consistently outperforms other baseline and SOTA methods given different number of target samples. In this section, we conduct additional experiments with extremely limited number of target samples: 1-shot and 5-shot. We also conduct experiments with more training samples during adaptation to show that our method consistently outperforms existing FSIG methods. 

\textbf{Results.}
The results are shown in Figure \ref{fig:fig6_extension}. Here we additionally include Adaptive Data Augmentation as an important baseline, and qualitative results can be found in Figures \ref{fig:supp_ada_babies} and \ref{fig:supp_ada_cat}.
}

{

\section{Discussion: What form of visual information is encoded by high FI kernels?}
\label{sec:rebuttal_visualizing_high_FI_kernels}

In this section, we attempt to discover what form of visual information is encoded/generated by a specific high FI kernel identified by our importance probing method.
This is a complex problem and to our best knowledge, methods on visualizing generative models/GANs are still rather restrictive in terms of concepts that can be visualized. 
Nevertheless, we leverage on GAN Dissection method \cite{bau2019gan}, a more established visualization method to visualize the high FI internal representations. 

\textbf{Experiment setup:} We use Church as the source domain as official GAN Dissection method
\footnote{https://github.com/CSAILVision/gandissect} is more suitable for scene-based image generation models (This is due to limitation of the semantic segmentation pipeline in GAN Dissection \cite{bau2019gan}). We use 2 target domains: haunted houses (proximal domain) and palace (distant domain). Following official GAN Dissection implementation \cite{bau2019gan}, we use the ProGAN \cite{karras2017progan} model. For fair comparison, we strictly follow the exact experiment setup discussed in Section \ref{subsec-supp:additional_source_target_domains}.

\textbf{Results.} 
\begin{itemize}
    \item Visualizing high FI kernels for Church $\rightarrow$ Haunted Houses adaptation : The results for FI estimation for kernels and several distinct semantic concepts learnt by high FI kernels are shown in Figure \ref{fig:rebuttal_gan_dissection_haunted_houses}.
    In Figure \ref{fig:rebuttal_gan_dissection_haunted_houses}, we visualize four examples of high FI kernels: (a), (b), (c), (d) corresponding to concepts building, building, tree and wood respectively. Using GAN Dissection, we observe that a notable amount of high FI kernels correspond to useful source domain concepts including building, tree and wood (texture) which are preserved when adapting to Haunted Houses target domain. We remark that these preserved concepts are useful to the target domain for adaptation. 
    
    \item Visualizing high FI kernels for Church $\rightarrow$ Palace adaptation : The results for FI estimation for kernels and several distinct semantic concepts learnt by high FI kernels are shown in Figure \ref{fig:rebuttal_gan_dissection_palace}.
    In Figure \ref{fig:rebuttal_gan_dissection_palace}, we visualize four examples of high FI kernels: (a), (b), (c), (d) corresponding to concepts grass, grass, building and building respectively. 
    Using GAN Dissection, we observe that a notable amount of high FI kernels correspond to useful source domain concepts including grass and building which are preserved when adapting to Palace target domain. We remark that these preserved concepts are useful to the target domain (Palace) for adaptation. 
\end{itemize}

\begin{figure}[h]
    \centering
    \includegraphics[width=\textwidth]{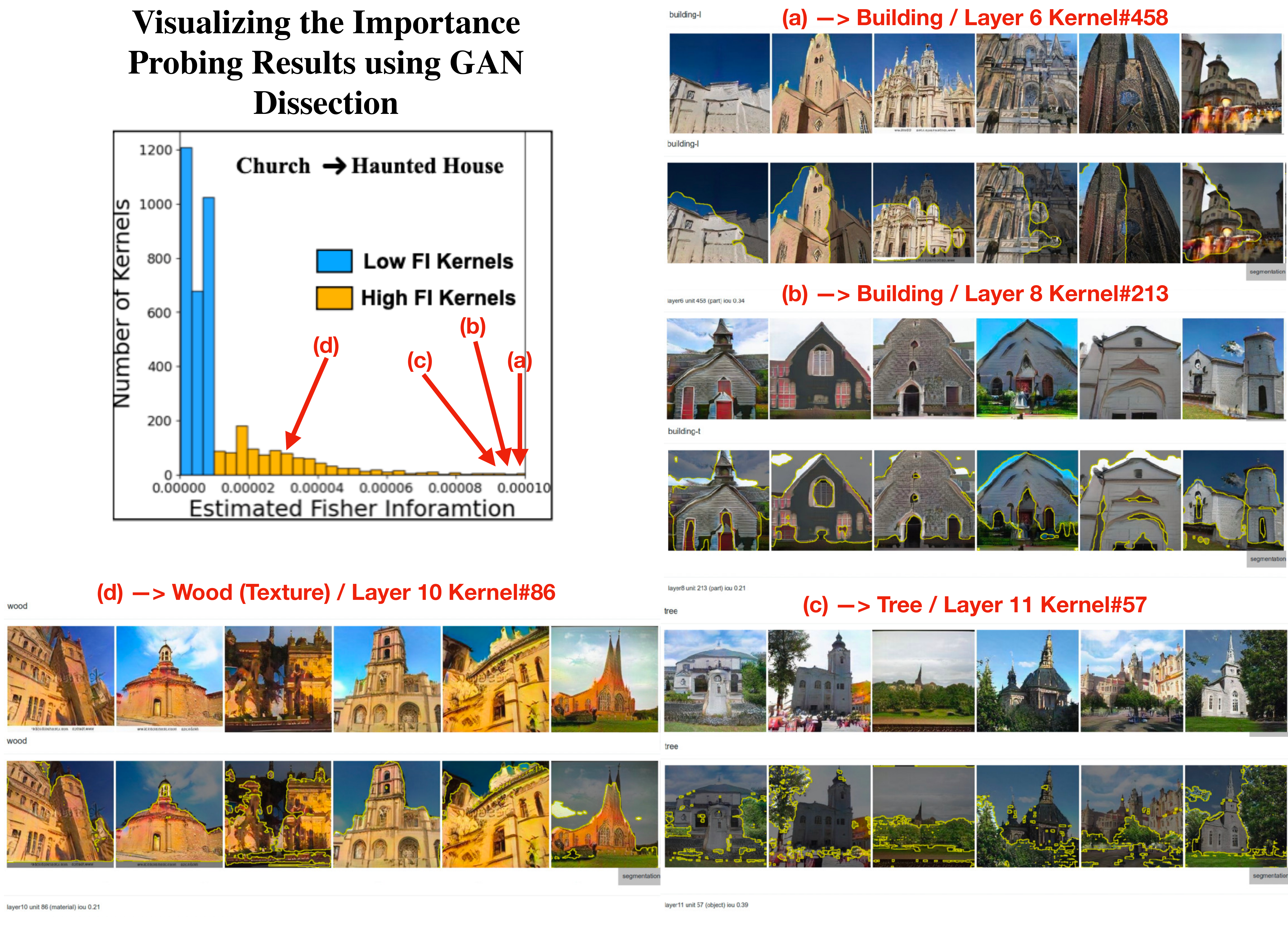}
    \caption{
    Visualizing high FI kernels using GAN Dissection \cite{bau2019gan} for Church $\rightarrow$ Haunted Houses 10-shot adaptation.
    In visualization of each high FI kernel, the first row shows different images generated by the source generator, and the second row highlights the concept encoded by the corresponding high FI kernel as determined by GAN Dissection.
    We observe that a notable amount of high FI kernels correspond to useful source domain concepts including building (a, b), tree (c) and wood (d) which are preserved when adapting to Haunted Houses target domain. We remark that these preserved concepts are useful to the target domain (Haunted House) for adaptation.}
    \label{fig:rebuttal_gan_dissection_haunted_houses}
\end{figure}

\begin{figure}[h]
    \centering
    \includegraphics[width=\textwidth]{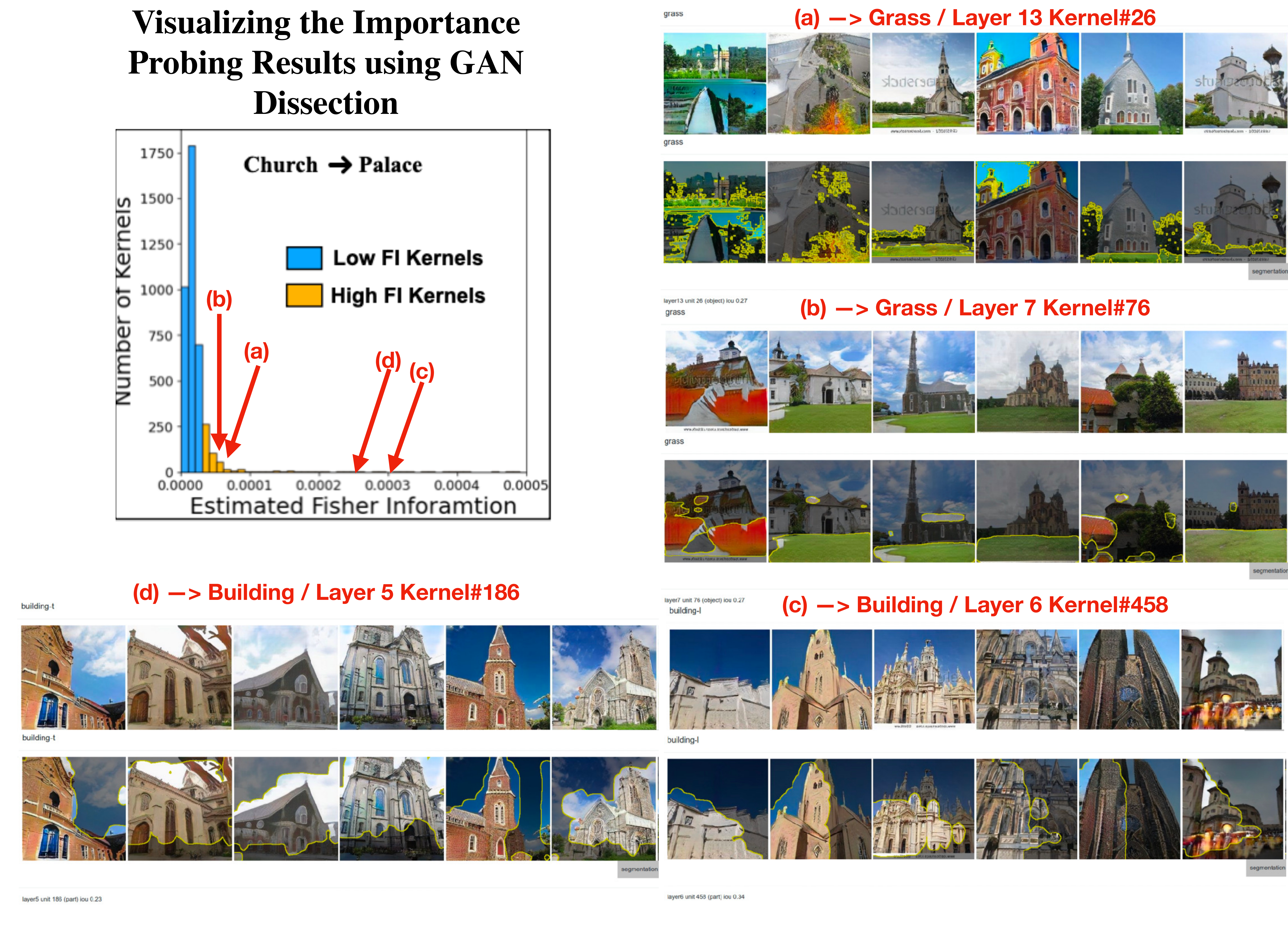}
    \caption{
    Visualizing high FI kernels using GAN Dissection \cite{bau2019gan} for Church $\rightarrow$ Palace 10-shot adaptation.
    In visualization of each high FI kernel, the first row shows different images generated by the source generator, and the second row highlights the concept encoded by the corresponding high FI kernel as determined by GAN Dissection.
    We observe that a notable amount of high FI kernels correspond to useful source domain concepts including grass (a, b) and building (c, d) which are preserved when adapting to Palace target domain. We remark that these preserved concepts are useful to the target domain (palace) for adaptation.}
    \label{fig:rebuttal_gan_dissection_palace}
\end{figure}

\textbf{Limitations of GAN Dissection / Future Work :} Although GAN Dissection can uncover useful semantic concepts preserved by high FI kernels, GAN Dissection method \cite{bau2019gan} is limited by the dataset used for semantic segmentation. Hence this method is not able to uncover concepts that are not present in semantic segmentation dataset (They use Broaden Dataset \cite{bau2017network}). Therefore, using GAN dissection we are currently unable to discover and visualize more fine-grained concepts preserved by our high FI kernels. We hope to further address this problem in future work.

}

\section{Main Paper Experiments : Additional Results / Analysis}
\label{sec-supp:additional_main_paper_results}

\subsection{KID / Intra-LPIPS / Standard Deviation of Experiments}
\label{sec-supp:kid_intr_lpips}

\textbf{KID / Intra-LPIPS.} 
In addition to FID scores reported in the main paper, we evaluate KID \cite{li2018mmd-aae} and Intra-LPIPS \cite{zhang2018lpips}.
We remark the KID ($\downarrow$) is another metric in addition to FID ($\downarrow$) to measure the quality of generated samples, and Intra-LPIPS ($\uparrow$) measures the diversity of generated samples. 
In literature, the original LPIPS \cite{zhang2018lpips} evaluates the perceptual distance between images. We follow CDC \cite{ojha2021fig_cdc} and DCL \cite{zhao2022dcl} to measure the Intra-LPIPS, a variant of LPIPs, to evaluate the degree of diversity. Firstly, we generate 5,000 images and assign them to one of 10-shot target samples, based on the closet LPIPS distance. Then, we calculate the LPIPS of 10 clusters and take average.
KID and Intra-LPIPS results are reported in Tables \ref{table-supp:kid} and \ref{table-supp:intra_lpips} respectively.
As one can observe, our proposed adaptation-aware FSIG method outperforms SOTA FSIG methods \cite{li2020fig_EWC, ojha2021fig_cdc, zhao2022dcl} and produces high quality images with good diversity.

\begin{table}[!h]
    \centering
    \caption{
   KID ($\downarrow$) score of different methods with the same checkpoint of Table {\color{red}2}
    in the main paper. The values are in $10^3$ units, following \cite{karras2020ADA, chai2021ensembling}.
    \label{tab:supp_kid}
    }
    \begin{adjustbox}{width=0.8\textwidth}
    \begin{tabular}{l|cccccccccc}
    \toprule
         Method & TGAN & FreezeD & EWC & CDC & DCL & AdAM (Ours) \\ \hline
         Babies  & $81.92$ & $65.14$ & $51.81$ & $51.74$ & $43.46$ & \textbf{28.38}\\
         AFHQ-Cat & $41.912$ & $38.834$ & $58.65$ & $196.60$ & $117.82$ & \textbf{32.78} \\
    \bottomrule
    \end{tabular}
    \end{adjustbox}
    \label{table-supp:kid}
\end{table}

\begin{table}[!h]
    \centering
    \caption{
    Intra-LPIPS ($\uparrow$) of different methods, the standard deviation is calculated over 10 clusters. Compared to the baseline models (TGAN/FreezeD) or state-of-the-art FSIG methods (EWC/CDC/DCL), our proposed method can achieve a good trade-off between diversity and quality of the generated images, see Table {\color{red} 2} in main paper for FID score.
    }
    \begin{adjustbox}{width=\textwidth}
    \begin{tabular}{l|cccccccccc}
    \toprule
         Method & TGAN & FreezeD & EWC & CDC & DCL & AdAM (Ours) \\ \hline
         Babies  & $0.517 \pm 0.04$ & $0.518 \pm 0.05$ & $0.521 \pm 0.03$ & $0.578 \pm 0.03$ & $0.580 \pm 0.02$ & $0.590 \pm 0.03$\\
         AFHQ-Cat & $0.490 \pm 0.02$ & $0.492 \pm 0.04$ & $0.587 \pm 0.04$ & $0.629 \pm 0.03$ & $0.616 \pm 0.05$ & $0.557 \pm 0.02$ \\
    \bottomrule
    \end{tabular}
    \end{adjustbox}
    \label{table-supp:intra_lpips}
\end{table}

\textbf{Standard Deviation of FID scores.}
We report standard deviation of FID scores for Babies and Cat corresponding to the  main paper experiments (Table 2: main paper) in Table \ref{table-supp:fid_std}. 
As one can observe, the standard deviations are within acceptable range.
\begin{table}[!h]
    \centering
    \caption{
    FID score ($\downarrow$) with standard deviation over 3 different runs.
    }
    \begin{adjustbox}{width=\textwidth}
    \begin{tabular}{l|cccccccccc}
    \toprule
         \textbf{Method} & TGAN & FreezeD & EWC & CDC & DCL & AdAM (Ours) \\ \hline
         Babies & $101.69 \pm 0.50$ & $97.15 \pm 1.02$ & $79.59 \pm 0.26$ & $66.98 \pm 1.58$ & $56.64 \pm 0.90$ & \bm{$47.92 \pm 0.87$} \\
         AFHQ-Cat & $64.60 \pm 0.68$ & $64.56 \pm 0.69$ & $74.69 \pm 0.32$ & $174.5 \pm 2.55$ & $154.60 \pm 1.98$ & \bm{$57.59 \pm 0.36$} \\
    \bottomrule
    \end{tabular}
    \end{adjustbox}
    \label{table-supp:fid_std}
\end{table}

\subsection{10-shot Adaptation Results}
\label{sup-sec:extended_experiments_10_shot_results}
\begin{figure}[!h]
    \centering
    \includegraphics[width=\textwidth]{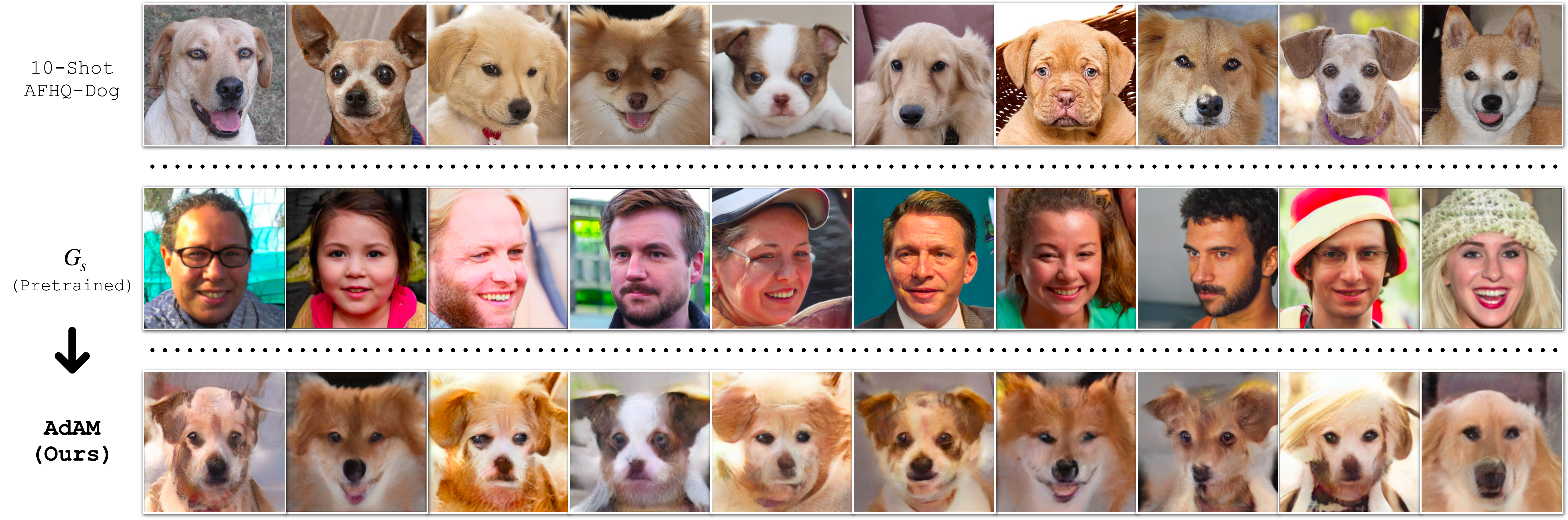}
    \caption{FFHQ $\rightarrow$ AFHQ-Dog (distant domain)}
    \label{fig:supp_dog}
\end{figure}

\begin{figure}[!h]
    \centering
    \includegraphics[width=\textwidth]{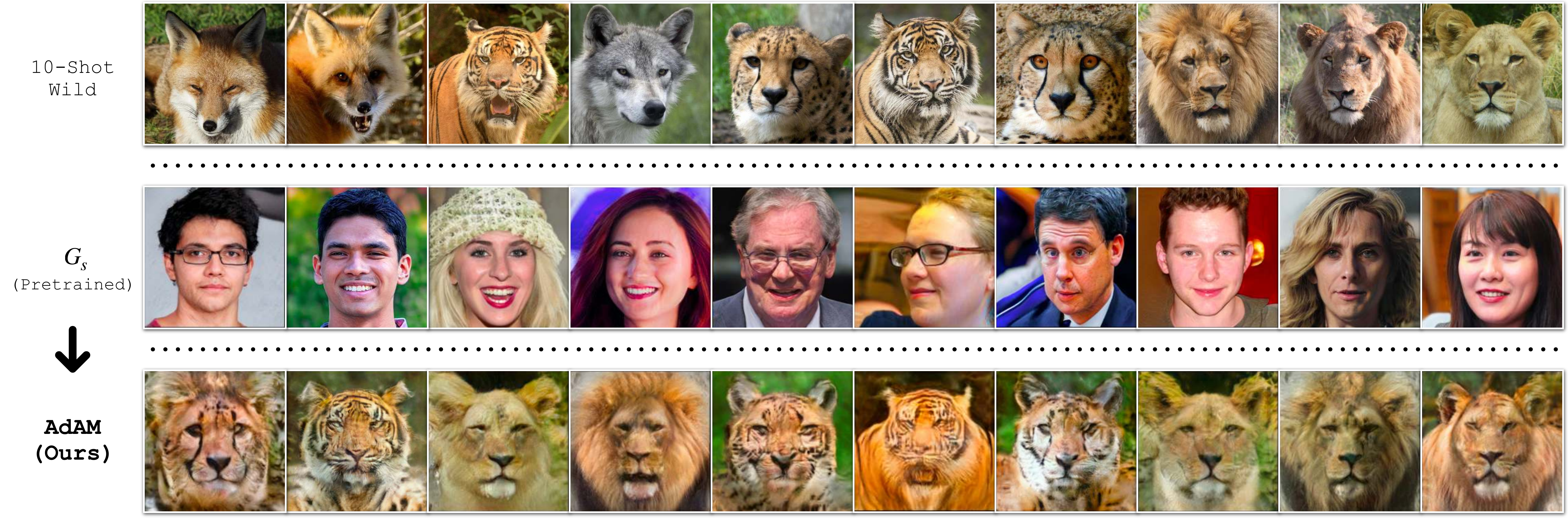}
    \caption{FFHQ $\rightarrow$ AFHQ-Wild (distant domain)}
    \label{fig:supp_wild}
\end{figure}

\begin{figure}[!h]
    \centering
    \includegraphics[width=0.6\textwidth]{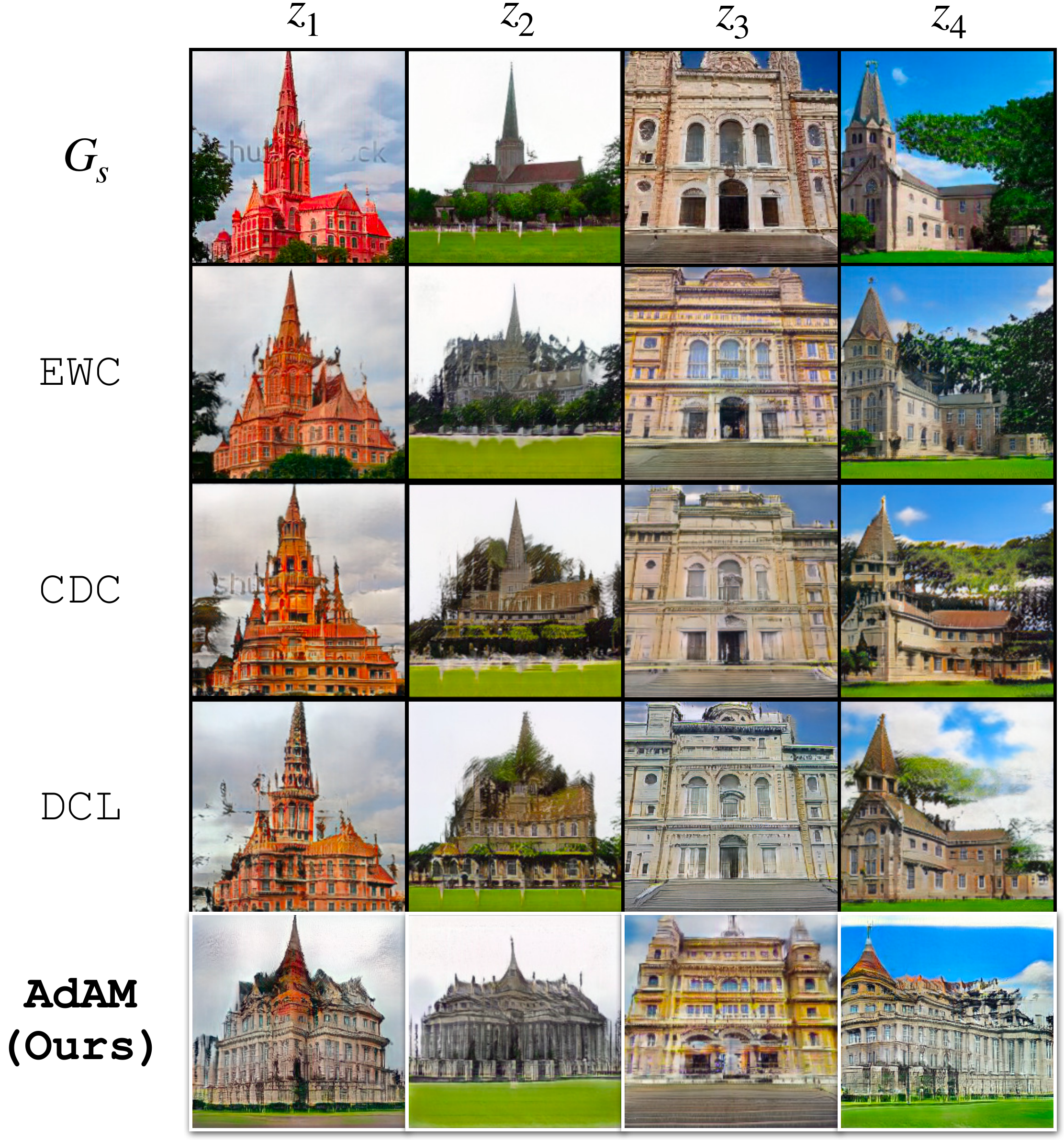}
    \caption{Church $\rightarrow$ Palace (distant domain)}
    \label{fig:supp_failure_palace}
\end{figure}

We show complete 10-shot adaptation results for our proposed adaptation-aware FSIG method and existing FSIG methods \cite{wang2018transferringGAN, li2020fig_EWC, ojha2021fig_cdc, zhao2022dcl} for distant target domains. 
Results for FFHQ $\rightarrow$ Dog and FFHQ $\rightarrow$ Wild are shown in Figures \ref{fig:supp_dog} and \ref{fig:supp_wild} respectively.
As one can observe, SOTA FSIG methods \cite{li2020fig_EWC, ojha2021fig_cdc, zhao2022dcl} are unable to adapt well to distant target domains (palace, yurt) due to \textit{due to only considering source domain / task in knowledge preservation.}
We remark that TGAN \cite{wang2018transferringGAN} suffers severe mode collapse.
We clearly show that our proposed adaptation-aware FSIG method outperforms SOTA FSIG methods \cite{li2020fig_EWC, ojha2021fig_cdc, zhao2022dcl} and produces high quality images with good diversity.

\begin{figure}[]
    \centering
    \includegraphics[width=\textwidth]{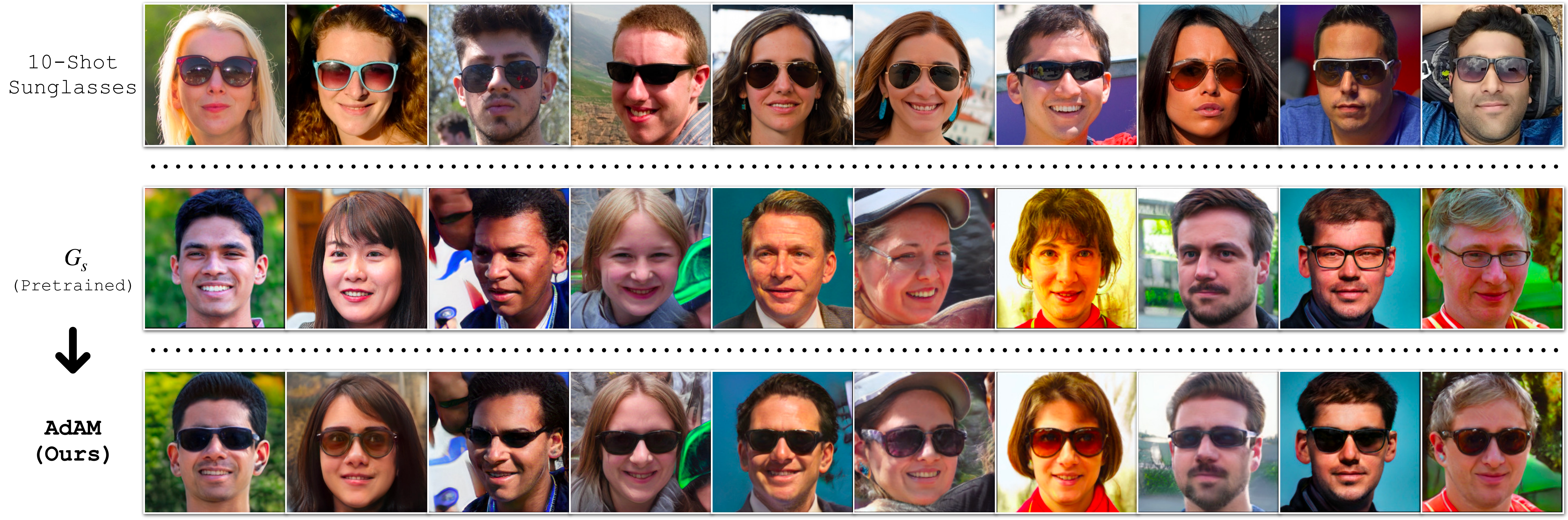}
    \caption{FFHQ $\rightarrow$ Sunglasses}
    \label{fig:supp_sunglasses}
\end{figure}

\begin{figure}[]
    \centering
    \includegraphics[width=\textwidth]{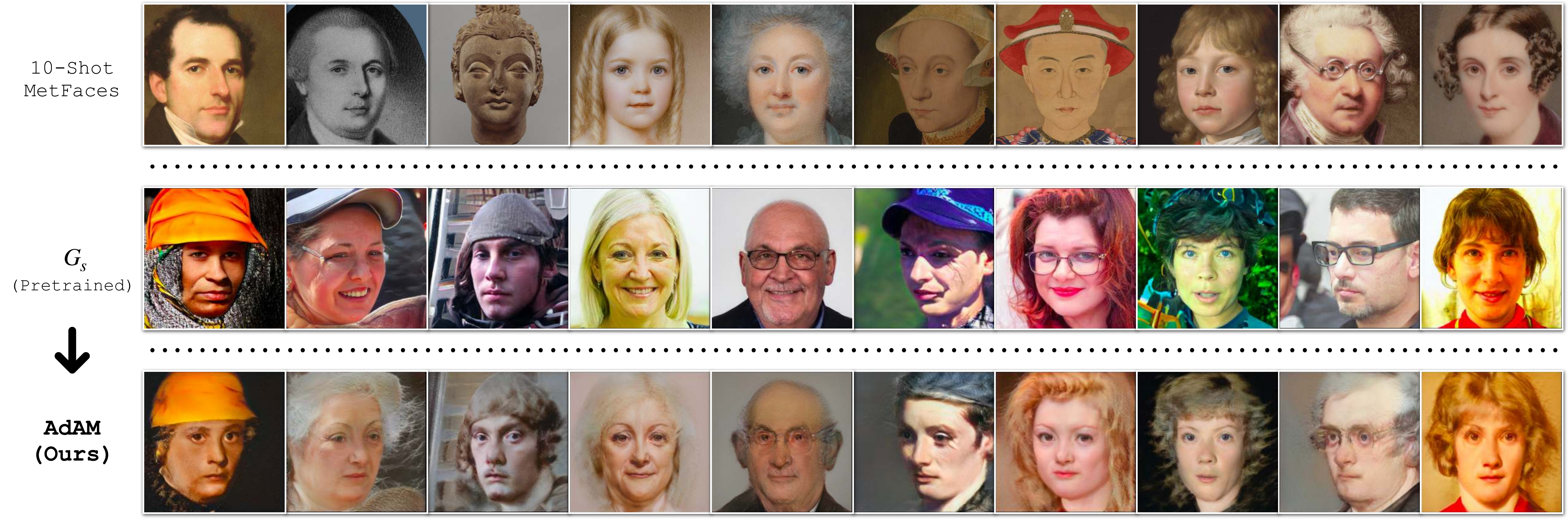}
    \caption{FFHQ $\rightarrow$ MetFaces}
    \label{fig:supp_metface}
\end{figure}

\begin{figure}[]
    \centering
    \includegraphics[width=\textwidth]{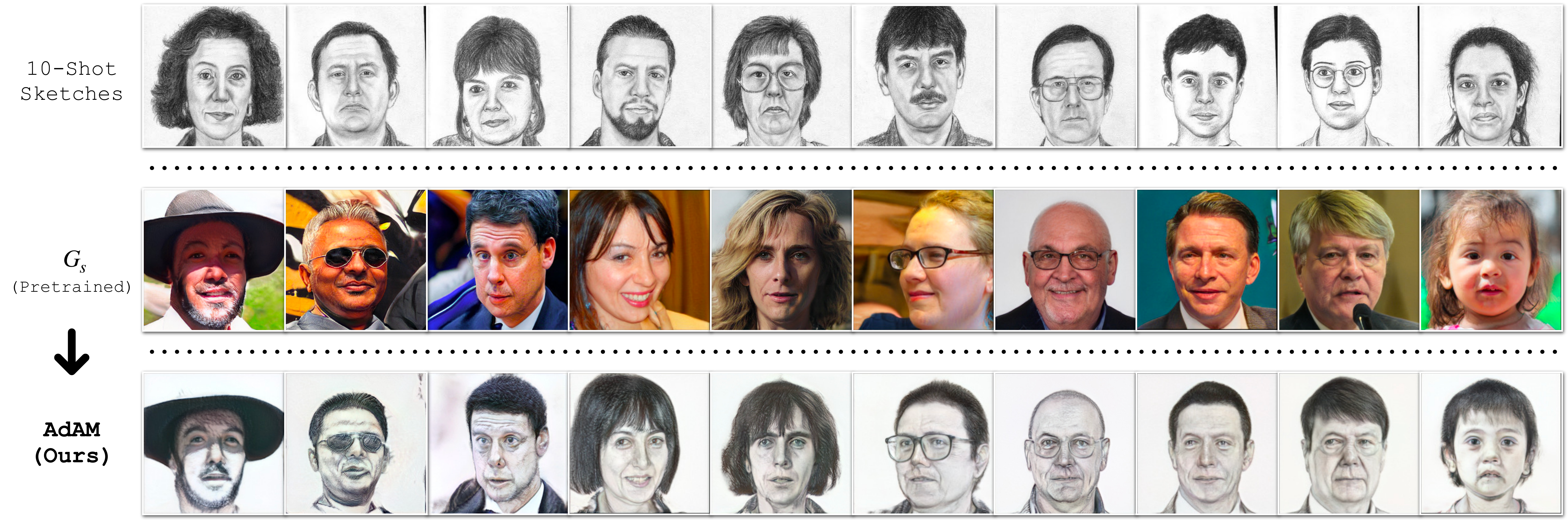}
    \caption{FFHQ $\rightarrow$ Sketches}
    \label{fig:supp_sketches}
\end{figure}

\begin{figure}[]
    \centering
    \includegraphics[width=\textwidth]{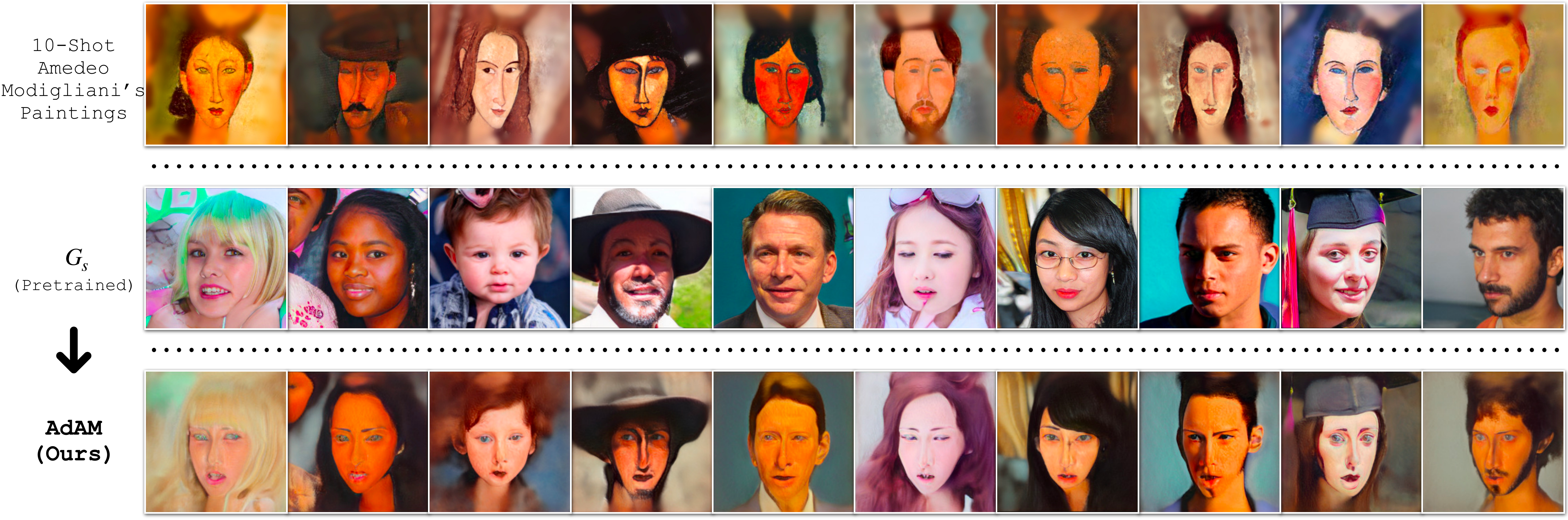}
    \caption{FFHQ $\rightarrow$ Amedeo Modigliani’s Paintings}
    \label{fig:supp_amedeo}
\end{figure}

\begin{figure}[]
    \centering
    \includegraphics[width=\textwidth]{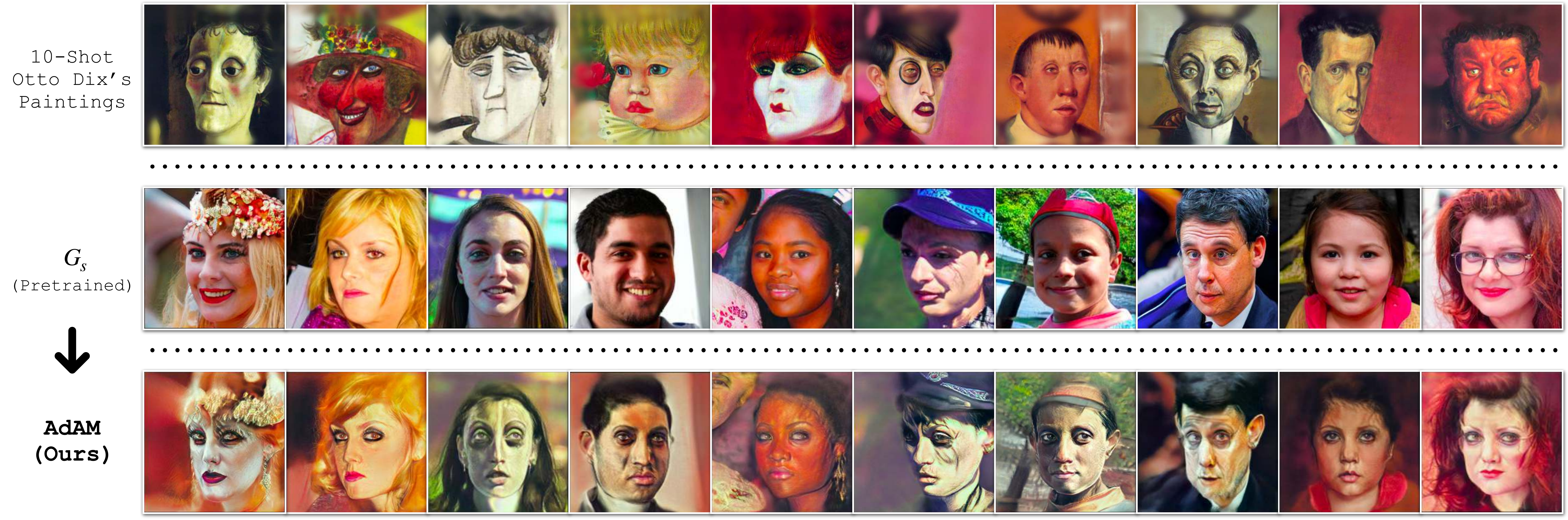}
    \caption{FFHQ $\rightarrow$ Otto Dix’s Paintings}
    \label{fig:supp_otto}
\end{figure}

\begin{figure}[]
    \centering
    \includegraphics[width=\textwidth]{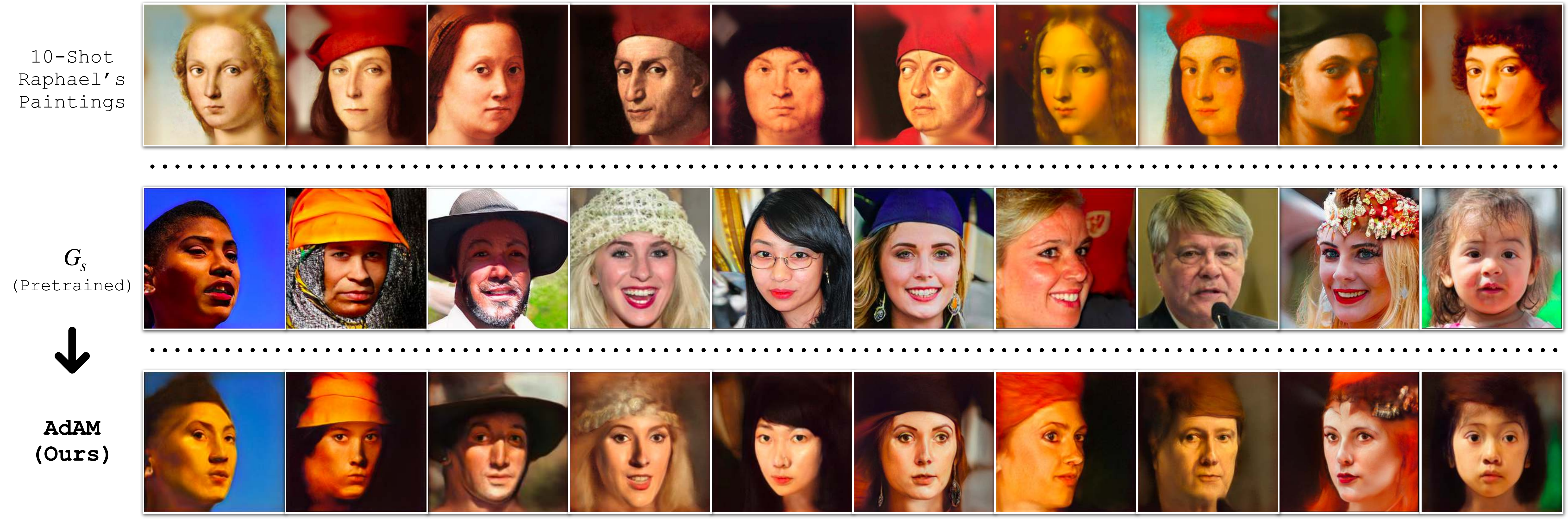}
    \caption{FFHQ $\rightarrow$ Raphael’s Paintings}
    \label{fig:supp_raphael}
\end{figure}

\begin{figure}[]
    \centering
    \includegraphics[width=\textwidth]{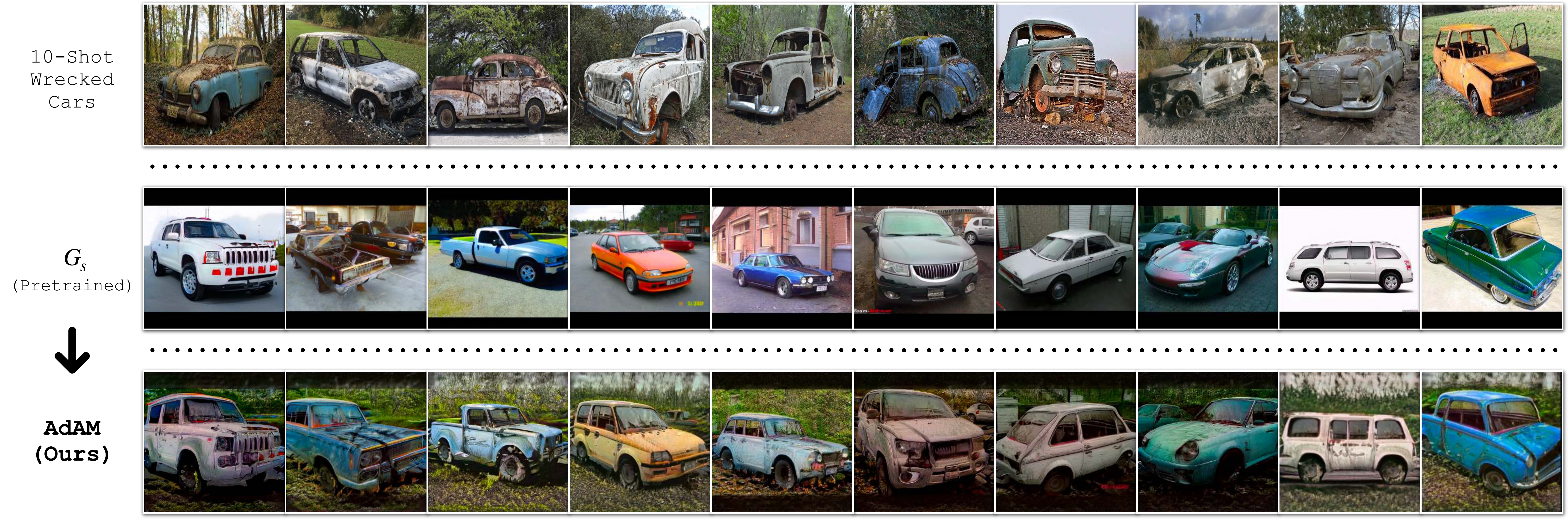}
    \caption{Cars $\rightarrow$ Wrecked Cars}
    \label{fig:supp_cars}
\end{figure}

We further show 10-shot adaptation results for our proposed adaptation-aware FSIG method for additional setups.
We show 10-shot adaptation results for 
FFHQ $\rightarrow$ MetFaces \cite{karras2020ADA} (Figure \ref{fig:supp_metface}), 
FFHQ $\rightarrow$ Sketches (Figure \ref{fig:supp_sketches}), 
FFHQ $\rightarrow$ Sunglasses (Figure \ref{fig:supp_sunglasses}), 
FFHQ $\rightarrow$ Amedeo Modigliani’s Paintings (Figure \ref{fig:supp_amedeo}),
FFHQ $\rightarrow$ Otto Dix’s Paintings (\ref{fig:supp_otto}) and 
Cars $\rightarrow$ Wrecked Cars (Figure \ref{fig:supp_cars}).

\subsection{100-shot adaptation}
\begin{figure}[]
    \centering
    \includegraphics[width=0.8\textwidth]{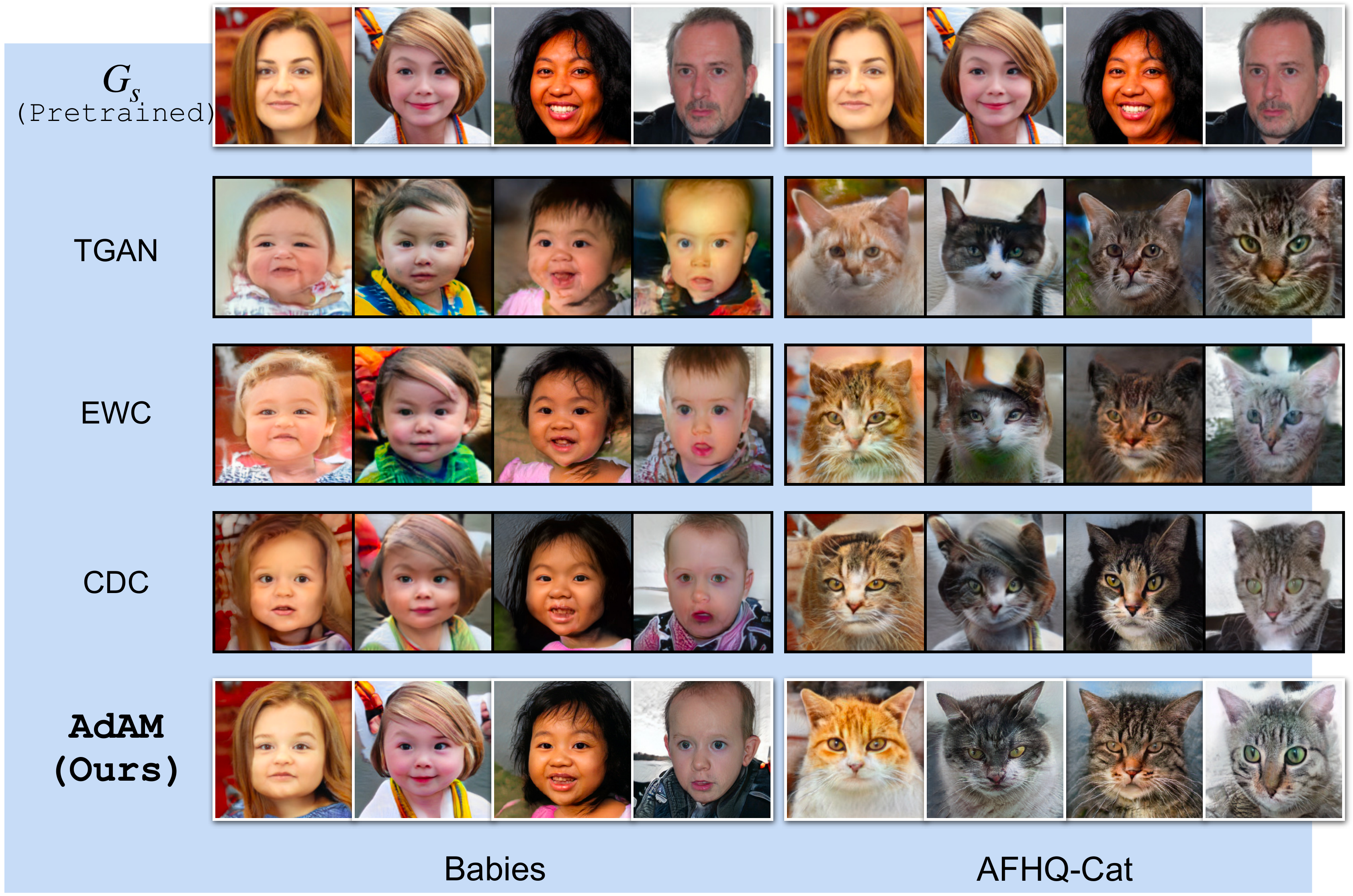}
    \caption{FFHQ $\rightarrow$ Babies (Left) and AFHQ-Cat (Right) with 100 samples for adaptation.}
    \label{fig:100-shot}
\end{figure}

In addition to the analysis of increasing the number of shots for target adaptation in Figure {\color{red} 6} of main paper, here we additionally show the generated images with 100-shot training data, on Babies and AFHQ-Cat. The results are shown in Figure \ref{fig:100-shot} where each column represents a fixed noise. Compared to baseline and SOTA methods, our generated images can still produce the best quality and  diversity.

\subsection{FID measurements with limited target domain samples}
\label{supp-sec:fid-analysis}
To characterize source $\rightarrow$ target domain proximity, we used FID and LPIPS measurements. 
FID involves distribution estimation using first-order (mean) and second-order (trace) moments, i.e.: $FID = mean_{component}+ trace_{component}$ \cite{heusel2017FID}
Generally, 50K real and generated samples are used for FID calculation
\footnote{Chong, Min Jin, and David Forsyth. "Effectively unbiased fid and inception score and where to find them." Proceedings of the IEEE/CVF conference on computer vision and pattern recognition. 2020.}.
Given that our target domain datasets contain limited samples, i.e.: Cat \cite{choi2020starganv2}, Dog \cite{choi2020starganv2}, Wild \cite{choi2020starganv2} datasets contain $\approx$ 5K samples, we conduct extensive experiments to show that FID measurements with limited samples give reliable estimates, thereby reliably characterizing source $\rightarrow$ target domain proximity.
Specifically, we decompose FID into mean and trace components and study the effect of target domain sample size to show that our proximity measurements using FID are reliable.

\textbf{Experiment Setup.}
We use 3 large datasets namely FFHQ \cite{karras2020styleganv2} (70K samples), LSUN-Bedroom \cite{yu15lsun} (70K samples) and LSUN-Cat \cite{yu15lsun} (70K samples). 
We use FFHQ (70K samples) as the source domain and study the effect of sample size on FID measure.
Specifically, we decompose FID into mean and trace components in this study.
We consider FFHQ (self-measurement), LSUN-Bedroom and LSUN-cat as target domains. 
We sample 13, 130, 1300, 2600, 5200, 13000, 52000 samples from the target domain and measure the FID with FFHQ (70K samples), and compare it against the FID obtained by using the entire 70K samples from the target domain.

\textbf{Results / Analysis.}
The results are shown in Table \ref{table-supp:fid-analysis}.
As one can observe, with $\approx$ 2600 samples, we can reliably estimate FID as it becomes closer to the FID measured using the entire 70K target domain samples.
Hence, we show that our source $\rightarrow$ target proximity measurements using FID are reliable.

\begin{table}[!h]
\caption{
\textit{FID measurements with limited target domain samples give reliable estimates to characterize source $\rightarrow$ target domain proximity:}
FFHQ (70K) is the source domain.
We use FFHQ (self-measurement), LSUN-Bedroom and LSUN-Cat as target domains.
We use different number of samples from target domain to measure FID. We also decompose FID into mean and trace components in this study.
We sample 13, 130, 1300, 2600, 5200, 13000, 52000 images from the target domain and measure the FID with source domain (FFHQ / 70K), and compare it against the FID obtained by using the entire 70K samples from the target domain.
Each experiment is repeated \textit{100 times} and we report the results with standard deviation.
We also report mean and trace components separately.
As one can observe, with $\approx$ 2600 samples, we can reliably estimate FID as it becomes closer to the FID measured using the entire 70K target domain samples.
Therefore, this study shows that our source $\rightarrow$ target proximity measurements using FID are reliable.
}
  \begin{adjustbox}{width=\textwidth}
  \begin{tabular}{l|c|cccccccc}\toprule
\textbf{FID} &\textbf{} &\textbf{13} &\textbf{130} &\textbf{1300} &\textbf{2600} &\textbf{5200} &\textbf{13, 000} &\textbf{52, 000} &\textbf{70, 000} \\ \toprule
\multirow{3}{*}{FFHQ}  &FID &196.3 $\pm$ 11.8 &83.4 $\pm$ 2.2 &15.3 $\pm$ 0.2 &\textbf{7 $\pm$ 0.1} &3.3 $\pm$ 0 &1.2 $\pm$ 0 &0.1 $\pm$ 0 &0 $\pm$ 0 \\
&mean &12 $\pm$ 2.7 &1.3 $\pm$ 0.3 &0.1 $\pm$ 0 &\textbf{0.1 $\pm$ 0} &0 $\pm$ 0 &0 $\pm$ 0 &0 $\pm$ 0 &0 $\pm$ 0 \\
&trace &184.3 $\pm$ 10.3 &82.2 $\pm$ 2 &15.2 $\pm$ 0.2 &\textbf{6.9 $\pm$ 0.1} &3.3 $\pm$ 0 &1.1 $\pm$ 0 &0.1 $\pm$ 0 &0 $\pm$ 0 \\ \midrule
\multirow{3}{*}{Bedroom} &FID &358.5 $\pm$ 9.3 &301.9 $\pm$ 2.4 &251 $\pm$ 1.2 &\textbf{243.6 $\pm$ 0.8} &240.1 $\pm$ 0.5 &238.2 $\pm$ 0.4 &237.2 $\pm$ 0.2 &237.2 $\pm$ 0.1 \\
&mean &139.3 $\pm$ 8 &131.8 $\pm$ 2.5 &131.4 $\pm$ 0.9 &\textbf{131.1 $\pm$ 0.6} &131.1 $\pm$ 0.4 &131.1 $\pm$ 0.3 &131.1 $\pm$ 0.1 &131.1 $\pm$ 0.1 \\
&trace &219.1 $\pm$ 9.9 &170.1 $\pm$ 1.9 &119.6 $\pm$ 0.6 &\textbf{112.5 $\pm$ 0.4} &109.1 $\pm$ 0.3 &107.1 $\pm$ 0.2 &106.1 $\pm$ 0.1 &106 $\pm$ 0.1 \\ \midrule
\multirow{3}{*}{Cat} &FID &370.2 $\pm$ 18.7 &283.7 $\pm$ 4.4 &209.7 $\pm$ 1.2 &\textbf{199.9 $\pm$ 0.8} &195.3 $\pm$ 0.6 &192.8 $\pm$ 0.4 &191.4 $\pm$ 0.2 &191.3 $\pm$ 0.1 \\
&mean &105.7 $\pm$ 8.4 &93 $\pm$ 2.2 &91.7 $\pm$ 0.8 &\textbf{91.7 $\pm$ 0.5} &91.6 $\pm$ 0.4 &91.6 $\pm$ 0.2 &91.6 $\pm$ 0.1 &91.6 $\pm$ 0.1 \\
&trace &264.5 $\pm$ 15.7 &190.7 $\pm$ 3.6 &118 $\pm$ 0.9 &\textbf{108.2 $\pm$ 0.6} &103.7 $\pm$ 0.4 &101.2 $\pm$ 0.3 &99.9 $\pm$ 0.1 &99.7 $\pm$ 0.1 \\
\bottomrule
\end{tabular}
\end{adjustbox}
\label{table-supp:fid-analysis}
\end{table}

{
\section{Discussion: How much can the proximity between source and target be relaxed?}
\label{sec-supp:rebuttal_proximity_relaxation}
In this section, we explore the proximity limitation between source and target domains in our experiment setups.
First, we remark that the upper bound on proximity between the source domain S and the target domain T could be conditioning on (a) the number of available samples (shots) from the target domain, and (b) the method used for knowledge transfer.

(a) Proximity bound conditioning on the number of target domain samples. In this paper, we focus on few-shot setups, e.g. 10 shots. However, with more target domain samples available, proximity between S and T can be further relaxed, and the proximity bound would increase, i.e. for a given generative model on S, we could learn an adapted model for T which is more distant. Intuitively, increasing the number of target domain samples can provide more diverse knowledge for T, and as a result, there is less reliance on the knowledge of S that is generalizable for T (which would decrease as S and T are more apart). In the limiting cases when abundant target domain samples are available, knowledge of S would not be critical, and proximity constraints between S and T may be totally relaxed (ignored).

(b) Proximity bound conditioning on the knowledge transfer method. Given a generative model pretrained on S and a certain number of available samples from T, the method used for knowledge transfer plays a critical role. If the method is superior in identifying suitable transferable knowledge from S to T, the proximity between S and T can be relaxed, and the proximity bound would increase. In our work, our first contribution is to reveal that existing SOTA approaches (which are based on target-agnostic ideas) are inadequate in identifying transferable knowledge from S to T. As a result, when proximity between S and T is relaxed, the performance of the adapted models is miserably poor, as discussed in Sec.3, Sec.5, and Appendix. Therefore, our second contribution is to propose a target-aware approach that could identify more meaningful transferable knowledge from S to T, allowing relaxation of the proximity constraint.

In this section, we provide experimental results for the adaptation between two very distant domains: FFHQ$\rightarrow$Cars using only 10-shots, aiming to answer two main questions: (1) Is there transferable knowledge from FFHQ to Cars for the FSIG task? (2) How does our proposed method compare with other methods in this setup? For this, in addition to transfer learning approaches discussed in the paper, we also add the results for training from scratch using only the same 10 Car samples. The quantitative results are in Table \ref{table:ffhq_cars}. 
}

\begin{table}[h]  
    \caption{
    We conduct experiments for {FFHQ $\rightarrow$ Cars} adaptation and evaluate the performance in such a challenging setup. We show that our method can achieve similar diversity as ADA \cite{karras2020ADA} and the overall performance (FID) is better than other baseline and SOTA methods.
    }
   \centering
        \begin{tabular}{l| c c}
        \toprule
        \textbf{Domain}
         & \multicolumn{2}{c}{\textbf{FFHQ $\rightarrow$ Cars}}
         \\ 
         & {FID ($\downarrow$)} & {Intra-LPIPS ($\uparrow$)} \\
                \hline
        Train from Scratch & 201.34 & 0.300 \\ 
        ADA \cite{karras2020ADA} & 171.98 & 0.438 \\
        EWC \cite{li2020fig_EWC} & 276.19 & \textbf{0.620} \\
        CDC \cite{ojha2021fig_cdc} & 109.53 & 0.484 \\
        DCL \cite{zhao2022dcl} & 125.96 & 0.464 \\
        AdAM (Ours) & \textbf{80.55} & 0.425
        \\
        \bottomrule
        \end{tabular}
    \label{table:ffhq_cars}
\end{table}

The results suggest that even though domain FFHQ and domain Cars are apart, there is still useful and transferable knowledge from FFHQ to Cars (e.g. low-level edges, shapes), leading to better performance (FID, Intra-LPIPS) in the adapted model using proposed method compared to the one which is trained from scratch. In addition, our proposed method can identify and transfer more meaningful knowledge compared to other baselines and SOTA methods, resulting in lower FID and higher diversity in generated images.


\section{Additional information for Checklist}
\label{sec-supp:checklist}

\subsection{Potential Societal Impact}
\label{sec-supp:societal_impact}
\begin{figure}[!t]
    \centering
    \includegraphics[width=\textwidth]{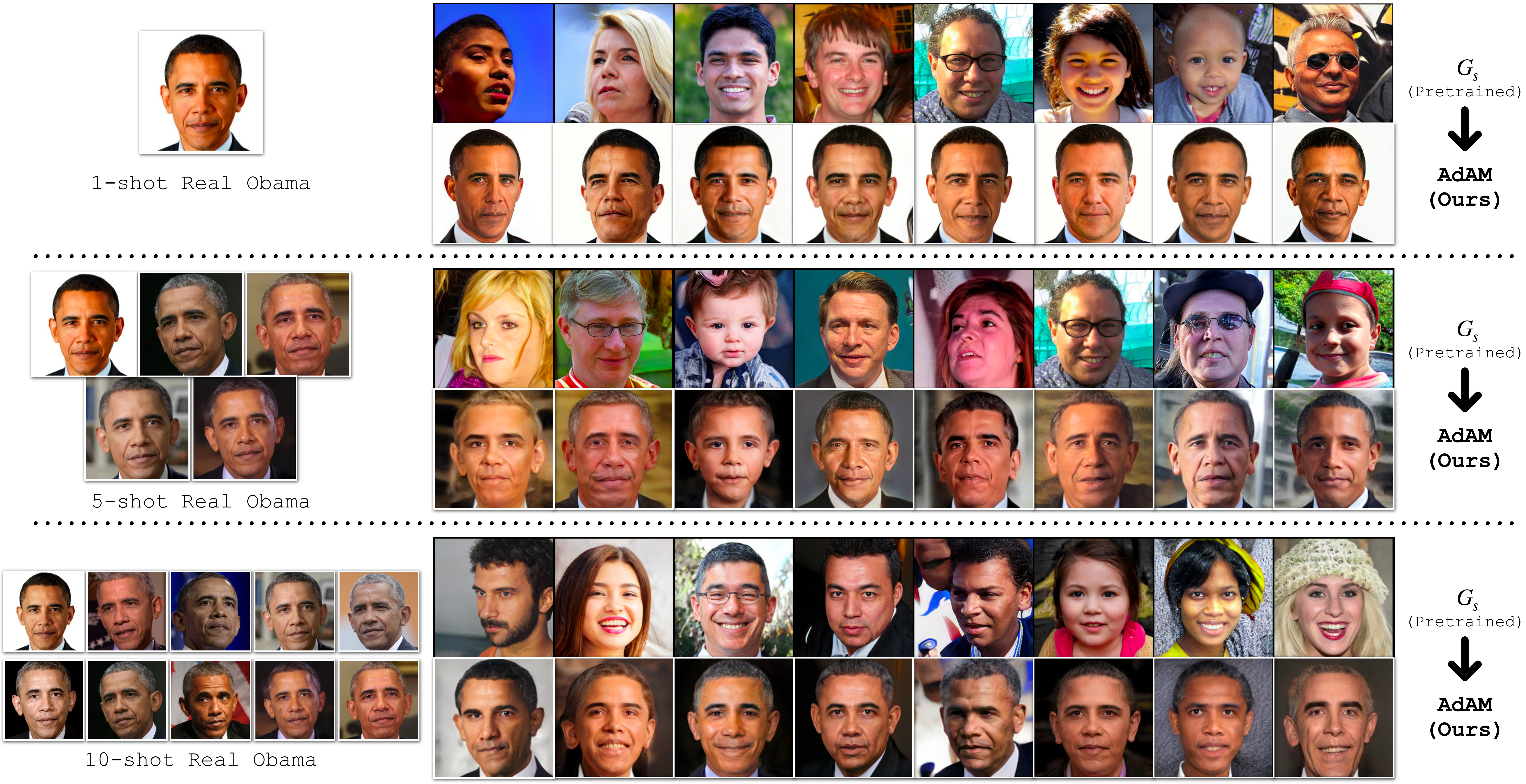}
    \caption{FFHQ $\rightarrow$ Obama Dataset}
    \label{fig:supp_obama}
\end{figure}

Given very limited target domain samples (i.e.: 10-shot), our proposed method achieves SOTA results in FSIG with different source / target domain proximity.
Though our work shows exciting results by pushing the limits of FSIG, we urge researchers, practitioners and developers to use our work with privacy, ethical and moral concerns. 
{
In what next, we bring an example to our discussion.

\textbf{Adapting the pretrained GAN to a particular person.}
The idea of FSIG aims to adapt a pretrained GAN to a target domain with limited samples. It is reasonable that a user of FSIG can take few-shot images of a particular person and generate diverse images of the person, which leads to potential safety and privacy concerns. We conduct an experiment to adapt a pretrained StyleGAN2 generator to Obama dataset \cite{zhao2020differentiable} in 1-shot, 5-shot and 10-shot setups, and the results are shown in \ref{fig:supp_obama}.

Potentially, the method can be adapted to generate images of the same person by applying a more restrictive selection of the source model’s knowledge. However, this would degrade the diversity of the outputs and may not be suitable for general FSIG which our paper focuses on. We will explore such interesting application as our future work and verify with state-of-the-art face recognition systems to understand any potential threats.
}

\subsection{Amount of Compute}
\label{sec-supp:amount_compute}
The amount of compute in this project is reported in Table \ref{table-supp:compute}. 
We follow
NeurIPS guidelines to include the amount of compute for different experiments along with $CO_2$ emission.

\begin{table}[!ht]
\caption{
Amount of compute in this project. The GPU hours include computations for initial
explorations / experiments to produce the reported values. $CO_2$ emission values are
computed using \url{https://mlco2.github.io/impact/}
}
  \begin{adjustbox}{width=\textwidth}
  \begin{tabular}{l|c|c|c}\toprule
\textbf{Experiment} &\textbf{Hardware} &\textbf{GPU hours} &\textbf{Carbon emitted in kg} \\ \toprule
Main paper : Table 2 (Repeated 3 times) &Tesla V100-PCIE (32 GB) &306 & 52.33 \\ \midrule
Main paper : Figure 5 / Figure 6 &Tesla V100-PCIE (32 GB) &136 & 23.26 \\ \midrule
Main paper : Figure 2 &Tesla V100-PCIE (32 GB) &6 &1.03 \\ \midrule
Supplementary : Extended Experiments &Tesla V100-PCIE (32 GB) &68 &11.63 \\ \midrule
Supplementary : Ablation Study &Tesla V100-PCIE (32 GB) &14 &4.1 \\ \midrule
Additional Compute for Hyper-parameter tuning &Tesla V100-PCIE (32 GB) &24 &2.16 \\ \midrule
\textbf{Total} &\textbf{} &\textbf{554} &\textbf{94.73} \\
\bottomrule
\end{tabular}
\end{adjustbox}
\label{table-supp:compute}
\end{table}

\newpage

\end{document}